\documentclass[letterpaper]{article} % DO NOT CHANGE THIS
\usepackage{aaai2026}  % DO NOT CHANGE THIS
\usepackage{times}  % DO NOT CHANGE THIS
\usepackage{helvet}  % DO NOT CHANGE THIS
\usepackage{courier}  % DO NOT CHANGE THIS
\usepackage[hyphens]{url}  % DO NOT CHANGE THIS
\usepackage{graphicx} % DO NOT CHANGE THIS
\def\UrlFont{\rm}  % DO NOT CHANGE THIS
\usepackage{natbib}  % DO NOT CHANGE THIS AND DO NOT ADD ANY OPTIONS TO IT
\usepackage{caption} % DO NOT CHANGE THIS AND DO NOT ADD ANY OPTIONS TO IT
\usepackage[utf8]{inputenc} % allow utf-8 input
\usepackage{booktabs}       % professional-quality tables
\usepackage{amsfonts}       % blackboard math symbols
\usepackage{nicefrac}       % compact symbols for 1/2, etc.
\usepackage{microtype}      % microtypography
\usepackage{xcolor}         % colors

\usepackage{amssymb}            % Defines common symbols like \mathbb R
\usepackage{mathtools}          % Extends amsmath, providing common math tools
\usepackage{mathrsfs}           % Enables \mathscr, which can work in cases that \mathcal does not
\usepackage{graphicx}           % For including images
\usepackage{url}                % For displaying URLs
\usepackage{lipsum}             % For placeholder text
\usepackage{svg}

\usepackage{amsthm}

\usepackage{microtype}
\usepackage{subfigure}
\usepackage{booktabs} % for professional tables

\usepackage{algorithm}
\usepackage{algorithmic}
\usepackage{amsmath}
\usepackage{amsfonts}
\usepackage{amsthm}

\DeclareMathOperator*{\argmax}{arg\,max}
\DeclareMathOperator*{\argmin}{arg\,min}
\theoremstyle{plain}
\newtheorem{theorem}{Theorem}[section]

\newtheorem{lemma}[theorem]{Lemma}

\theoremstyle{definition}
\newtheorem{definition}[theorem]{Definition}
\newtheorem{assumption}[theorem]{Assumption}
\theoremstyle{remark}

\usepackage{newfloat}
\usepackage{listings}
\DeclareCaptionStyle{ruled}{labelfont=normalfont,labelsep=colon,strut=off} % DO NOT CHANGE THIS
\floatstyle{ruled}
\newfloat{listing}{tb}{lst}{}
\floatname{listing}{Listing}
\newcommand{\jhedit}[1]{\textcolor{black}{#1}}
\newcommand{\jhhedit}[1]{\textcolor{black}{#1}}

\newcommand{\ttyedit}[1]{\textcolor{black}{#1}}
\newcommand{\tyedit}[1]{\textcolor{black}{#1}}
\newcommand{\Tyedit}[1]{\textcolor{black}{#1}}
\newcommand{\tyyedit}[1]{\textcolor{black}{#1}}
\newcommand{\ztyedit}[1]{\textcolor{black}{#1}}
\newcommand{\tedit}[1]{\textcolor{black}{#1}}

\newcommand{\cwinline}[1]{\textcolor{black}{#1}}
\newcommand{\cwwinline}[1]{\textcolor{black}{#1}}
\newcommand{\cwwwinline}[1]{\textcolor{black}{#1}}

\makeatletter
\newcommand{\printfnsymbol}[1]{%
  \textsuperscript{\@fnsymbol{#1}}%
}

\usepackage[table]{xcolor}
\definecolor{mountainrow}{gray}{0.90}

\title{Structure Detection for Contextual Reinforcement Learning}
\author{
Tianyue Zhou\equalcontrib, Jung-Hoon Cho\equalcontrib, Cathy Wu
}
\affiliations{
    Massachusetts Institute of Technology\\
    \{tianyuez, jhooncho, cathywu\}@mit.edu
}

\usepackage{bibentry}
\begin{document}

\maketitle
\begin{abstract}
Contextual Reinforcement Learning (CRL) tackles the problem of solving a set of related Contextual Markov Decision Processes (CMDPs) that vary across different context variables. Traditional approaches---independent training and multi-task learning---struggle with either excessive computational costs or negative transfer. A recently proposed multi-policy approach, Model-Based Transfer Learning (MBTL), has demonstrated effectiveness by strategically selecting a few tasks to train and zero-shot transfer. However, CMDPs encompass a wide range of problems, exhibiting structural properties that vary from problem to problem. As such, different task selection strategies are suitable for different CMDPs. In this work, we introduce Structure Detection MBTL (SD-MBTL), a generic framework that dynamically identifies the underlying generalization structure of CMDP and selects an appropriate MBTL algorithm. For instance, we observe \textsc{Mountain} structure in which generalization performance degrades from the training performance of the target task as the context difference increases. We thus propose M/GP-MBTL, which detects the structure and adaptively switches between a Gaussian Process-based approach and a clustering-based approach. Extensive experiments on synthetic data and CRL benchmarks—covering continuous control, traffic control, and agricultural management—show that M/GP-MBTL surpasses the strongest prior method by 12.49\% on the aggregated metric. These results highlight the promise of online structure detection for guiding source task selection in complex CRL environments. 
% Code, datasets, and a project webpage are available at \url{https://mit-wu-lab.github.io/SD-MBTL/}.
\end{abstract}

\begin{links}
    \link{Code}{https://github.com/mit-wu-lab/SD-MBTL/}
    \link{Webpage}{https://mit-wu-lab.github.io/SD-MBTL/}
    % \link{Extended version}{https://aaai.org/example/extended-version}
\end{links}

\section{Introduction}
\jhedit{Despite the recent success of deep reinforcement learning (RL), deep RL often \jhedit{struggles to solve} r}eal-world applications \jhedit{that} involve families of related tasks that differ in a few key parameters but \jhedit{mostly} share underlying dynamics \citep{bellemare2020autonomous, degrave2022magnetic,benjamins2023contextualizecasecontext}.
Such contextual variations naturally arise in robotics (e.g., different payload weights or terrain conditions) \citep{yu2020meta} and traffic control (e.g., varying traffic inflows or signal timing) \citep{jayawardana2025intersectionzoo}.
Formally, these settings can be modeled as Contextual Markov Decision Processes (CMDPs)~\citep{hallak2015contextual, modi2018markov, benjamins2023contextualizecasecontext}, where each task is specified by a context. 
\tyyedit{Moreover, we extend prior studies to more challenging multi-dimensional CMDP, rather than a single-dimensional context.}

\jhedit{Due to the curse of dimensionality in context variables, a} central challenge in CMDPs is how to efficiently train policies that generalize to \jhedit{numerous related} tasks without starting \jhedit{training} from scratch. Existing paradigms \jhedit{for CMDPs} include: (1) Independent training of a separate policy for each task, which is straightforward but computationally expensive for large task families; (2) Multi-task training of a single universal policy conditioned on the context, which can suffer from \jhedit{limited model capacity or} negative transfer if tasks are too dissimilar~\citep{kang_learning_2011, standley_which_2020}; and (3) Multi-policy training \jhedit{on a small subset of} source tasks \jhedit{\emph{with zero-shot transfer} to new tasks, raising the key question of how to choose that subset.}
Recent work has shown that carefully selecting which tasks to train on can lead to strong generalization and improved sample efficiency \citep{cho2023temporaltransferlearningtraffic, MBTL}\jhedit{, since the cost of full training far outweighs the \jhhedit{cost of policy evaluation and the task-selection.}} 

\begin{figure}[t]
    \centering
    \includegraphics[width=\linewidth]{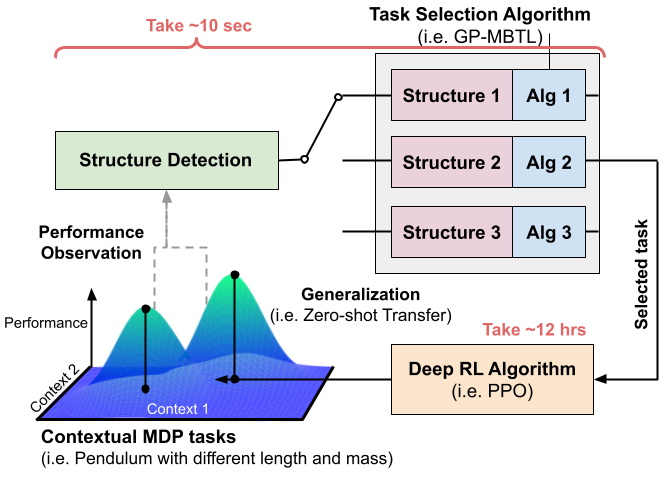}
    \caption{\textbf{Conceptual overview.} \jhedit{SD-MBTL detects an underlying structure from observed generalization performance and selects an appropriate algorithm.}}
    \label{fig:sd-mbtl-overview}
\end{figure}
\jhedit{Multi-policy training requires selecting a subset of tasks in CMDPs, formulated as a \jhedit{Greedy} Tasks Selection Problem (\jhedit{G}STS). However, many CMDPs exhibit specific structural patterns that can be exploited. 
\tyyedit{For example, \cite{MBTL} splits generalization performance into training performance and a generalization gap, modeling the gap as a linear function. 
Inspired by this, we define a task structure as a functional decomposition of generalization performance in which certain components satisfy designated properties.} 
\jhedit{W}e \jhedit{thus} propose a generic \emph{Structure Detection Model-Based Transfer Learning} (SD-MBTL) framework, which detects the underlying structure and adapts its task selection strategy accordingly. 
In this paper, we instantiate the SD-MBTL framework with \textbf{M/GP-MBTL}, which uses two specific structures and dynamically switches between task selection algorithms--Gaussian Process-based and clustering-based MBTL (Figure~\ref{fig:sd-mbtl-overview}).}
% \cwww{Figure text is very small!}

\jhedit{For example, we \cwwwinline{observe experimentally} that} \jhedit{generalization in CMDPs \cwwwinline{often} follow some structure, in which we find some task selection approach works better.}
\jhedit{Thus, d}etecting such structure enables a more targeted approach to source task selection than a one-size-fits-all approach.
To capitalize on this insight, we \jhedit{also devise} a fast and effective algorithm\jhedit{, \textsc{Mountain} Model-Based Transfer Learning (}M-MBTL\jhedit{), b}y reducing \jhedit{GSTS} problem to a \jhedit{sequential version of} clustering problem upon a certain \textsc{Mountain} structure. It enables us to leverage clustering \tyedit{loss to efficiently guide the search of the training task in a continuous space.}
% \jhfn{Could you double check if this is true?} 
% This simplified model is both conceptually intuitive and empirically effective for multi-dimensional contexts.

% Recognizing that the Constant-Peak structure may not hold in every complex real-world CMDP, we further introduce \textit{M/GP-MBTL} that detects the underlying structure and accordingly selects the most suitable MBTL algorithm (Figure~\ref{fig:hybrid-mbtl-overview}). When the Constant-Peak structure is detected, \jhedit{M-MBTL} is employed to efficiently guide source task selection; otherwise, a revised BO-based MBTL algorithm is used.

\jhedit{Our main contributions are summarized as follows:}
\begin{itemize}
    \itemsep0em 
    % \item We introduce a \jhedit{generic structure detection} framework \jhedit{for CRL, which leverages a suite of structure detection criteria and task selection algorithms.}\cwww{Be more direct about the main contribution. (So what?)}
    \item We introduce \jhedit{SD-MBTL, a unified framework that detects the underlying generalization structure of CMDPs and adapts its source‐task selection strategy, resulting in improved sample efficiency in diverse CRL settings.}
    \item \jhedit{We propose \jhedit{M/GP-MBTL, a practical structure detection algorithm,} using two specific structures \jhedit{in} the generalization performance.} \jhedit{We also develop a clustering-based MBTL algorithm for \textsc{Mountain} structure.}
    \item We \jhedit{empirically validate our approach in a \Tyedit{multi-dimensional} synthesized dataset and \jhedit{real-world} CMDP benchmarks, \cwwwinline{including}} continuous control (\jhedit{CartPole and BipedalWalker \cite{benjamins2023contextualizecasecontext}), cooperative eco-driving (IntersectionZoo \cite{jayawardana2025intersectionzoo}), and crop management (CyclesGym \cite{turchetta2022learning})}. \tedit{On the aggregated metric normalized between the random baseline and the oracle, M/GP-MBTL outperforms the previous best method by 12.49\%.}
\end{itemize}
\cwwinline{The rest of the article is structured as follows}\jhedit{: Section~\ref{sec:preliminaries} reviews CMDP and GSTS problem. Section~\ref{sec:sd-mbtl} introduces the SD-MBTL framework. Section~\ref{sec:M/GP-MBTL} introduces M/GP-MBTL. Section~\ref{sec:experiments} presents evaluations on benchmarks, Section~\ref{sec:related-works} discusses related work, and Section~\ref{sec:conclusion} concludes.}
% \cw{Convey the logic of the article.}
\section{Preliminaries}\label{sec:preliminaries}
\subsection{Contextual Markov Decision Process}
We consider a \emph{contextual} Markov decision process (CMDP) defined over a family of MDPs indexed by a \emph{context} (or \emph{task}) variable in a multi-dimensional space. 
Let the finite set of all MDPs (or tasks) be denoted by $Y = \{ y_1, y_2, \dots, y_N \}$,
where each task $y_n \in \mathbb{R}^D$ for some dimension $D$, and $N = |Y|$ is the total number of tasks (or target tasks). Specifically, let each task $y \in Y \subset \mathbb{R}^D$ parameterize a distinct MDP $\mathcal{M}_y = (S, A, P_y, R_y, \rho_y),$ where $S$ is the state space, $A$ is the action space, $P_y$ represents transition dynamics, $R_y$ is the reward function, and $\rho_y$ is the initial state distribution for the task $y$ \citep{hallak2015contextual,modi2018markov}. Hence, the CMDP is the collection $\{\mathcal{M}_y\}_{y \in Y}$.

We define a continuous set of tasks \(X \subset \mathbb{R}^D\) such that \(X\) has the same bound as \(Y\). 
% \jhfn{We had a comment why $X$ should be defined separately as $Y$. I personally think it's fine, but for the clarity of the paper, let's revisit this.}
Intuitively, \(X\) captures an underlying continuous space of tasks from which \(Y\) can be viewed as a discrete subset ($Y\subset X$).
\jhedit{Given a} context \(x \in X\), \jhedit{one can} train a policy \(\pi_x\) \jhedit{(using, for instance, an off-the-shelf RL algorithm}) and \jhedit{get} the \textit{(source) training performance} on task \(x\) by $J(\pi_x,\,\tyedit{x}) = \mathbb{E}\bigl[\text{return of } \pi_x \text{ in } \mathcal{M}_{\tyedit{x}} \bigr].$
% \tyfn{Write this formally?}
% \cw{Do you mean $J(\pi_x,\,x)$?}
\jhedit{To zero-shot transfer} from a source task \(x\) to a target task \(y\), we apply the policy \(\pi_x\) (trained on \(\mathcal{M}_x\)) within \(\mathcal{M}_y\), \tyyedit{potentially} at reduced performance compared to training directly on \(y\). 
The resulting \emph{generalization performance} is \(J(\pi_x,\,y)\).  
We quantify the \jhedit{zero-shot} generalization gap when transferring from \(x\) to \(y\) as $\Delta J(\pi_x,\,y)=J(\pi_x,\,x)-J(\pi_x,\,y)$. 
\subsection{Problem Formulation}

\textbf{\jhedit{Greedy} source task selection} \tyedit{(\jhedit{G}STS)} problem\footnote{\jhedit{ This was defined as the Sequential Source Task Selection problem in \citep{MBTL}.} \ztyedit{We renamed the problem because its goal is to greedily select a training task at each step, and the term “sequential” fails to convey this aspect.}}
\tyyedit{Given a CMDP with a finite set of \emph{target tasks} $Y$, our goal is to identify a sequence of tasks on which to perform full RL training, so as to achieve high generalization performance on all tasks in \(Y\). 
Specifically, at each step \(k\), we pick a new source task \(x_k\in X\) to train a corresponding policy \(\pi_{x_k}\). 
For each task in $Y$, we choose the best performing policy from a set of trained policies, which maximizes the performance.
The objective at each round \(k\) is to \textit{greedily} select a new source task \(x_k\) that maximizes the expected performance across the entire set \(Y\). }
% \tyyedit{However, since the search space $X^K$ grows exponentially with $K$, solving this optimization problem directly is computationally intractable. To address this, we consider a greedy variant of the objective, referred to as \textbf{Greedy Source Task Selection}.}
% \jhfn{These two sentences also hold for SSTS (above).}
% After each new source task \(x_k\) is trained, we can update the best available performance for each target task \(y\), denoted as $\jhedit{J}_k(\jhedit{y})=\max\Bigl(\jhedit{J}_{k-1}(\jhedit{y}),\, J(\pi_{x_k},\,\jhedit{y})\Bigr),$ with \(\jhedit{J}_0(\jhedit{y})\) initialized to zero at the outset.
% Hence, the objective at round \(k\) is to select a new source task \(x_k\) that maximizes the \emph{expected improvement} in performance across the entire set \(Y\). 
One can write this as:
% \begin{equation}
%     x_k=\argmax_{\,x \in \jhedit{X} \setminus \{x_{1:k-1}\}}
%     \mathbb{E}_{\jhedit{y} \sim \mathcal{U}(Y)} 
%     \Bigl[\max\Bigl(\jhedit{J}_{k-1}(\jhedit{y}),\; J(\pi_x,\,\jhedit{y})\Bigr)\Bigr]\tyyedit{.}
% \end{equation}
\begin{equation}
    x_k=\argmax_{\,x \in \jhedit{X} \setminus \{x_{1:k-1}\}}
    \mathbb{E}_{\jhedit{y} \sim \mathcal{U}(Y)} \tyyedit{[\max_{x^\prime\in x_{1:k-1}\cup \{x\}} J(\pi_{x^\prime}, y)]}
\end{equation}
where $x_{1:K}=x_1,\dots,x_{K}$, and \(\mathcal{U}(Y)\) is a uniform distribution over the finite task set \(Y\).
% where $x_{1:k-1}=x_1,\dots,x_{k-1}$ and \(\mathcal{U}(Y)\) is a uniform distribution over the finite task set \(Y\).
In \cwwinline{general}, an exact solution to GSTS problem\jhedit{s} can be intractable for large-scale or continuous task spaces. \cwwinline{In practice, many applications have fairly small task spaces (e.g., 3-10 context dimensions, each discretized to 100 values). The core challenge is that solving \textit{any} task MDP is fairly expensive and thus the number of task MDPs to solve should be minimized.}
MBTL \cwwinline{aimed to circumvent naive exploration} by modeling the training performance and generalization gap and optimizing the selection via Bayesian optimization~\cite{MBTL}.
\subsection{Model-Based Transfer Learning}
\label{sec:gp-mbtl}
A practical solution to solve \jhedit{G}STS problem is provided by MBTL~\citep{MBTL}, which we refer to as \textbf{GP-MBTL} in this work.
GP-MBTL discretizes the continuous source-task space, models source-task returns with Gaussian-Process regression, and approximates the generalization gap as a linear function of context similarity. At each iteration \(k\), it selects the source task \(x_k\) that maximizes an acquisition function combining predicted source task performance \(\mu_{k-1}(x)\), uncertainty \(\sigma_{k-1}(x)\), and the estimated generalization gap to target tasks. The policy trained on \(x_k\) is then evaluated, and the GP posterior updated, progressively improving coverage of the target set \(Y\). We additionally use observed transfer performance to refine the acquisition rule and estimate the gap’s slope online. Full derivations, algorithmic details, and hyperparameters are provided in Appendix~\ref{sec:detail_GP-MBTL}.

\section{Structure Detection Model-Based Transfer Learning}\label{sec:sd-mbtl}
\jhedit{CMDPs with m}ulti-dimensional context spaces typically demand \jhedit{significantly} more data to keep model predictions accurate, which challenges the Gaussian-process module in \tyyedit{GP-MBTL} and complicates the exploration–exploitation trade-off.  \tyedit{When a CMDP obeys a recognisable structure, however, that structure can be \cwinline{detected and then} exploited to curb unnecessary exploration.  We therefore introduce a generic \textbf{Structure Detection Model-Based Transfer Learning} (SD-MBTL) framework for task selection.}

\subsection{\tyedit{Generalization Performance Structure Decomposition}}
\tedit{To support the identification and analysis of CMDP structures, we decompose the generalization-performance structure inspired by the Sobol–Hoeffding (functional-ANOVA) decomposition, which uniquely and orthogonally splits any square-integrable multivariate function into additive main-effect terms and interaction terms \cite{hoffding1948class, sobol1990sensitivity}. More details on how we derive this decomposition are provided in Appendix~\ref{sec:decomposition}.}
% \jhfn{maybe you can add some sentences that relate the Sobol-Hoeffding decomposition to our definition.}
\begin{definition}[\jhedit{Generalization Performance Structure Decomposition}]\label{def:general_model}
    \tedit{For any task pair $(x,y)$, define $C\!:=\!\mathbb{E}_{x\in X,y\in Y}[J(\pi_x,y)]$, 
    $g(y)\!:=\!\mathbb{E}_{x\in X}[J(\pi_x,y)]-C$,
    $f(x)\!:=\!J(\pi_x,x)-g(x)-C$,  and $h(x,y)\!:=\!J(\pi_x,y)-f(x)-g(y)-C$, thus we can get:}
    \begin{equation}
        J(\pi_x, y)=f(x)+g(y)+h(x,y)\tedit{+C},
    \end{equation}
    and $h(x,y)=0$ if $x=y$.
\end{definition}
% Based on empirical observations detailed in Section~\ref{sec:structure-observations},
\tedit{Based on empirical observations, we found three components that affect generalization performance: 1) The intrinsic quality of source policy influences whether it has good generalization across many tasks. 2) The difficulty of the target task affects whether transferred policies can achieve high performance on it. 3) The dissimilarity between source and target tasks degrades generalization performance, as large differences in context usually yield poor generalization. Thus, we name $f(x)$ as the \textbf{policy quality}, $g(y)$ as the \textbf{task difficulty}, and $h(x,y)$ as the \textbf{task dissimilarity} between source and target tasks.}
\begin{figure*}[!t]%
    \centering
    \subfigure[CartPole]{%
        \includegraphics[width=0.33\textwidth]{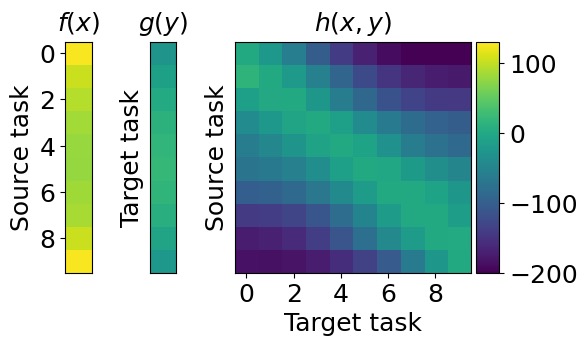}
        \label{fig:cartpole_structure}%
        }%
        \subfigure[BipedalWalker]{%
        \includegraphics[width=0.33\textwidth]{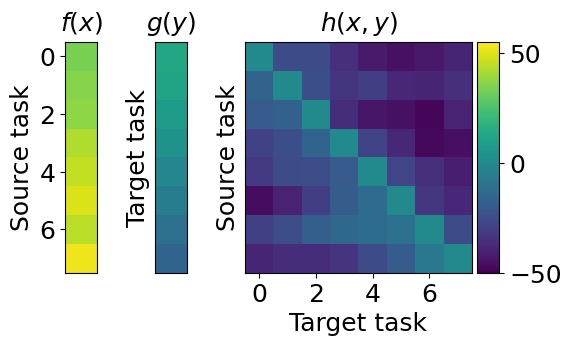}
        \label{fig:walker_structure}%
        }%
    \subfigure[\jhedit{CyclesGym}]{
        \includegraphics[width=0.33\textwidth]{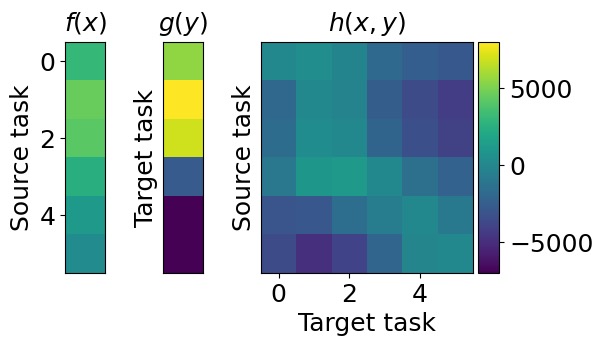}
        \label{fig:crop_structure}
        }%
    \caption{\tedit{Heatmap of policy quality $f(x)$, task difficulty $g(y)$, and task dissimilarity $h(x,y)$ for CartPole (mass of cart), BipedalWalker (scale), and CyclesGym (precipitation) averaged over three different random seeds. In these tasks, $f(x)$ is nearly constant. $h(x, y)$ decreases approximately linearly as the context difference increases, resembling a distance metric.}}
    % \caption{\tyyedit{\jhedit{Heatmap of t}ransfer matrices for CartPole (\jhedit{mass of cart}), BipedalWalker (gravity), and \jhedit{CyclesGym} (temperature) \jhedit{averaged over three different random seeds.} Each element represents the \jhedit{generalization} performance when transferring from the source task context to the target task. In the subtracted matrix \jhedit{(rightmost)}, each column is \ztyedit{subtracted} by the task difficulty. 
    % }}
    \label{fig:average_transfer_matrix}
\end{figure*}

\tedit{Based on the decomposition, we outline the procedure for addressing a CMDP by exploiting a specific structure.} Within a CMDP structure, one or more of $f(x)$, $g(y)$, and $h(x,y)$ may satisfy particular assumptions or properties, allowing us to treat them as known components of $J(\pi_x, y)$. Thus, a corresponding algorithm can be used to learn the remaining unknown components and select training tasks based on the model so as to maximize the objective.
\subsection{SD-MBTL}\label{sec:SD-MBTL}
SD-MBTL embeds a dynamic structure detection mechanism inside the MBTL loop \tedit{and chooses the corresponding MBTL algorithm based on the detected structure}.
Its inputs are (i) a set of candidate CMDP structures~$\mathcal{S}$, (ii) a detection routine  
$\text{Detect}:\mathbb{R}^{k\times N}\!\rightarrow\!\mathcal{S}$, and  
(iii) a library of MBTL algorithms $\{\text{Alg}_1,\dots,\text{Alg}_{|\mathcal{S}|}\}$ matched one-to-one with the structures in~$\mathcal{S}$.  
At each decision round, SD-MBTL uses the observed generali\jhedit{z}ation performances  
$\{J(\pi_{x_\kappa},y_n)\}_{n\in[N],\,\kappa\in[k]}$  
to infer the current structure $s_i=\text{Detect}(\cdot)$, then invokes the corresponding algorithm~$\text{Alg}_i$ to choose the next training task~$x$.  After training, the new policy~$\pi_x$ is transferred to all target tasks, yielding zero-shot generalization performances $\{J(\pi_x,y_n)\}_{n\in[N]}$.  
% \tyedit{Accurate detection allows SD-MBTL to run an algorithm tailored to the true structure, thereby cutting data requirements and training cost.}
\ttyedit{More details in SD-MBTL are provided in Appendix \ref{sec:detail_SD-MBTL}.}
% Below, we present the details about key algorithms that form the backbone of our SD-MBTL framework.
% Algorithm~\ref{alg:SD-MBTL} outlines the generic Structure Detection Model-Based Transfer Learning (SD-MBTL) framework. \Tyedit{The input of SD-MBTL includes a task set $Y$, a CMDP structures set $S$, a detection algorithm $\text{Detect}:\mathbb{R}^{k\times N}\to S$, and MBTL algorithms ($\text{Alg}_1,\text{Alg}_2,...,\text{Alg}_{|S|}$).} In each decision round, the algorithm first detects the underlying CMDP structure based on observed generalization performance and then uses a mapping function to select the corresponding MBTL algorithm. The chosen algorithm then selects the next training task, which is subsequently used to update the performance estimates.

\section{\tedit{M/GP Model-Based Transfer Learning}}\label{sec:M/GP-MBTL}
\tyedit{As a concrete instantiation of SD-MBTL, we propose \textbf{M/GP-MBTL}, which targets a special CMDP structure---\textsc{Mountain}}.
When the CMDP satisfies \textsc{Mountain}, the \jhedit{G}STS problem reduces to clustering, and we employ the clustering-based \jhedit{M-MBTL}.  Otherwise, we revert to the more general GP-MBTL. This two-way strategy combines the sample efficiency of M-MBTL with the robustness of GP-MBTL.

\subsection{\textsc{Mountain} Structure}\label{sec:mountain}
\tedit{In our experiments, we observe that some CMDP problems exhibit similar model structures. Figure~\ref{fig:average_transfer_matrix} shows the policy quality $f(x)$, task difficulty $g(y)$, and task dissimilarity $h(x,y)$ in CartPole, BipedalWalker, and CyclesGym benchmarks. In these tasks, $f(x)$ is nearly constant. $h(x, y)$ decreases approximately linearly as the context difference increases, resembling a distance metric. These similar characteristics suggest that they may originate from the same underlying structure.}
\tedit{Thus, we identify a specific generalization structure---termed \textsc{mountain} structure---that relies on the following key assumptions.}
\begin{assumption}[Constant Source Task Influence]
    \label{assump:constant}
    % \ztyedit{Policy quality is constant, i.e., (very) low variance: $f(x)=C_1$.}
    \ztyedit{Policy quality is constant: $f(x)=C_1$.} 
    % The influence of source task on generalization gap is constant, \cwwinline{i.e., (very) low variance}: .\cwww{Is this a different interpretation as provided in Section 3.1?}
    % \begin{equation}\label{eq:constant_f}
    %         f(x)=C_1.
    %     \end{equation}
\end{assumption}
\begin{assumption}[Distance Metric for Task Dissimilarity]
    \label{assump:distance}
    The task dissimilarity $h(x,y)$ is represented as a distance metric: $h(x,y)=-\text{dist}(x,y)$.
    \tyedit{The distance metric \citep{Cech1969} is a function $\text{dist}:\mathbb{R}^D\times\mathbb{R}^D\rightarrow \mathbb{R}$ which satisfies $\text{dist}(x,x)=0\,\  \forall x$, positivity, symmetry, and the triangle inequality. In this paper, we use the $L_1$ norm as the distance metric for our algorithms.}
\end{assumption}
The linear generalization gap assumption in \cite{cho2023temporaltransferlearningtraffic} and \cite{MBTL} \tyyedit{assumes $J(\pi_x,y)-J(\pi_x,x)=g(x)-g(y)+h(x,y)=k|x-y|$. It} can be considered a specific case of Assumption~\ref{assump:distance}, where the CMDP is one-dimensional and target task difficulties \( g(y) \) are constant.
% \jhedit{It also reflects our observations that more \jhedit{distant} contexts lead to lower generalization performance.}

\begin{definition}[\jhedit{\textsc{Mountain} Structure}]
\label{def:mountain}
\jhedit{\textsc{Mountain}} is a CMDP structure where Assumption \ref{assump:constant} and \ref{assump:distance} holds. Or equivalently, $J(\pi_x,y)=J(\pi_y,y)-\text{dist}(x,y).$
% \begin{equation}\label{z}
% J(\pi_x,y)=J(\pi_y,y)-\text{dist}(x,y).
% \end{equation}
\end{definition}
The proof of equivalence is provided in Appendix \ref{sec:equivalence}. We refer to this as \jhedit{\textsc{Mountain}} because the training performance of the target task resemble\jhedit{s} mountain peaks, and the generalization performance acts \jhedit{like a slope that decreases as the distance between source and target tasks increases}. \jhedit{This perspective effectively reduces} \jhedit{G}STS to a clustering problem, which will be discussed in Section \ref{sec:SC-MBTL}.

\subsection{\jhedit{Structure} Detection}\label{sec:detection_approach}
For an unknown CMDP, we aim to determine whether it satisfies \textsc{Mountain} structure. At each step $k$ we have observed the generalization performances $J(\pi_x,y)$ for all $x\in\{x_{1},\ldots,x_{k}\}$. Our goal is to test whether $f(x)$ and $\operatorname{dist}(x,y)$ individually obey Assumptions \ref{assump:constant} and \ref{assump:distance}. Because \textsc{Mountain} structure does not include an explicit \cwwwinline{task difficulty} term $g(y)$, we first remove the influence of \cwwwinline{task difficulty} on $J(\pi_x,y)$ when performing \ttyedit{structure detection}. \ztyedit{Then we apply two criteria: (i) a Small Variance Criterion to check whether policy quality is almost constant (Assumption \ref{assump:constant}), and (ii) a Slope Criterion to check that the majority of slopes are decreasing like a distance metric (Assumption \ref{assump:distance}). If both criteria are satisfied, we tag the CMDP \jhedit{as} \textsc{Mountain}}.

\textbf{Remov\cwwwinline{ing} the influence of \cwwwinline{task difficulty}.} \ztyedit{For each step $k$,} we \jhedit{estimate} the difficulty \jhedit{of target task} by the mean generalization performance of trained policies.
\begin{equation}\label{eq:approx_J_target}
    \tedit{g(y) \approx \tyedit{\mathbb{E}_{x\in \Tyedit{x_{1:k}}}}[J(\pi_x, y)]}-C.
\end{equation}
\tedit{This estimate replaces $x \in X$ in the Definition \ref{def:general_model} with $x \in x_{1:k}$, because at step $k$ we only have access to the information in $x_{1:k}$.}
% It also ignores the constant $C$, which has no effect on structure detection.
% \ztyedit{We make this estimate because, when the target task is simpler, different policies generally achieve better generalization performance, and vice versa.} 
% \tedit{We also provide Lemma \ref{lemma:estimation} in Appendix \ref{sec:lemma_est} which justifies the low error of this estimation.}
\ztyedit{For any training task set $x_{1:k}$ at step $k$,} we define the relative generalization performance as the difference between real generalization performance and its expectation over training tasks: $\overline{J}(\pi_x,y):=J(\pi_x,y)-\mathbb{E}_{x^\prime \in \Tyedit{x_{1:k}}}[J(\pi_{x^\prime},y)]$.
By removing the influence of task difficulty $g(y)$ by Equation \ref{eq:approx_J_target}, we can have
$\overline{J}(\pi_x,y)\approx f(x)+h(x,y)$.
% \begin{equation}\label{eq:constant_peak}
%     \overline{J}(\pi_x,y)\approx \ttyedit{f(x)+h(x,y)+\tedit{2C}.}
% \end{equation}
% \jhedit{We thus} propose two criteria: \Tyedit{Small Variance} Criterion to detect Assumption \ref{assump:constant} and Slope Criterion to detect \jhedit{Assumption}~\ref{assump:distance}. 

\textbf{\Tyedit{Small Variance} Criterion.}
\ttyedit{Although Assumption \ref{assump:constant} treats \ztyedit{policy quality} $f(x)$ as a constant, a small variance in $f(x)$ does not materially distort the structure. We therefore compare the standard deviations of $f(x)$ \jhedit{with} \ztyedit{task dissimilarity} \ttyedit{$h(x,y)$} to verify that $f(x)$ indeed exhibits \jhedit{small} deviation.}
\begin{lemma}[\tyyedit{Relative Influence of \ztyedit{Policy Quality and Task Dissimilarity}}]\label{lemma:compare_std} 
    \ztyedit{If $\text{std}_{x\in x_{1:k}}(\overline{J}(\pi_x,x))<\mathbb{E}_{x\in x_{1:k}}[\text{std}_{y\in Y}(\overline{J}(\pi_x,y))]$, then we have} $\tyyedit{\text{std}_{x\in x_{1:k}}(f(x))<\mathbb{E}_{x\in x_{1:k}}[\text{std}_{y\in Y} (\ttyedit{h(x,y)})].}$
    % \cwww{Fix this line to make it easier to read. Also, which way is the implication? Is it an if and only if? The writing is not clear. The implication seems to be if A then B, but not sure.}
\end{lemma}
When Lemma \ref{lemma:compare_std} holds, 
\ztyedit{policy quality} $f(x)$ has a relatively small variance. The proof is provided in Appendix \ref{sec:proof-compare_std}. Define $\mathbb{I}(\cdot)$ as the indicator function. \ztyedit{Thus, for any training task set $x_{1:k}$ at step $k$,} we have the following criterion:
% \cwww{Same comment. Not well defined. Need conditioning on $k$.}
\begin{equation}
\begin{aligned}
   & \text{\Tyedit{Small Variance} Criterion}\\
    :=&\mathbb{I}\left(\text{std}_{x\Tyedit{\in x_{1:k}}}(\overline{J}(\pi_x,x))<\mathbb{E}_{x\Tyedit{\in x_{1:k}}}[\text{std}_{y\Tyedit{\in Y}}(\overline{J}(\pi_x,y))]\right).
\end{aligned}
\end{equation}
\textbf{Slope Criterion}. 
A distance metric $\text{dist}(x,y)$ shows both downward slopes on both sides of the source task, while Assumption \ref{assump:distance} can be violated by the case where $h(x,y)$ have different slope signs.
\ztyedit{To capture this, we regress the observed generalization performance on signed context differences and extract the left- and right-hand slope vectors \(\theta_{\text{L}},\theta_{\text{R}}\in \mathbb{R}^D\).}
We verify this structure by checking whether the majority of slopes are indeed decreasing, which is equivalent to confirming that the slope signs on both sides are the same, excluding cases where both sides slope upward. \jhedit{Thus, we have:}
% \cwww{I would like to see this section simplified.}
\begin{equation}\label{eq:slope_criterion}
    % \text{Slope Criterion}=\mathbb{I}\left(\frac{1}{D}\sum_{d=1}^D \mathbb{I}(\text{sgn}(\theta_{\text{left}}^d)=\text{sgn}(\theta_{\text{right}}^d))>0.5\right),
    \text{Slope criterion}:= \mathbb{I}\!\left( \tfrac{1}{D}\sum_{d=1}^D \mathbb{I}[\operatorname{sgn}(\theta_L^d) = \operatorname{sgn}(\theta_R^d)] > \tfrac12 \right),
\end{equation}
\ttyedit{where $\theta^d_{\text{L}}$ and $\theta^d_{\text{R}}$ represents the slope in the $d$-th dimension.} \ztyedit{More details are provided in Appendix \ref{sec:slope_criterion}.} \jhedit{W}e define \( \text{Detect}(\{ J(\pi_\kappa, y_n) \}_{n \in [N], \kappa \in [k-1]}) = \textsc{Mountain} \) if both \jhedit{criteria} are satisfied; otherwise, the result is \textsc{None}.

\subsection{\jhedit{\textsc{Mountain}} Model-Based Transfer Learning}\label{sec:SC-MBTL}
\tyyedit{Under \textsc{Mountain}, \jhedit{G}STS problem is reduced to a sequential clustering problem (Lemma \ref{lemma:reduce}). \jhedit{Proof is provided in Appendix~\ref{sec:proof-reduction}.} Specifically, the reduced problem is a sequential minimization \jhedit{of} the total distance between target tasks and their corresponding source tasks. Thus, a clustering loss function can be used to search for training tasks fast and accurately in a continuous set $X$.}
\begin{lemma}[Reduction of \jhedit{G}STS to Clustering]\label{lemma:reduce} 
With the \jhedit{\textsc{Mountain}} structure, \jhedit{G}STS problem reduces to the sequential version of the clustering problem. Thus, a clustering loss function can be used to search for training tasks fast and accurately in a continuous set $X$.
\begin{equation}
    \tyyedit{\text{for } k\in[K]:\, x_k=\argmin_{x} \,\mathbb{E}_{y_n}[\min_{x^\prime\in x_{1:k-1}\cup \{x\}}\text{dist} (x^\prime, y_n)]}
\end{equation}
\end{lemma}
\tyyedit{Our \textsc{Mountain} Model-Based Transfer Learning (M-MBTL) extends K-Means to this sequential setting. Because fixing earlier centroids can trap the search in local optima, we adopt a random restart technique \citep{random_restart}: sample $M$ target tasks as initial centroids, locally refine each by the clustering loss above, and choose the one with the lowest loss as the training task for that round.  We repeat this procedure for $K$ rounds, training a policy on each selected task and evaluating it on all targets. Full algorithms, including our techniques to reduce the time complexity, are provided in Appendix \ref{sec:M-MBTL}. We also provide a comparison between M-MBTL and GP-MBTL in Appendix \ref{sec:comparison}.}

\begin{algorithm}[H]
    \caption{M/GP-MBTL}
    \label{alg:Hybrid-MBTL}
    \begin{algorithmic}
        \STATE {\bfseries Input:} Task set $Y=\{y_n\}_{n\in[N]}$ 
        \FOR{$k\in[K]$}
            \STATE $s=\text{Detect}(\{\Tyedit{J(\pi_\kappa, y_n)}\}_{n\in[N], \kappa\in[\Tyedit{k-1}]})$
            \IF{\jhedit{$s=$\textsc{Mountain}}}
                \STATE \jhedit{$x_k\gets$ Run M-MBTL}
            \ELSE
                \STATE \jhedit{$x_k\gets$ Run GP-MBTL}
            \ENDIF
            \STATE Train on $x_k$, receive $\{\Tyedit{J(\pi_k, y_n)\}}_{n\in[N]}$
        \ENDFOR
    \end{algorithmic}
\end{algorithm}

\subsection{\cwinline{M/GP-MBTL}}
\tyyedit{Based on the two algorithms and the detection approach above, we introduce a concrete instantiation of SD-MBTL, named M/GP-MBTL (Algorithm \ref{alg:Hybrid-MBTL}). This algorithm uses the function $\text{Detect}(\cdot)$ mentioned in Section \ref{sec:detection_approach} to detect the problem structure. Then it dynamically chooses between M-MBTL for \textsc{Mountain} structures and GP-MBTL for more general settings, enabling the algorithm to tailor its task selection strategy to the underlying CMDP structure.}
% \begin{equation}
%     \Tyedit{\text{Detect}(\{\Delta J(\pi_\kappa, y_n)\}_{n\in[N], \kappa\in[k]})=\text{Small Variance Criterion }\land \text{ Slope Criterion}}
% \end{equation}
% Based on this decision, the algorithm dynamically chooses between a clustering-based method (M-MBTL) for \textsc{Mountain} structures and a Gaussian Process–based method (GP-MBTL) for more general settings. This adaptive approach enables the algorithm to tailor its task selection strategy to the underlying CMDP structure.

\section{Experiments}\label{sec:experiments}
In this section, we present an extensive evaluation of our approaches, \ttyedit{organi\jhedit{z}ed around two guiding questions: }
\textbf{Q1}: \tyyedit{Under what conditions does M-MBTL perform well, and under what conditions does GP-MBTL perform well?}
\textbf{Q2}: \tyyedit{Can M/GP-MBTL always achieve the best of both worlds, or approach optimal performance?}
\ttyedit{We investigate these questions on CMDP experiments, including synthetic data, continuous control, eco-friendly traffic control, and crop management. Appendix \ref{sec:ablation} reports ablation studies to assess how different combinations of algorithms in SD-MBTL affect performance, and Appendix \ref{sec:time_compare} compares the runtime. Appendix \ref{sec:IQM} shows the algorithm selection at each round and the number of training policies required to achieve $\epsilon$-suboptimal.}

To evaluate the performance of our algorithm in higher-dimensional settings, we conducted experiments on 5-dimensional and 7-dimensional synthetic datasets, which demonstrate the generalizability of M/GP-MBTL for multi-dimensional CMDP problems. Detailed results are provided in Appendix \ref{sec:higher_dim}. 
Moreover, to show that the decision-round parameter K in our results is not cherry-picked, we provide curves of the aggregated metrics as K varies in Appendix~\ref{sec:diff_noise_term} and \ref{sec:IQM}. After only a small number of decision rounds, M/GP-MBTL consistently maintains the best overall performance, demonstrating its robustness.

% \textbf{Q1}: \tyyedit{Why do we need structure detection to address CMDP problems?}\cwww{Re-phrase this question to be more objective and less strong. For instance, I don't think we can convey in a single paper that we \textit{need} SD (necessary condition).}
% \cwwinline{To answer these questions, we first describe our experimental setup. We then report the main findings on synthetic data, followed by results on benchmark datasets. Finally, we perform ablation studies in Appendix \ref{sec:ablation} to assess how different combination of algorithms in SD-MBTL affect performance.}
% \begin{enumerate}
%     \item \textbf{Q1}: \tyyedit{Why do we need structure detection to address CMDP problems?}
%     \item \textbf{Q2}: \tyyedit{Under what conditions does M-MBTL perform well, and under what conditions does GP-MBTL perform well?}
%     \item \textbf{Q3}: \tyyedit{Can M/GP-MBTL always achieve the best of both worlds, or at least approach optimal performance?}
%     % \jhfn{Can M/GP-MBTL always achieve the best of both worlds?}
% \end{enumerate}
% \jhfn{Please double check if the experiments section is structured to answer these questions.}
% TODO: alg for each problem structure
% How to solve it, M-MBTL, GP-MBTL
\subsection{Setup}
% To ensure sufficient task density, we discretize each context dimension into multiple values and generate tasks by enumerating all combinations across dimensions.
% This results in an exponential growth in the number of tasks—and hence training time—with increasing dimensionality. 
\tedit{Although our aim is to tackle high-dimensional CMDPs, the number of tasks—and hence the training time—grows exponentially with dimensionality. Therefore, we confine our experiments to three-dimensional settings.}
% While our goal is to address multi-dimensional CMDPs, training time increases exponentially with dimensionality. Thus, we consider three-dimensional context in all experiments.
% three dimensions offer a practical balance between problem complexity and computational feasibility. 
% Extending to higher-dimensional contexts remains an important direction for future work.

\textbf{Baselines.}
We consider two types of baselines: canonical and multi-policy. The canonical baselines include: 
(1) \textbf{Independent training}, which trains separate policies on each \tyyedit{MDP} task; and (2) \textbf{Multi-task training}, where a universal context-conditioned policy is trained for all tasks. 
% \cw{Because it can be ambiguous, make sure we define ``task'' to be a context-MDP, and that we never use it to refer to a CMDP.}
The multi-policy baselines include: (3) \textbf{Random selection}, which selects training tasks uniformly at random; (4) \textbf{\jhedit{GP-MBTL}} \citep{MBTL}; (5) \textbf{\jhedit{\jhedit{M-MBTL}}}; and (6) \textbf{\jhedit{Myopic} Oracle}, 
\tyyedit{an optimal solution for GSTS problem which has access to the generalization performance of each source-target pair $(x,y)$ in all experiment trials and selects the training task myopically in each step.} \cwwinline{More details, including the formal definition and relationship to prior work, are provided in Appendix~\ref{sec:oracle}.}
% \jhfn{revise to our new stochastic oracle definition}
% which has access to the \jhedit{generalization performance} of all source policies to all tasks, and uses that information to greedily select the best source task at each decision round.

% \textbf{Distance metric.}\jhfn{double check if we need this}
% \tyedit{We use the L\jhedit{$_1$} norm distance in \jhedit{M-MBTL}.} This aligns with our experimental observations, where each context dimension influences the generalization gap independently.
%  \begin{equation}
%      \begin{aligned}
%          \Delta J (\pi_{x}, y)\simeq-\text{dist}_w(x,y)=-w^\intercal|x-y|
%      \end{aligned}
%  \end{equation}
% Compared to other distance metrics, such as the L\jhedit{$_2$} norm, the L\jhedit{$_1$} norm treats each dimension independently. This aligns with our experimental observations, where each context dimension influences the generalization gap independently.

\textbf{\jhedit{Train and zero-shot transfer.}}
% We discretize the three-dimensional continuous task set \( X \) into \( N_1\times N_2\times N_3 =N\) tasks, where $N_i$ represents the number of context variations for dimension $i$.
We discretize the continuous task set \( X \) into $N$ tasks. 
For the ease of our problem, we set these discretized tasks \jhedit{the} same as $Y$. 
We use Proximal Policy Optimization (PPO) \citep{schulman2017proximal} for training. We zero-shot transfer the \jhedit{policies} obtained after training each task to all other tasks.
\tedit{The generalization performance constructs a transfer matrix whose rows represent source tasks and columns represent target tasks.}
\tyyedit{We set the number of decision rounds $K=N$, and train three times with different random seeds}.
\Tyedit{More details on deep \jhedit{RL} implementation and hyperparameters are provided in Appendix~\jhedit{\ref{sec:implementation_details}}}.

\textbf{Bootstrapping.} 
\tedit{However, three matrices are not enough for evaluating our different source task selections.
To address this, we employ \textit{bootstrapping} \citep{tibshirani1993introduction} to synthesize an expanded dataset.
Since each task is independently trained three times, the rows of the transfer matrix are independent. 
Therefore,} we generate 100 bootstrapped transfer matrices \Tyedit{with different random seeds} by resampling \tyyedit{rows of the transfer matrices} with replacement from the original set of trained policies. 
\tyyedit{Through bootstrapping, each row of the transfer matrix is still sampled from its true distribution, but the number of transfer matrices is significantly increased.}
\tyedit{\jhedit{It} allows us to more reliably approximate the real distribution of the transfer matrix and thereby mitigate the effects of limited training runs.}
% \cw{Why is bootstrap justified?} 
% \cw{Make sure we record the random seeds for reproducibility.}
% \tyfn{I revised this sentense. The original one "Bootstrapping allows us to approximate the sampling distribution of our performance estimates more reliably and thereby..." seems hard to understand.}
% \jhfn{Please review this.}

\textbf{Performance measure.}
% When evaluating algorithms on the \jhedit{G}STS problem, the training task context output by the algorithm in each round is mapped to the nearest 
% \tyedit{task} in the discrete set for training.
% \jhfn{what do you mean by nearest neighbor here?} 
% neighbor 
% The corresponding row in the transfer matrix is then used as the \jhedit{generalization performance} for evaluation. 
We apply min-max normalization to the performance across all tasks \jhedit{after the algorithm is run}.
We use the mean as the performance metric\tedit{, and use bootstrap resampling to compute the 95\% confidence intervals (CI) of the performance. Because the upper and lower bounds of the CI are roughly symmetric, we only display $\pm\frac{(\text{upper} - \text{lower})}{2}$ as the CI half-width.}
We also evaluated the IQM and the median of the performance, which are presented in Appendix \ref{sec:IQM}. Both of them are close to the mean.
% , suggesting that our conclusions are unlikely to depend strongly on outliers. 
\tedit{Additionally, to compare methods across different benchmarks and cases, we use an \textbf{aggregated performance} metric adapted from the human-normalized score~\cite{agent57, Mnih2015HumanlevelCT}: each algorithm’s score on a benchmark is linearly rescaled so that the Random baseline maps to 0 and the Myopic Oracle to 1, and we then average these normalized scores over all benchmarks. This single 0–1 value shows both how much an algorithm surpasses Random and how close it comes to the oracle; full definitions and formulas appear in Appendix \ref{sec:aggregated_metric}.}

% \tedit{Additionally, on the CMDP benchmarks, we design an \textbf{aggregated performance} metric to evaluate how baseline algorithms perform across multiple benchmarks—particularly suited for MBTL-based methods. For each benchmark, we take an algorithm’s performance, subtract the Random baseline, and divide by the difference between the Myopic Oracle’s performance and the Random baseline. This yields a normalized performance, analogous to min-max normalization. We then average an algorithm’s normalized performance across the four benchmarks to obtain its aggregated performance. This metric scales each MBTL-based algorithm’s score between 0 and 1, indicating how much it outperforms the Random baseline and how closely it approaches the Myopic Oracle.}
% \jhedit{In addition, we use the Interquartile Mean (IQM) as a performance metric that reduces the influence of outliers \citep{agarwal2021deep}.}
% \tyyedit{We are not concerned with these outliers, as they may result from training instability. In some cases, an algorithm might happen to select a training task that generalizes exceptionally well, but such outcomes are unreliable in real-world scenarios.}
% \cw{Do we not care about outliers? Why or why not?}
\subsection{Synthetic Data}
\textbf{Data description.}
We \jhedit{first} evaluate the performance of \jhedit{M/GP}-MBTL under \jhedit{synthesized dataset}, including the cases where Assumptions \ref{assump:constant} and \ref{assump:distance} are satisfied or not\tyyedit{, as well as cases with varying noise levels}. 
We discretize the \tyyedit{three dimensional continuous} task set \( X \) into $N=8\times8\times8$ tasks, each characterized by three context \ttyedit{dimensions}, where each \ttyedit{dimension} is an integer ranging from 1 to 8. We generate $J(\pi_x,y)=f(x)+g(y)+h(x,y)+C \Tyedit{+\epsilon_{x,y}}$
\ttyedit{, where $C=500$ is a constant that does not affect any algorithmic decisions, $\epsilon_{x,y}\sim \mathcal{N}(0,\sigma^2)$ is a Gaussian noise sampled independently for all $x$ and $y$, and $\sigma=5$.}
% \cwww{The connection $512 = 8^3$ could be clearer. It's totally non-obvious that you can discretize 512 tasks in this way. I would start by specifying 3 context dimensions, not start with 512 tasks.}
% \cwww{I need more information on the ranges of $f(\cdot), g(\cdot), h(\cdot, \cdot)$. My interpretation so far is that noise is at roughly a 5\% level (25/500).}
% \cwww{Do you mean $\epsilon$ (fixed) or $\epsilon_{xy}$? Write $\sigma$ instead of 5 and then set the parameter. Check all notation...}
We include \ttyedit{eight} conditions, which is the combination of \ttyedit{three} variations: 1) constant and non-constant $f(x)$; 2)  constant and non-constant $g(y)$; 3) distance metric holds or \jhedit{does not hold}.
% \cwww{For numbers under ten, spell out the number, i.e., eight.}
% \tyyedit{In the non-constant setting, we model $f(x)$ and $g(y)$ as linear functions. When the distance metric holds, we use $L_1$ norm distance; when they are violated, we flip the slope signs in two of the three dimensions to break metric consistency.\cwww{Not clear what this means.} Additional implementation details appear in Appendix \ref{sec:synthetic-data}.}
\ttyedit{We use a linear function $f(x)=[4,4,4]\cdot x$ and $g(y)=[3,3,3]\cdot y$ for the non-constant case, and $f(x)=0$, $g(y)=0$ for constant case.
When the distance metric holds, $h(x,y)=-[3,3,3]\cdot |x-y|$, where $|\cdot|$ means the absolute value for each element in a vector. Otherwise, $h(x,y)=-([3,3,3]\cdot [x-y]_+ + [1, 1, -3]\cdot[x-y]_-)$.}
\Tyedit{We also tested our approaches on the synthetic data with different noise \jhedit{levels} in Appendix~\ref{sec:diff_noise_term}.}
% \cw{Plot these transfer matrices (or other visuals) and include them in Supplementary.}
% \jhfn{is it common to use $|\cdot|$ for vector norm?}
% \jhfn{did you define $[\cdot]_+$ in vector? or scalar?}.

\begin{table*}[!ht]
\begin{small}
\begin{center}
  \begin{tabular}{lllcccc|c}
    \toprule
    \textbf{$f(x)$} & \textbf{$g(y)$} & \textbf{$h(x,y)$}  & \textbf{Random} & \textbf{GP-MBTL} & \textbf{M-MBTL (Ours)} & \textbf{M/GP-MBTL (Ours)} & \textbf{Myopic Oracle}\\
    \midrule
    Constant & Linear & Non-distance & {0.6956} $\pm$ 0.0009 & \textbf{0.7212} $\pm$ 0.0006 & {0.6976} $\pm$ 0.0002 & {0.7168} $\pm$ 0.0010 & {0.7260} $\pm$ 0.0001\\
    \rowcolor{mountainrow} Constant & Linear & $L_1$ norm & {0.7011} $\pm$ 0.0002 & {0.6961} $\pm$ 0.0011 & \textbf{0.7038} $\pm$ 0.0002 & \textbf{0.7038} $\pm$ 0.0002 & {0.7048} $\pm$ 0.0002\\
    Constant & None & Non-distance & {0.7906} $\pm$ 0.0014 & \textbf{0.8369} $\pm$ 0.0003 & {0.7936} $\pm$ 0.0003 & {0.8334} $\pm$ 0.0004 & {0.8375} $\pm$ 0.0002\\
    \rowcolor{mountainrow} Constant & None & $L_1$ norm & {0.7722} $\pm$ 0.0004 & {0.7747} $\pm$ 0.0005 & \textbf{0.7762} $\pm$ 0.0003 & \textbf{0.7762} $\pm$ 0.0003 & {0.7778} $\pm$ 0.0003\\
    Linear & Linear & Non-distance & {0.5773} $\pm$ 0.0015 & \textbf{0.6088} $\pm$ 0.0001 & {0.5782} $\pm$ 0.0001 & \textbf{0.6086} $\pm$ 0.0001 & {0.6090} $\pm$ 0.0001\\
    Linear & Linear & $L_1$ norm & {0.5880} $\pm$ 0.0020 & \textbf{0.6218} $\pm$ 0.0001 & {0.5900} $\pm$ 0.0002 & \textbf{0.6216} $\pm$ 0.0002 & {0.6221} $\pm$ 0.0001\\
    Linear & None & Non-distance & {0.6672} $\pm$ 0.0020 & \textbf{0.7083} $\pm$ 0.0002 & {0.6683} $\pm$ 0.0002 & {0.7073} $\pm$ 0.0006 & {0.7088} $\pm$ 0.0002\\
    Linear & None & $L_1$ norm & {0.7401} $\pm$ 0.0021 & \textbf{0.7744} $\pm$ 0.0012 & {0.7422} $\pm$ 0.0002 & \textbf{0.7751} $\pm$ 0.0002 & {0.7756} $\pm$ 0.0001\\
\midrule
\multicolumn{3}{c}{\textbf{Aggregated Performance}} & {-0.0000} $\pm$ 0.0107 & {0.2127} $\pm$ 0.0511 & {0.0147} $\pm$ 0.0089 & \textbf{0.3099} $\pm$ 0.0290 & {1.0000} $\pm$ 0.0088 \\
    \bottomrule
  \end{tabular}
  \end{center}
  \caption{Performance comparison on Synthetic Data ($\epsilon=\mathcal{N}(0, 5^2)$) with $K=50$. Values are reported as the mean performance ± half the width of the 95\% confidence interval. Rows shaded in gray indicate the settings that satisfy our proposed \textsc{Mountain} structure. Bold values represent the highest value(s) within the statistically significant range for each task, excluding the oracle.}
  \label{tab:synthetic-performance}
  % \scriptsize{* \textit{Note}: Values are reported as the mean performance ± half the width of the 95\% confidence interval. Rows shaded in light gray indicate the synthetic settings that satisfy our proposed \textsc{Mountain} structure assumption. Bold values represent the highest value(s) within the statistically significant range for each task, excluding the oracle.}
  \end{small}
\end{table*}

\textbf{Results.} 
Table~\ref{tab:synthetic-performance} presents the performance comparison on synthetic data.  
 \tyyedit{Rows shaded in light gray indicate the synthetic settings that satisfy our proposed \textsc{Mountain} structure assumption.}
\tyyedit{\textbf{Answer to Q1}: When \textsc{Mountain} structure holds, M-MBTL achieves the best performance among multi-policy baselines and closely approaches the performance of the myopic oracle. This indicates that M-MBTL} effectively exploits this structure to get near-optimal performance with reduced training cost. 
Conversely, when \textsc{Mountain} structure is violated, GP-MBTL achieves the highest performance, demonstrating a general ability to handle CMDP problems.
\tyyedit{\textbf{Answer to Q2}: \jhedit{Overall}, M/GP-MBTL attains the best \tedit{aggregated performance}, demonstrating that it can successfully detect whether a CMDP satisfies \textsc{Mountain} structure and choose the appropriate algorithm accordingly.} 

\subsection{\jhedit{CMDP} Benchmarks}
\jhedit{\textbf{Benchmarks.}}
\jhedit{We consider four CMDP benchmarks: CartPole, BipedalWalker, IntersectionZoo, and \jhedit{CyclesGym}.}
% \textbf{Continuous Control Benchmarks.} 
In CartPole, the objective is to keep a pole balanced \jhedit{upright} on a moving cart. We vary three key context variables---length of the pole, mass of the pole, and mass of the cart---in ranges spanning from 0.2 to 2 times their standard values \jhedit{($N=9\times10\times10$)}.
BipedalWalker is a 4-joint walker robot environment. Context variables are friction, gravity, and scale, ranging from 0.2 to 1.6 times their default values ($N=8\times8\times8$). 
We use CARL \jhedit{environments} \citep{benjamins2023contextualizecasecontext} \jhedit{for} these CartPole and BipedalWalker.
% \textbf{Traffic Benchmark.}
IntersectionZoo \citep{jayawardana2025intersectionzoo} is \jhedit{a multi-agent CRL benchmark for eco-driving in urban road networks}.
We run experiments on a synthetic intersection network configuration with three \ttyedit{context variables}---traffic inflow, autonomous vehicle (AV) penetration rate, and traffic signal green-phase duration---from 0.2 to 1.2 times their default values \jhedit{($N=6\times6\times6$)}\jhedit{, providing} a broad range of \jhedit{complex} traffic scenarios \jhedit{for evaluation}.
CyclesGym \citep{turchetta2022learning} is an agricultural management simulation environment in which an RL agent is tasked with managing various crop-related parameters.
Context variables are temperature, sunlight, and precipitation. These variables are discretized into a \(6 \times 6 \times 6\) grid, yielding different agricultural scenarios.
\Tyedit{More details on CMDP benchmarks are provided in Appendix \ref{sec:benchmark_details}.}

\begin{table*}[!ht]
    \begin{small}
\begin{center}
  % \caption{Performance comparison of different methods on CMDP benchmarks}
  % \label{tab:performance}
  \begin{tabular}{ccccccc|c}
    \toprule
    \textbf{Benchmark (CMDP)} & \textbf{Independent} & \textbf{Multi-task} & \textbf{Random} & \textbf{GP-MBTL} & \textbf{M-MBTL (Ours)} & \textbf{M/GP-MBTL (Ours)} & \textbf{Myopic Oracle} \\
    \midrule
\begin{tabular}[c]{@{}c@{}}\textbf{CartPole}\\($K=12$)\end{tabular} & \begin{tabular}[c]{@{}c@{}}0.9346\\$\pm$ 0.0003\end{tabular} & \begin{tabular}[c]{@{}c@{}}\textbf{0.9967}\\$\pm$ 0.0024\end{tabular} & \begin{tabular}[c]{@{}c@{}}0.9861\\$\pm$ 0.0017\end{tabular} & \begin{tabular}[c]{@{}c@{}}0.9919\\$\pm$ 0.0013\end{tabular} & \begin{tabular}[c]{@{}c@{}}0.9896\\$\pm$ 0.0016\end{tabular} & \begin{tabular}[c]{@{}c@{}}0.9898\\$\pm$ 0.0016\end{tabular} & \begin{tabular}[c]{@{}c@{}}0.9998\\$\pm$ 0.0000\end{tabular} \\
\begin{tabular}[c]{@{}c@{}}\textbf{BipedalWalker}\\($K=12$)\end{tabular} & \begin{tabular}[c]{@{}c@{}}0.7794\\$\pm$ 0.0011\end{tabular} & \begin{tabular}[c]{@{}c@{}}0.5680\\$\pm$ 0.0919\end{tabular} & \begin{tabular}[c]{@{}c@{}}0.8051\\$\pm$ 0.0045\end{tabular} & \begin{tabular}[c]{@{}c@{}}0.8073\\$\pm$ 0.0044\end{tabular} & \begin{tabular}[c]{@{}c@{}}\textbf{0.8315}\\$\pm$ 0.0029\end{tabular} & \begin{tabular}[c]{@{}c@{}}\textbf{0.8261}\\$\pm$ 0.0030\end{tabular} & \begin{tabular}[c]{@{}c@{}}0.8629\\$\pm$ 0.0011\end{tabular} \\
\begin{tabular}[c]{@{}c@{}}\textbf{IntersectionZoo}\\($K=50$)\end{tabular} & \begin{tabular}[c]{@{}c@{}}0.2045\\$\pm$ 0.0008\end{tabular} & \begin{tabular}[c]{@{}c@{}}0.3788\\$\pm$ 0.1059\end{tabular} & \begin{tabular}[c]{@{}c@{}}0.5288\\$\pm$ 0.0108\end{tabular} & \begin{tabular}[c]{@{}c@{}}\textbf{0.5840}\\$\pm$ 0.0092\end{tabular} & \begin{tabular}[c]{@{}c@{}}0.4878\\$\pm$ 0.0064\end{tabular} & \begin{tabular}[c]{@{}c@{}}\textbf{0.5682}\\$\pm$ 0.0069\end{tabular} & \begin{tabular}[c]{@{}c@{}}0.6305\\$\pm$ 0.0082\end{tabular} \\
\begin{tabular}[c]{@{}c@{}}\textbf{CyclesGym}\\($K=50$)\end{tabular} & \begin{tabular}[c]{@{}c@{}}0.2133\\$\pm$ 0.0002\end{tabular} & \begin{tabular}[c]{@{}c@{}}0.2081\\$\pm$ 0.0002\end{tabular} & \begin{tabular}[c]{@{}c@{}}0.2198\\$\pm$ 0.0001\end{tabular} & \begin{tabular}[c]{@{}c@{}}0.2193\\$\pm$ 0.0001\end{tabular} & \begin{tabular}[c]{@{}c@{}}\textbf{0.2205}\\$\pm$ 0.0001\end{tabular} & \begin{tabular}[c]{@{}c@{}}0.2201\\$\pm$ 0.0001\end{tabular} & \begin{tabular}[c]{@{}c@{}}0.2214\\$\pm$ 0.0001\end{tabular} \\
\midrule
\begin{tabular}[c]{@{}c@{}}\textbf{Aggregated}\\\textbf{Performance}\end{tabular} & \begin{tabular}[c]{@{}c@{}}-2.9063\\$\pm$ 0.1470\end{tabular} & \begin{tabular}[c]{@{}c@{}}-3.1120\\$\pm$ 1.8408\end{tabular} & \begin{tabular}[c]{@{}c@{}}0.0000\\$\pm$ 0.0467\end{tabular} & \begin{tabular}[c]{@{}c@{}}0.1681\\$\pm$ 0.0539\end{tabular} & \begin{tabular}[c]{@{}c@{}}0.1822\\$\pm$ 0.0503\end{tabular} & \begin{tabular}[c]{@{}c@{}}\textbf{0.2930}\\$\pm$ 0.0403\end{tabular} & \begin{tabular}[c]{@{}c@{}}1.0000\\$\pm$ 0.0236\end{tabular} \\
\bottomrule
    \end{tabular}
\end{center}
  \caption{Performance comparison of different methods on CMDP benchmarks. Values are reported as the mean performance ± half the width of the 95\% confidence interval. Bold values represent the highest value(s) within the statistically significant range for each task, excluding the oracle.}
  \label{tab:performance}\end{small}
\end{table*}

\textbf{Results.}
Table~\ref{tab:performance} summarizes the results across four benchmarks.
\tyyedit{\textbf{Answer to Q1}:} 
On the whole, \cwwinline{the gap between the Myopic Oracle and} canonical baselines (independent and multi-task training) \cwwinline{indicate} the potential of \ttyedit{MBTL}\cwwinline{-based} strategic task selection.
% \cwww{Be more direct!} 
\ttyedit{Multi-task training performs the best on CartPole, which may be because the problem is relatively easy, such that a single straightforward context-conditioned policy can achieve strong performance \jhedit{across the} CMDP. However, it performs worse in other benchmarks, potentially due to increased task complexity, where it may suffer from model capacity limitations.}
\tyedit{By contrast, M-MBTL and GP-MBTL show better aggregated metric performance than independent training and multi-task training.}
% Multi-task training outperforms baselines only on the simplest task (CartPole) and degrades on the other three benchmarks.\cwww{Double check this.} 
% By contrast, MBTL-based algorithms exhibit consistently robust\cwww{What do we mean by robust? Did we use this language earlier? Try to be consistent with the language.} results overall. 
% However, GP-MBTL and M-MBTL each attain optimal performance only under specific conditions\cwww{Re-phrase. I don't think we have enough evidence with 4 tasks to say that ``only under specific conditions''.}, revealing substantial room for structure-aware task selection.
\ttyedit{In tasks that satisfy \textsc{Mountain} structure (BipedalWalker, \jhedit{CyclesGym}), M-MBTL achieves the highest performance. }
% Tasks that satisfy Mountain structure (BipedalWalker, \jhedit{CyclesGym}) favour M-MBTL, which rapidly approaches the Myopic Oracle. 
When the structure is violated (IntersectionZoo), GP-MBTL becomes superior, showing the value of a GP model for source-task performance.
% \tyyedit{In CMDPs with \textsc{Mountain} structure, such as BipedalWalker, and Crop Management, M-MBTL rapidly approaches Myopic Oracle and shows the the best performance. However, in the IntersectionZoo benchmark, which does not exhibit the \textsc{Mountain} structure, GP-MBTL outperforms all baselines. This suggests that modeling the source task performance with GP is essential when \textsc{Mountain} structure is not detected.
In addition, CartPole satisfies the assumptions of both M-MBTL and GP-MBTL, so both algorithms perform well. 
\tyyedit{\textbf{Answer to Q2}:} 
% Across all four benchmarks, M/GP-MBTL performs almost as well as the better of M-MBTL and GP-MBTL, indicating its effectiveness for solving CMDP.
\tedit{Across all four benchmarks, M/GP-MBTL consistently matches the stronger of M-MBTL and GP-MBTL. It achieves the best aggregated performance and shows an improvement of 0.1249 over GP-MBTL, indicating that it moves 12.49\% closer to the Myopic Oracle within the MBTL framework. This highlights the method’s effectiveness in addressing various CMDP problems.}
% \ttyedit{Across all four benchmarks, M/GP-MBTL consistently matches the stronger of M-MBTL and GP-MBTL. Its gap to the myopic oracle stays below 5\% at $K=12$ on the classic-control tasks (CartPole and BipedalWalker) and below 10\% at $K=50$ on IntersectionZoo and CyclesGym, underscoring the method’s effectiveness on CMDPs.}

% \jhedit{However, \cwwinline{M/GP-MBTL} achieve close to GP-MBTL, indicating} that it correctly recognizes the violation of the \textsc{Mountain} structure and selects GP-MBTL as the appropriate task selection strategy for the problem.  
% Multi-task training performs the best in the long run on CartPole, \cwwinline{which may be because} the problem is relatively \cwwinline{easy, such that} a single \cwwinline{straightforward context-conditioned} policy can achieve strong performance \jhedit{across \cwwinline{the} CMDP}. However, it performs worse in other benchmarks, \cwwinline{potentially} due to increased task complexity, where it \cwwinline{may} suffers from model capacity limitations.\cw{Qualify your statements when you don't have strong evidence!}
\section{Related Works}\label{sec:related-works}
\jhedit{\textbf{Contextual Reinforcement Learning \jhhedit{and Multi-Policy Approaches}.} CRL provides a framework for managing variations of MDP in environment dynamics, rewards, and initial states through a context variable, effectively yielding a family of related tasks parameterized by these contexts \citep{hallak2015contextual, modi2018markov, benjamins2023contextualizecasecontext}. When the context is observed, a popular approach is to train a \jhedit{single} policy that explicitly conditions on this contextual information \citep{teh2017distral,sodhani2021multi} \jhedit{or by encoding it with a latent representation \citep{yu2020meta, sun_paco_2022}.}}
\jhedit{Although performance can be strong when the tasks are homogeneous, negative transfer and capacity limits arise when context variation is large.} 
\jhhedit{An alternative is to train a small set of expert policies. This idea has been explored through policy committees \citep{ge2025learning} and represented-MDPs \citep{ivanov2024personalized}, where the primary goal is to cluster the task space upfront and train one expert policy per cluster.}
% Our work tackles this limitation by multi-policy training, which selects a set of source tasks to train individually, then transferring the resulting policies zero-shot to the remaining tasks.
% \cw{Add relation to our work.}
% \jhedit{\textbf{Multi-policy Training.} 
\jhedit{Recent work shows that training only a few carefully chosen source tasks can outperform both independent training and multi-task training} \citep{cho2023temporaltransferlearningtraffic, MBTL}. 
% \jhedit{GP-MBTL \cite{MBTL} formulates selection as Bayesian optimization in one-dimensional context spaces.} 
\jhedit{We generalize these ideas in multi-dimensional contexts and a \textit{meta} layer that first detects what generalization structure is present before choosing which task-selection to deploy.}
% \cwww{My gut instinct is that this is not the related work to be discussing in this section, in part because you have already discussed it in the intro. Or like, maybe this can be compressed to just 1 sentence.} 
% \cwww{Just brainstorming. I'd like to see a section discussing cross-task generalization in ML/RL, e.g., what approaches people have taken to identify or exploit structure.}
% MBTL tackles this problem by modeling both training performance and generalization gaps with Bayesian optimization to adaptively pick new source tasks. MBTL has been shown to be effective in one-dimensional contextual domains but faces scalability challenges as the dimensionality of the context increases \citep{MBTL}.}\cw{Add relation to our work. I'm also not convinced we need a whole section on this. Also, no need to reiterate if we have already conveyed this earlier in the paper; a 1-sentence summary would suffice. Maybe broaden it to transfer learning in CRL?}

\jhedit{\textbf{Structure Detection.}} \cwwinline{Structure detection appears widely in machine learning, for example in \jhedit{detecting convexity for optimization problems \cite{ahmadi2013np} and discovering GP kernels for nonparametric regression \cite{duvenaud2013structure}}. Typical frameworks include \jhedit{finding good approximations or decomposing the unknown functions}. However, structure detection is fairly nascent in the context of CRL. In CRL, structure detection appears \jhedit{appears as (i) \emph{context change detection}, where hierarchical RL agents infer when environment dynamics have shifted and switch options accordingly \citep{yucesoy_hierarchical_2015}; (ii) \emph{model selection}, where online statistical tests reject over-complex dynamics models to curb over-fitting \citep{lee2021online}; and (iii) \emph{policy-reuse selection}, where an agent \jhhedit{leverages a pre-existing library of policies.}
\jhhedit{These methods decide whether a source policy can be safely transferred to a new target task \citep{fernandez2006probabilistic} or combine multiple source policies to solve a target task, for example, through a mixture-of-experts model \citep{gimelfarb2021contextual}.}
\jhhedit{Our work also detects} structure online, but we leverage it to choose a source-task, rather than to re-weight or gate a fixed library of policies; in principle, both ideas can be combined.}}

\ttyedit{Appendix \ref{sec:related_works} surveys additional literature on clustering methods and solutions for CRL.}
% Structure Detection in CRL could be seen as algorithm selection or hierarchical RL. \cite{yucesoy_hierarchical_2015} integrates hierarchical RL with context detection to quickly adapt to non-stationary environments by inferring shifts in dynamics. \cite{lee2021online} proposes an online meta-algorithm for model selection that rejects overly complex models through simple statistical tests. Our approach employs a similar idea to use simple criteria for detecting the structure.\cw{I don't get the relation to our work, may need more context.} \cw{I wrote a sketch of a revised structure.} \cw{Other related work that appears to be missing: Detection of task similarity, detection of conflicting gradients to avoid negative transfer (the paper I shared recently from Peter Stone's group is an example), detection for policy reuse, etc.}

\section{Conclusion}\label{sec:conclusion}
We propose a structure detection framework for CRL that infers the underlying structure of CMDPs and adaptively selects the task selection strategy. Our M/GP-MBTL algorithm switches between a clustering strategy for \textsc{Mountain} cases and a GP-based strategy otherwise. Experiments on synthetic and real benchmarks show consistent improvements over prior methods, highlighting the promise of structure detection for scalable and robust transfer learning in complex CRL environments.
% In this work, we introduced a novel structure detection framework for MBTL (SD-MBTL) in CRL, which automatically infers the underlying structure of CMDPs and adaptively selects the most appropriate task selection strategy. In particular, our M/GP-MBTL algorithm dynamically switches between a clustering-based approach tailored for \textsc{Mountain} structure and a Gaussian Process–based method for more general settings. Extensive experiments on synthetic datasets and diverse real-world benchmarks demonstrate that our approach achieves consistent improvements in generalization performance over existing methods\jhedit{, highlighting the promise of structure detection for scalable and robust transfer learning in complex CRL environments.}
Nevertheless, our study has several \textbf{limitations}. First, the current detector focuses on a single structure; 
% Second, structure inference relies on an initial set of transfer evaluations, which can be costly when the context space is very large or when policies are expensive to evaluate;
Second, structure inference relies on \ttyedit{transfer evaluations of training tasks}, which can be costly when the context space is very large or when policies are expensive to evaluate;
% \cwww{Algorithm 1 does not indicate that there is an initial set of transfer evaluations.}
Third, all experiments were conducted in three-dimensional context spaces; additional work is needed to confirm scalability to higher-dimensional settings.
Future work will explore \jhedit{richer} CMDP structures and develop specialized algorithms \jhedit{to reduce the number of required source-task trainings while maintaining (or improving) generalization performance.}
% \jhfn{We should mention \textbf{limitation} of our work.}

%%%%%%%%%%%%%%%%%%%%%%%%%%%%%%%%%%%%%%%%%%%%%%%%%%%%%%%%%%%%%%%%
%% Acknowledgement
%%%%%%%%%%%%%%%%%%%%%%%%%%%%%%%%%%%%%%%%%%%%%%%%%%%%%%%%%%%%%%%%
\section*{Acknowledgements}
This work was supported by the National Science Foundation (NSF) CAREER award (\#2239566) and the Kwanjeong Educational Foundation Ph.D. scholarship program. The authors would like to thank the anonymous reviewers for their valuable feedback.

% \jhfn{Check if the bibtex is up-to-date.}
{
\small
\bibliography{reference}
\medskip
}

%%%%%%%%%%%%%%%%%%%%%%%%%%%%%%%%%%%%%%%%%%%%%%%%%%%%%%%%%%%%%%%%
%% Appendices
%%%%%%%%%%%%%%%%%%%%%%%%%%%%%%%%%%%%%%%%%%%%%%%%%%%%%%%%%%%%%%%%
\clearpage
\appendix
\section*{Appendix}
\section{\jhedit{Additional Related Works}}\label{sec:related_works}

\tyedit{\textbf{Clustering.} 
Clustering is a fundamental technique for grouping unlabeled data by similarity \citep{macqueen1967some,dempster1977maximum,kmeans++}. Hierarchical clustering extends this principle by splitting or merging clusters dynamically, allowing a flexible number of centroids \citep{hierarchical_clus_1977, szekely2005hierarchical, hierarchical_clustering_2016}. 
% Building on these ideas, 
\jhedit{M-MBTL} further adapts clustering to an iterative setting: it incrementally adds new centroids while preserving previously learned ones, providing an efficient way to guide the search for training tasks in the \jhedit{G}STS problem.}

\jhedit{\textbf{Multi-task Training.} Multi-task RL seeks to learn a single policy (or a set of shared parameters) to solve \textit{multiple related tasks} \citep{teh2017distral,sodhani2021multi}. While multi-task methods such as PaCo \citep{sun_paco_2022} and MOORE \citep{hendawy_multi_2023} have demonstrated success in discrete and continuous domains, they often assume a fixed task set and can suffer from negative transfer when certain tasks are too dissimilar \citep{kang_learning_2011, standley_which_2020}.}

\jhedit{\textbf{Curriculum \jhhedit{and Self-Paced} Learning.} 
Task‑selection for multi-policy training is closely related to curriculum learning for RL, where an agent is exposed to an automatically ordered sequence of source tasks \cite{narvekar2020curriculum}. 
\jhhedit{Another related approach called self-paced learning selects only a few representative task instances to train on \citep{klink2020self, eimer2021self}.} 
Whereas curriculum methods aim to ease online learning of a \emph{single} policy, our goal is to minimi\jhedit{z}e the number of expensive training runs by deciding \emph{which} contexts to fully train on for zero‑shot transfer.}
\jhhedit{In addition, curriculum learning or self-paced learning typically aims to accelerate the online training of a single universal policy, while our framework focuses on minimizing the total, often computationally prohibitive, cost of training by selecting an optimal subset of source tasks for training multiple expert policies that provide broad zero-shot coverage.}

\jhedit{\textbf{Unsupervised Environment Design (UED).}
UED methods automatically generate training tasks whose difficulty adapts to the current agent, producing curricula without hand-crafted ordering rules. ACCEL \citep{dennis2020emergent} formulates environment creation as an RL problem whose reward is tasks on the brink of the agent’s competence. These works complement ours: they invent new source tasks, whereas we select from a fixed CMDP. Our structure detection could, in principle, guide a UED generator---e.g. to bias sampling toward contexts that satisfy a detected \textsc{Mountain} structure.}

\jhedit{\textbf{Meta-learning.}
Meta-RL algorithms such as MAML \citep{finn2017model} learn policies that adapt rapidly to a \emph{single} new task via a few gradient steps or a short context window. Our method differs in motivation: we want \emph{zero-shot} coverage of \textit{all} tasks in a CMDP while minimizing the number of full training runs. Nevertheless, meta-learning could boost each individual source-task training phase, and conversely our structure detector could decide \emph{when} to switch from meta-learning to plain fine-tuning as context variance grows.}

\section{\tyyedit{Notation Table}}
Table \ref{tab:notation} describes the notation used in this paper.
\begin{table}[htbp]
    \centering
    \caption{Notation Table}
    \label{tab:notation}
    \begin{tabular}{|c|c|}
    \hline
        \textbf{Symbol} & \textbf{Description} \\
        \hline
        $x$  & Source task ($x\in X$) \\
        $y$      & Target task ($y\in Y$)  \\
        $\pi_x$      & Trained policy from source task  \\
        $J(\pi_x,y)$ & Generalization Performance of $\pi_x$ on $y$\\
        $x_{1:K}$ & Selected source tasks $x_1,x_2,..,x_K$ \\
        $f(x)$ & Policy quality of source task $x$ \\
        $g(y)$ & Difficulty of target task $y$ \\
        $h(x,y)$ & Task dissimilarity between $x$ and $y$ \\
        $\text{dist}(x,y)$ & A distance metric \\
        $\overline{J}(\pi_x,y)$ & Relative generalization performance of $\pi_x$ on $y$\\
        \hline
    \end{tabular}
\end{table}

\section{\tedit{Details of the Generalization-Performance Structure Decomposition}}\label{sec:decomposition}

\tedit{The Sobol–Hoeffding (functional-ANOVA) decomposition provides a \emph{unique} and \emph{orthogonal} expansion of any square-integrable function of \emph{independent} inputs into a grand mean, additive main-effect terms, and a residual interaction term.  
We restate the result for the two-variable case, which underpins our generalization-performance model.}

\begin{theorem}[Sobol–Hoeffding decomposition for two variables~\cite{hoffding1948class,sobol1990sensitivity}]
\label{theorem:sobol}
Let \((X,Y)\) be independent random variables with supports \(\mathcal X\) and \(\mathcal Y\), and let \(F\in L^{2}(\mathcal X\times\mathcal Y)\).  
Define the global mean
\[
\mu \;=\; \mathbb{E}_{X,Y}\!\bigl[F(X,Y)\bigr].
\]
Then there exist unique functions
\(
F_X\in L^{2}(\mathcal X),\;
F_Y\in L^{2}(\mathcal Y),\;
F_{XY}\in L^{2}(\mathcal X\times\mathcal Y)
\)
such that
\[
F(x,y)\;=\;\mu + F_X(x) + F_Y(y) + F_{XY}(x,y),
\]
with the orthogonality and zero-mean conditions
\[
\mathbb{E}_X[F_X(X)]=\mathbb{E}_Y[F_Y(Y)]=\mathbb{E}_{X,Y}[F_{XY}(X,Y)]=0.
\]
Explicitly,
\begin{equation}
\begin{aligned}
F_X(x)      &= \mathbb{E}_Y\!\bigl[F(x,Y)\bigr]-\mu,\\
F_Y(y)      &= \mathbb{E}_X\!\bigl[F(X,y)\bigr]-\mu,\\
F_{XY}(x,y) &= F(x,y)-F_X(x)-F_Y(y)-\mu.
\end{aligned}
\end{equation}
\end{theorem}

\tedit{\paragraph{Connection to Definition~\ref{def:general_model}.}
Let \(J(\pi_x,y)\) denote the generalization-performance function and identify \(F(x,y)=J(\pi_x,y)\).
Set \(g(y):=F_Y(y)\) and introduce 
\begin{equation}
    \begin{aligned}
        f(x)\;:=&\;F_X(x)-F_{XY}(x,x), \\
        h(x,y)\;:=&\;J(\pi_x,y)-f(x)-g(y)-\mu.
    \end{aligned}
\end{equation}
\tedit{This construction yields exactly the decomposition stated in Definition~\ref{def:general_model}.} It also leads to a difference between the Sobol-Hoeffding decomposition and our definition. Because the diagonal term \(F_{XY}(x,x)\) is subtracted from \(F_X(x)\), the interaction term is normalized so that \(h(x,y)=0\) if $x=y$. However, $\mathbb{E}_x[f(x)]=0$ is not satisfied.}

\section{\tyyedit{Proof of Equivalence in Definition \ref{def:mountain}}}\label{sec:equivalence}
\begin{proof}
The equivalent relationship is because, since $f(x)=C_1$ and $h(x,y)=-\text{dist}(x,y)$, we can get $J(\pi_y,y)=g(y)+C_1+C$. Therefore, $J(\pi_x, y)=C_1+C+(J(\pi_y,y)-C_1-C)-\text{dist}(x,y)=J(\pi_y,y)-\text{dist}(x,y)$.
\end{proof}

\section{\tyyedit{Proof of Lemma \ref{lemma:compare_std}}}\label{sec:proof-compare_std}
\begin{proof}
\ttyedit{Since $h(x,y)=0$ if $x=y$, we can get $\overline{J}(\pi_x,x)=f(x)$. Therefore, $\text{std}_{x\in x_{1:k}}(\overline{J}(\pi_x,x))=\text{std}_{x\in x_{1:k}}(f(x))$, and $\text{std}_{y\in Y}(\overline{J}(\pi_x,y))=\text{std}_{y\in Y} (h(x,y))$. 
Thus, if $\text{std}_{x\in x_{1:k}}(\overline{J}(\pi_x,x))<\mathbb{E}_{x\in x_{1:k}}[\text{std}_{y\in Y}(\overline{J}(\pi_x,y))]$, we have $\text{std}_{x\in x_{1:k}}(f(x))<\mathbb{E}_{x\in x_{1:k}}[\text{std}_{y\in Y} (h(x,y))]$.}
\end{proof}
% \begin{proof}
% \tyyedit{If Assumption \ref{assump:constant} is violated, we can get $\overline{J}(\pi_x,x)=f(x)+C$, and $\overline{J}(\pi_x,y)=f(x)-\text{dist}(x,y)+C$. Therefore, $\text{std}_{x\in x_{1:k}}(\overline{J}(\pi_x,x))=\text{std}_{x\in x_{1:k}}(f(x))$, and $\text{std}_{y\in Y}(\overline{J}(\pi_x,y))=\text{std}_{y\in Y} (\text{dist}(x,y))$. 
% Thus, if $\text{std}_{x\in x_{1:k}}(\overline{J}(\pi_x,x))<\mathbb{E}_{x\in x_{1:k}}[\text{std}_{y\in Y}(\overline{J}(\pi_x,y))]$, we have $\text{std}_{x\in x_{1:k}}(f(x))<\mathbb{E}_{x\in x_{1:k}}[\text{std}_{y\in Y} (\text{dist}(x,y))]$.}
% \end{proof}

\section{\tyedit{Proof of Lemma \ref{lemma:reduce}}}\label{sec:proof-reduction}
Lemma \ref{lemma:reduce} shows that if Mountain structure holds, we have $J(\pi_x,y)=J(\pi_y,y)-\text{dist}(x,y)$. The \jhedit{G}STS problem can be reduced to a sequential version of clustering. Here is a proof.
\begin{proof}
    \begin{equation}
    \begin{aligned}
        % &\max_{x_k}\mathbb{E}_{y_n\tyyedit{\in\mathcal{U}(Y)}}[\max(\Tyedit{J}_{k-1}(y_n), J(\pi_{x_k}, y_n)]\\
        % &\quad=\max_{x_k}\mathbb{E}_{y_n\tyyedit{\in\mathcal{U}(Y)}}[\max_{\kappa\in[k]}J(\pi_{x_\kappa}, y_n)]\\
        &\max_{x_k}\mathbb{E}_{y_n\tyyedit{\in\mathcal{U}(Y)}}[\max_{\kappa\in[k]}J(\pi_{x_\kappa}, y_n)]\\
        =&\max_{x_k}\mathbb{E}_{y_n\tyyedit{\in\mathcal{U}(Y)}}[\max_{\kappa\in[k]} J(\pi_{\Tyedit{y_n}}, \Tyedit{y_n})-\text{dist}(x_\kappa, y_n)]\\
         \tyedit{\Leftrightarrow}&-\min_{x_k}\mathbb{E}_{y_n\tyyedit{\in\mathcal{U}(Y)}}[\min_{\kappa\in[k]}\text{dist} (x_\kappa, y_n)]
    \end{aligned}
\end{equation}
\end{proof}
\section{\ztyedit{Details in Slope Criterion}}\label{sec:slope_criterion}
\ztyedit{Below, we provide more details on calculating the slopes in Slope Criterion. First, we concatenate the context difference as $\text{diff}_{\kappa, n}=[[x_\kappa-y_{\Tyedit{n}}]_-, [x_\kappa-y_{\Tyedit{n}}]_+], \forall \kappa,n,$. Here we slightly abuse the notation that \([ \cdot ]_{+}\) and \([ \cdot ]_{-}\) represent \(\max(\cdot, 0)\) and \(\min(\cdot, 0)\) \tyedit{for each element in the vector,} respectively. 
Using linear regression on the pairs \(\{ \overline{J}(\tyedit{\pi_{x_\kappa}}, y_n) \}_{n\in[N], \kappa\in[k]}\) versus \(\{ \text{diff}_{\kappa, n} \}_{n\in[N], \kappa\in[k]}\), we obtain slopes from the left and right sides: \(\theta_{\text{left}}\in \mathbb{R}^D\) and \(\theta_{\text{right}}\in \mathbb{R}^D\).}
\section{M-MBTL}
\subsection{Main Algorithm}\label{sec:M-MBTL}
\begin{figure*}[ht]
    \centering
    \includegraphics[width=0.95\linewidth]{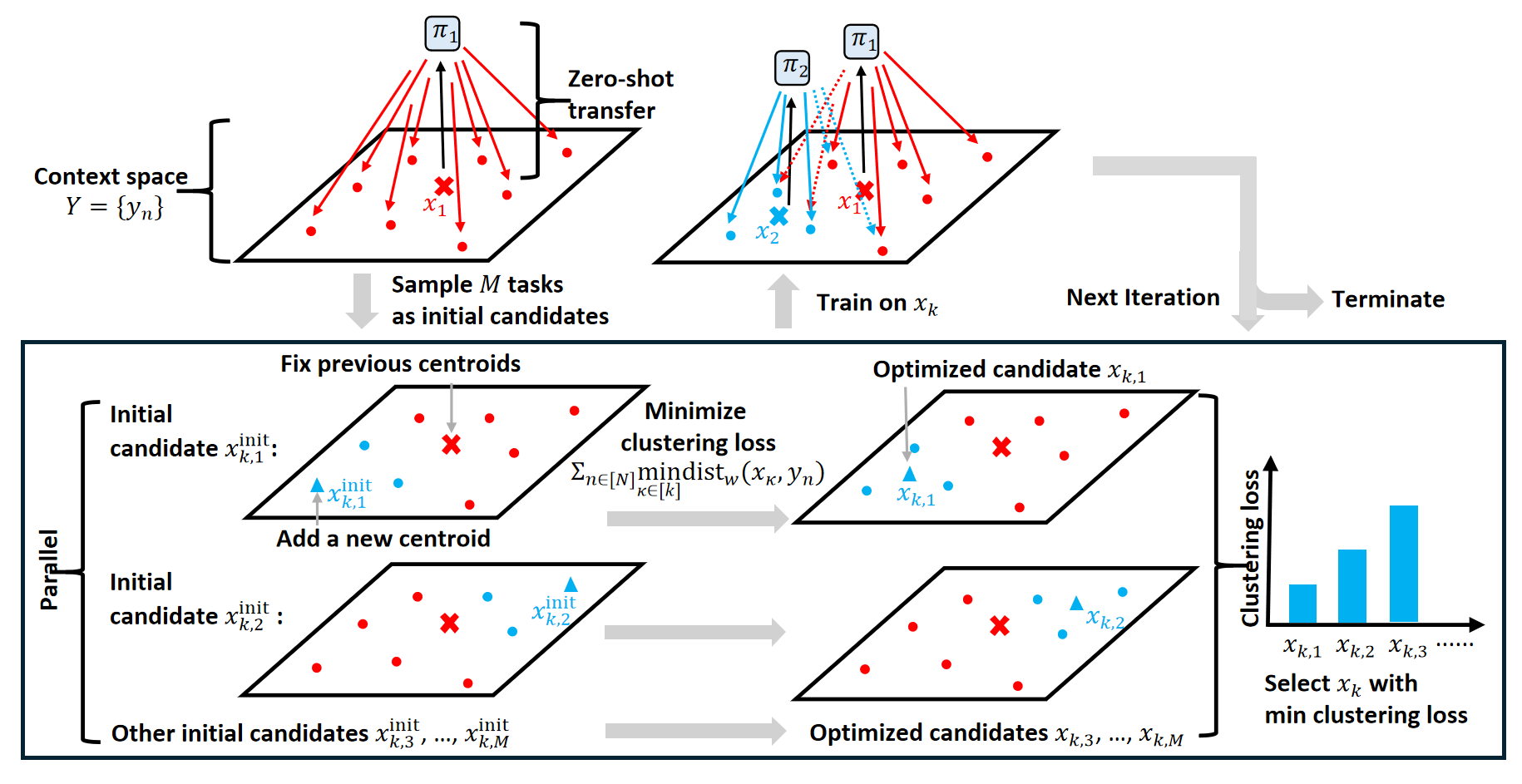}
    \caption{\textbf{Overview illustration for \jhedit{M-MBTL}.} 
    (a) A distance metric is used to estimate the generalization gaps between tasks and existing policies.  
    (b) A random restart approach is employed to search for the next training task. \( M \) tasks are sampled as initial candidates, with each initial candidate serving as a new centroid candidate while previous centroids are fixed. The candidate is optimized based on the clustering loss, and the optimized candidate with the minimum loss is selected as the next training task.  
    (c) Train on the selected task.}
    \label{fig:SC-MBTL} 
\end{figure*}

We propose \jhedit{\textsc{Mountain}} Model-Based Transfer Learning (\jhedit{M}-MBTL, \jhedit{Algorithm~\ref{alg:SC-MBTL})} as \jhedit{a} solution to \jhedit{G}STS, \jhedit{i}nspired by the K-Means clustering algorithm \citep{macqueen1967some}. However, K-Means and other traditional clustering methods \citep{dempster1977maximum, jain1988algorithms, kmeans++} do not support the sequential addition of centroids, where the added centroid corresponds to selecting a task for RL training.
To address this limitation, we extend the traditional clustering algorithm into a sequential form. 
Given a task set \( Y = \{y_n\}_{n \in [N]} \), \jhedit{M}-MBTL proceeds through \( K \) decision rounds. In each round, \jhedit{M}-MBTL aims to identify a new centroid as the training task \( x_k \), while keeping the previous centroids \( x_{1:k-1} \) fixed. However, this fixation of centroids---an important distinction between \jhedit{M}-MBTL and traditional clustering---cause\jhedit{s} the new centroid to converge to a local optimum positioned between the fixed centroids.
To address this issue, we employ the random restart technique \citep{random_restart}. This approach involves randomly sampling \( H \) task points \( \{x_{k,m}^{\text{init}}\}_{m \in [M]} \) from the target task set \( Y \), \tyyedit{updating} these candidates, and selecting the one with the best performance.  
During the call to $\text{Update}(x_{k,m}^{\text{init}}, x_{1:k-1})$, each initial candidate $x_{k,m}^{\text{init}}$ is added as a new centroid while the existing centroids $x_{1:k-1}$ remain fixed. We then refine this centroid by minimizing the clustering loss $\sum_{n=1}^{N}\min_{\kappa\le k}\operatorname{dist}(x_\kappa,y_n),$ yielding an updated candidate $x_{k,m}$ and its loss $l_{k,m}$. Full details appear in Algorithm \ref{alg:update}. \ttyedit{Figure \ref{fig:SC-MBTL} \jhedit{illustrates} an overview of M-MBTL.}
% During optimization, each initial candidate \( x_{k,m}^{\text{init}} \) is added as a new centroid while keeping \( x_{1:k-1} \) fixed. We then update the new centroid by minimizing the clustering loss: $\sum_{n \in [N]} \min_{\kappa \in [k]} \text{dist}(x_\kappa, y_n)$ \jhedit{(Algorithm~\ref{alg:update})}.
% This process yields an updated candidate \( x_{k,m} \) along with its corresponding loss \( l_{k,m} \). 
\jhedit{M}-MBTL selects the candidate with the lowest loss as the training task \( x_k \). 
\Tyedit{We introduce techniques to reduce the time complexity of M-MBTL, which are provided in Appendix \ref{sec:reduce_time_SC-MBTL}}. We also have a comparison between M-MBTL and GP-MBTL, which is provided in Appendix \ref{sec:comparison}. 

\begin{algorithm}[H]
\caption{\jhedit{M}-MBTL}
\label{alg:SC-MBTL}
\begin{algorithmic}
  \STATE {\bfseries Input:} Task set $Y=\{y_n\}_{n\in[N]}$, number of samples $M$.
  \STATE $\{x^{\text{init}}_{k,m}\}_{m\in[M]}=\text{Sample}(Y, M)$
  \FOR{$k\in [K]$}
      \FOR{$m\in[M]$}
          \STATE $x_{k,m}, l_{k,m}=\text{Update}(x^{\text{init}}_{k,m}, x_{1:k-1})$
      \ENDFOR
      \STATE $x_k=x_{k,\arg\min_{m} l_{k,m}}$
      \STATE Train on $x_k$
  \ENDFOR
\end{algorithmic}
\end{algorithm}

\begin{algorithm}[H]
\caption{Update}
\label{alg:update}
\begin{algorithmic}
  \STATE {\bfseries Input:} Initial point $x^{\text{init}}$, previous centroids $x_{1:k-1}$.
  \STATE $x_k=x^{\text{init}}$
  \WHILE{$x_k$ not converged}
      \FOR{$n\in[N],\kappa\in[k]$}
        \STATE $r_{n,\kappa}=\mathbb{I}(\kappa=\arg\max_{\kappa^\prime} \text{dist}(x_\kappa, y_n))$
      \ENDFOR
      \STATE $x_k = \arg\min_{x} \sum_{n\in[N]}r_{n,k} \text{dist}(x, y_n)$
  \ENDWHILE
  \STATE $l_k= \sum_{n\in[N],\kappa\in[k]}r_{n,\kappa}\text{dist}(x_\kappa, y_n)$
  \STATE {\bfseries Output:} $x_k$, $l_k$
\end{algorithmic}
\end{algorithm}

\subsection{Reducing Time Complexity of \jhedit{M-MBTL}}
\label{sec:reduce_time_SC-MBTL}
Since 1) \jhedit{M-MBTL} uses a fixed set of \( M \) samples as initial candidates, and 2) only the centroid \( x_k \) needs to be updated in the update process, we can reduce the time complexity in two key ways. Specifically, we introduce Get Performance (Algorithm \ref{alg:get_performance}) to replace Update (Algorithm \ref{alg:update}) in \jhedit{M-MBTL}.

First, some redundant calculations can be avoided in this process by updating each initial candidate. For a candidate indexed by \( m \) ($m\in [M]$), its optimized version from the previous round is \( x_{k-1,m} \). If the cluster of \( x_{k-1,m} \) does not intersect with the previous training task \( x_{k-1} \), this implies that the two centroids reduce the distances to different task points. In this case, we can leverage information from the previous round to update \( x_{k,m} \) and \( l_{k,m} \). 
Specifically, we define \( c(y_n, x_{1:k})\) as the cluster centroid of point \( y_n \) given all centroids \( x_{1:k} \):
\begin{equation}
   c(y_n, x_{1:k}) = \argmin_{x_\kappa \in x_{1:k}} \text{dist}(x_\kappa, y_n)
\end{equation}

Also, we define \( \text{Intersect}(x_{k-1,i}, x_{1:k-1})\) as the indicator that the cluster of \( x_{k-1,i} \) intersects with the cluster of \( x_{k-1} \):
\begin{equation}
\begin{aligned}
    &\text{Intersect}(x_{k-1,i}, x_{1:k-1})\\=& \mathbb{I}(\exists n \in [N],\, c(y_n, x_{1:k-1}) = x_{k-1}, \\
    &c(y_n, x_{1:k-2} \cup x_{k-1,i}) = x_{k-1,i}) 
\end{aligned}
\end{equation}
Define \( \Delta l_{k} = l_k - l_{k-1} \) as the reduced loss at round \( k \). If \( \text{Intersect}(x_{k-1,m}, x_{1:k-1}) \) is false, then \( x_{k,m} = x_{k-1,m} \) and \( l_{k,m} = l_{k-1,m} - \Delta l_{k-1} \), since the selected training task in this decision round and the previous round reduces the distances to different target tasks. Otherwise, we update \( x_{k,m} \) and \( l_{k,m} \) in the usual manner. The details are described in Algorithm \ref{alg:get_performance}.

Second, in the update process, since we only need to update \( x_k \), we do not need to calculate the distances between all task points and centroids in each iteration. Instead, we only need to compute the distance for the \( (n, \kappa) \) pairs influenced by \( x_k \). Define \( S(r)\) as the set of \( (n, \kappa) \) pairs where we need to update the distances and labels:
\begin{equation}
    \begin{aligned}
          S(r)= &\{(n,\kappa) \mid (r_{n,k} = 1, \kappa \in [k]) \, \\
          &\text{or} \, (r_{n,k} = 0, \kappa \in \{k\} \cup \{ \kappa^\prime \mid r_{n,\kappa^\prime} = 1 \})\} 
    \end{aligned}
\end{equation}
Here, \( r_{n,k} = 1, \kappa \in [k] \) indicates that if the point \( x_n \) belongs to cluster \( k \), we calculate its distance to all centroids. \( r_{n,k} = 0, \kappa \in \{k\} \cup \{ \kappa^\prime \mid r_{n,\kappa^\prime} = 1 \} \) indicates that if \( x_n \) does not belong to cluster \( k \), we only calculate its distance to its original centroid and the new centroid \( x_k \).
%\jhfn{maybe too lengthy for one sentence?}. 
This approach reduces the time complexity. The details are described in Algorithm \ref{alg:fast_update}.

\begin{algorithm}[tb]
   \caption{Get Performance}
   \label{alg:get_performance}
\begin{algorithmic}
   \STATE {\bfseries Input:} Initial point $x^{\text{init}}_{k,m}$, previous centroids $x_{1:k-1}$.
   \IF{not $\text{Intersect}(x_{k-1,m}$, $x_{1:k-1}$)}
    \STATE $x_{k,m}=x_{k-1, m}$
   \STATE $l_{k,m} = l_{k-1, m}-\Delta l_{k-1}$
   \ELSE
   \STATE $x_{k,m}, l_{k,m} = \text{Fast Update}(x^{\text{init}}_{k,m}, x_{1:k-1})$
   % \STATE $\{r_{n,k,i}\}_{n\in[N]} = \{r_{n,k-1,i}\}_{n\in[N]} $
   \ENDIF
   \STATE {\bfseries Output:} $x_{k,m}, l_{k,m}$
\end{algorithmic}
\end{algorithm}

\begin{algorithm}[tb]
   \caption{Fast Update}
   \label{alg:fast_update}
\begin{algorithmic}
   \STATE {\bfseries Input:} Initial point $x^{\text{init}}$, previous centroids $x_{1:k-1}$.
   \STATE $x_k=x^{\text{init}}$
   \STATE $d_{n,\kappa}=\text{dist}_{w_k}(x_{\kappa}, y_n), \forall n\in[N], \kappa\in[k]$
   \STATE $r_{n,\kappa}=\mathbb{I}(\kappa=\arg\max_{\kappa^\prime} d_{n,\kappa}), \forall n\in[N], \kappa\in[k]$
    \WHILE{$x_k$ not converged}
        \STATE $d_{n,\kappa}=\text{dist}_{w_k}(x_\kappa, y_n), \forall (n,\kappa)\in S(r)$
        \STATE $r_{n,\kappa}=\mathbb{I}(\kappa=\arg\max_{\kappa^\prime} d_{n,\kappa}), \forall (n,\kappa)\in S(r)$
        % \STATE $x_k=\frac{\sum_{n\in[N]} r_{n,k} x_n}{\sum_{n\in[N]} r_{n,k}}$
        \STATE $x_k = \arg\min_{x} \sum_{n\in[N]}r_{n,k} \text{dist}_{w_k}(x, y_n)$
    \ENDWHILE
    \STATE $l_k= \sum_{n\in[N],\kappa\in[k]}r_{n,\kappa}d_{n,\kappa}$
    \STATE {\bfseries Output:} $x_k$, $l_k$
\end{algorithmic}
\end{algorithm}

\subsection{Time comparison}\label{sec:time_compare}
\begin{figure*}[!t]%
    \centering
    \subfigure[CartPole]{
        \label{fig:result_time_cartpole_k100}
        \includegraphics[width=0.48\textwidth]{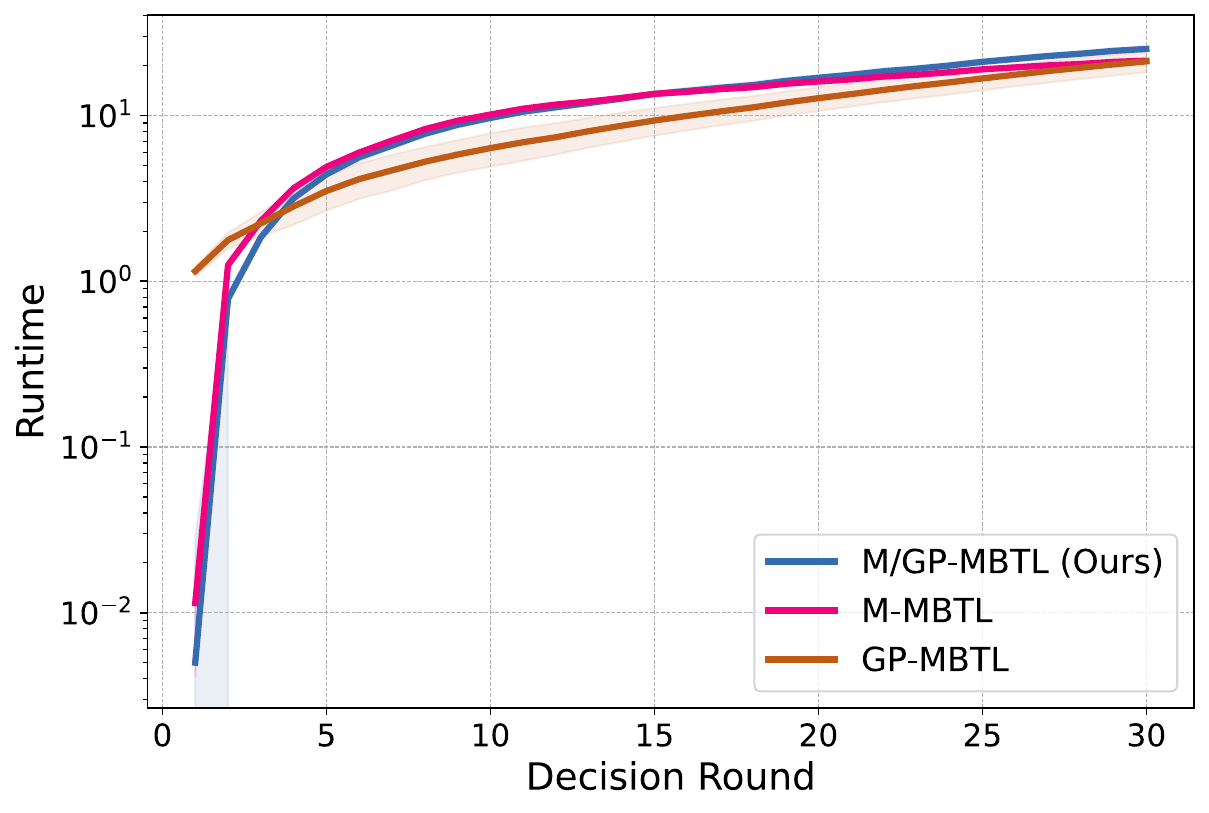}
        }
    \subfigure[BipedalWalker]{%
        \label{fig:result_time_walker_k256}%
        \includegraphics[width=0.48\textwidth]{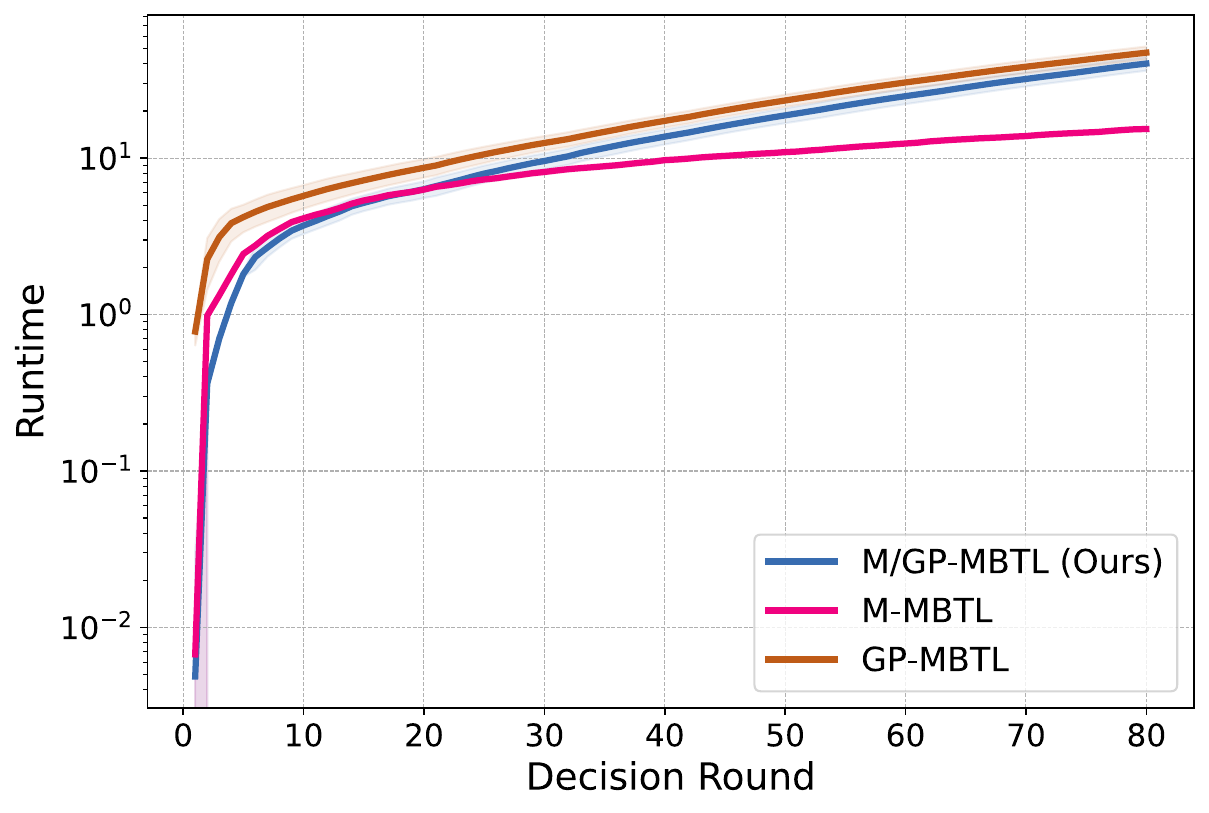}
        }%
    \\
    \subfigure[IntersectionZoo]{
        \label{fig:result_time_intersectionzoo_k216}
        \includegraphics[width=0.48\textwidth]{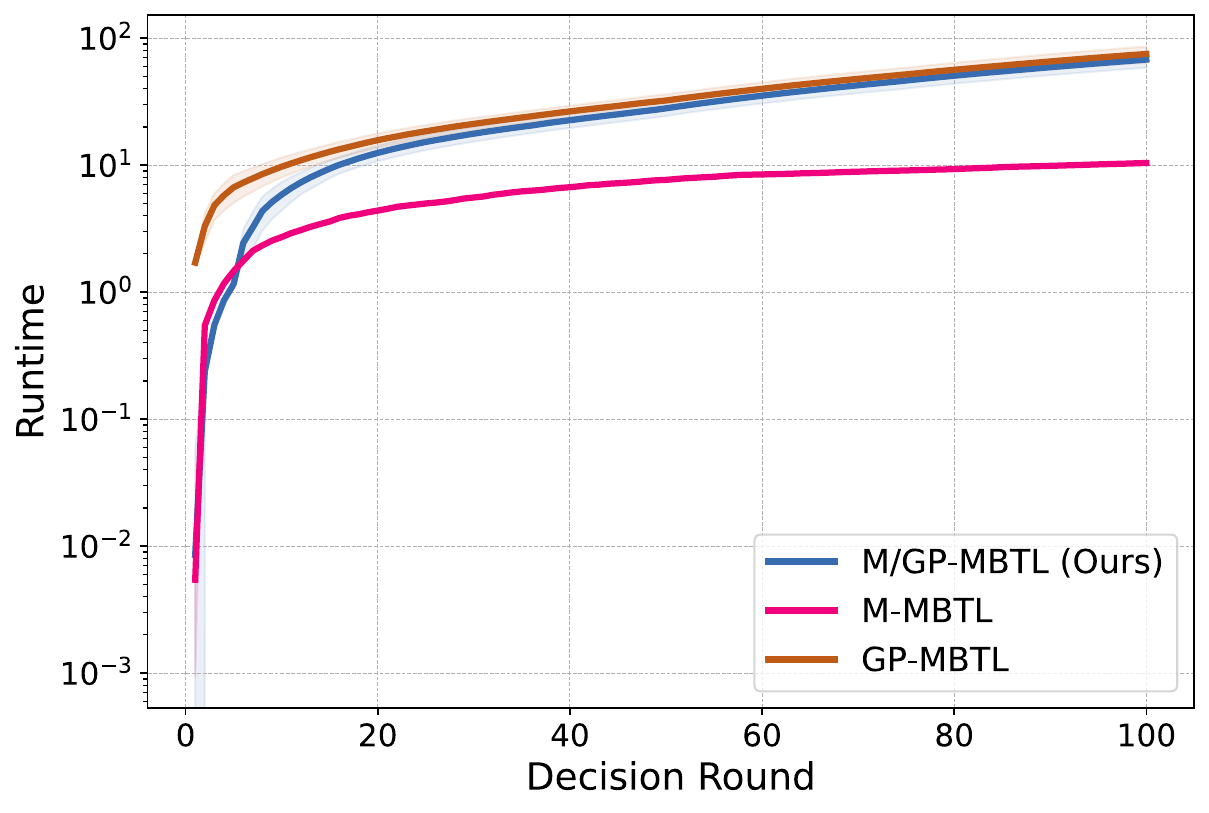}
        }%
    \subfigure[CyclesGym]{%
        \label{fig:result_time_crop_k216}%
        \includegraphics[width=0.48\textwidth]{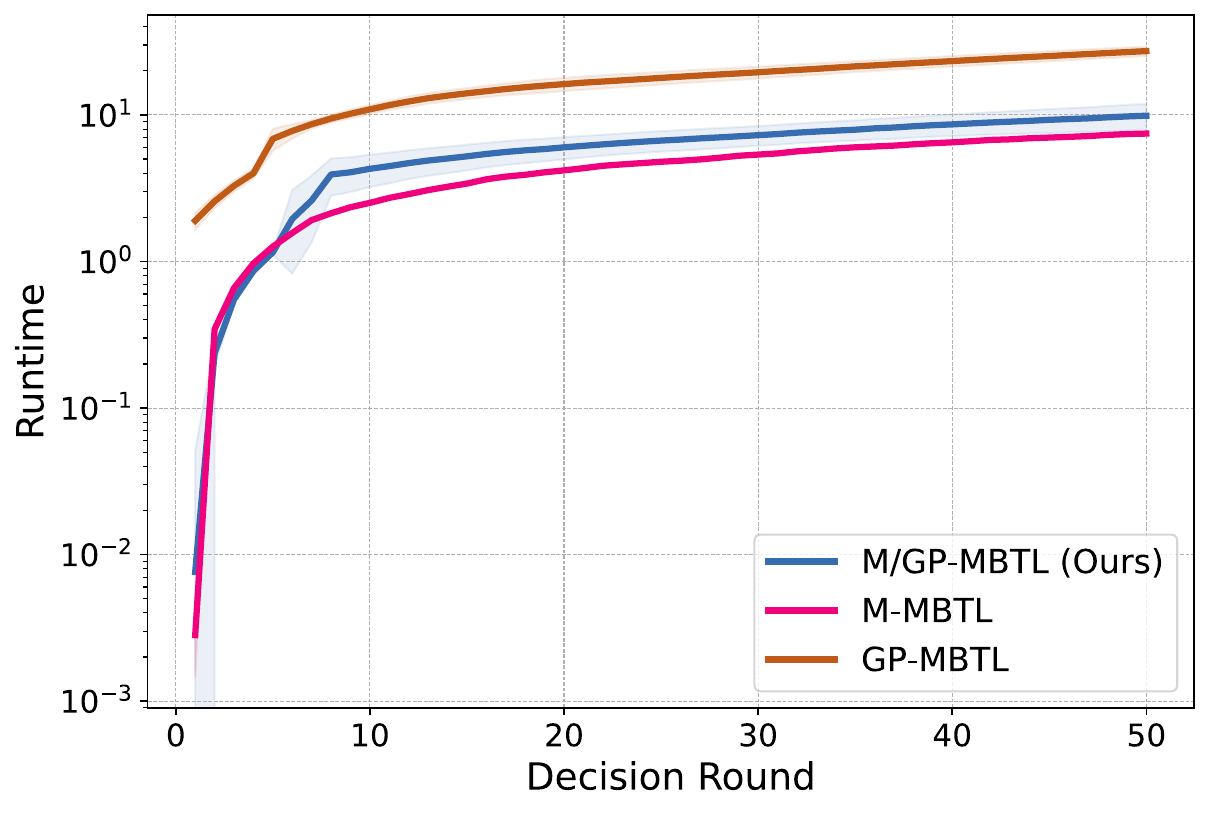}
        }%
    \caption{\Tyedit{CPU time comparison of M/GP-MBTL with M-MBTL and GP-\jhedit{MBTL}} on the CartPole, BipedalWalker, IntersectionZoo, and CyclesGym benchmarks.}
    \label{fig:result_time_all}
\end{figure*}

% \tyfn{Mistake here: Hybrid-MBTL in Figure \ref{fig:result_time_all} (a)}
\Tyedit{Figure \ref{fig:result_time_all} shows the CPU time comparison of M/GP-MBTL with M-MBTL and GP-MBTL.} When \( k \) is small, M-MBTL has a slightly higher runtime than GP-MBTL. However, as \( k \) increases, M-MBTL exhibits significantly lower runtime compared to GP-MBTL. This is because GP-MBTL requires considerable time to estimate the Gaussian Process model when handling a large number of training tasks. The runtime of M/GP-MBTL closely follows that of the algorithm it selects. \jhedit{Note that all reported runtimes exclude the time required for training the policies.}
% \cw{Clarify that this excludes the training time.}

\section{Details in \jhhedit{GP-MBTL}}\label{sec:detail_GP-MBTL}
\subsection{Gaussian Process Model-Based Transfer Learning \jhedit{\cite{MBTL}}}\label{sec:original_GP-MBTL}
\jhedit{A practical solution to solve \jhedit{G}STS problem is provided by MBTL~\citep{MBTL}, which we refer to as \textbf{GP-MBTL} in this work. The key idea is to separately estimate the source task performance with Gaussian Process (GP) Regression and model the generalization gap with a linear function of the context dissimilarity.
Although the original MBTL formulation targets one-dimensional contexts, it can be extended to higher dimensions by using norm-based distances.
Since continuous task spaces \(X\) can be large or even infinite in a multi-dimensional context space, GP-MBTL discretizes \(X\) into a finite grid, which enables applying BO methods on a manageable set of candidate source tasks.}

% \ty{TODO: introduce about GP-MBTL. Mention that GP-MBTL discretizes the continuous source task space. Mention about how to use GP to model the source task performance, and how to use acquisition function.}
\jhedit{At each iteration \(k\), GP-MBTL selects a new source task \(x_k\) by maximizing an acquisition function. Intuitively, the algorithm favors tasks \(x\) that (i) exhibit high estimated performance \(\mu_{k-1}(x)\), (ii) have high estimation uncertainty \(\sigma_{k-1}(x)\), and (iii) incur a small estimated generalization gap to other tasks in \(Y\). Concretely, an acquisition function is:}
% \cw{$\mu$ is not defined? Check that each variable is defined.}
\begin{equation}\label{eq:acquisition}
\begin{aligned}
    a(x;x_{1:k-1})=&\mathbb{E}_{y_n \in \tyyedit{\mathcal{U}(Y)}}[[\mu_{k-1}(x)+\beta_k^{1/2}\sigma_{k-1}(x)\\
    &-\Delta \hat{J}(\pi_x,y_n) - \Tyedit{\hat{J}_{k-1}(y_n)}]_+],
\end{aligned}
\end{equation}
\jhedit{where \(\beta_k\) is a user-specified exploration parameter, and \([\cdot]_+\) denotes \(\max(\cdot,\,0)\).}
\Tyedit{\(\Delta \hat{J}(\pi_x, y_n)\) models how much performance is \jhedit{\textit{estimated} to be} lost when transferring \(\pi_x\) to \(y_n\), while}
\(\jhedit{\hat{J}_{k-1}(y_n)}\) represents the \jhedit{estimated} best policy performance on \(y_n\) from previously trained source tasks.
\Tyedit{Thus, the acquisition function estimates the expected improvement from training on task $x$.}
\jhedit{By iterating this procedure---training a policy on the chosen source task, updating the GP estimates with the newly observed performance, and selecting the most promising next source task---GP-MBTL incrementally improves its selection of source tasks to maximize overall performance across \(Y\).}

\jhedit{Furthermore, we note that, in this paper, we leverage actual generalization performance when computing the acquisition function and enable online detection of the generalization gap’s slope beyond the original \Tyedit{GP-MBTL}. More details are provided in Appendix~\ref{sec:GP-MBTL}.}
\Tyedit{However, we found that GP-MBTL is not efficient in solving all CMDP problems. In the following section, we will show how we can detect and leverage CMDP structures for training task selection.} 

\subsection{Revised \jhhedit{GP-MBTL}}\label{sec:GP-MBTL}
\textbf{Update with Real Generalization Gaps.} The original \Tyedit{GP-MBTL}~\citep{MBTL} \cwwinline{uses \textit{modeled} generalization gaps despite having cheap} access to \textit{real} generalization gaps, \cwwinline{since} evaluation \jhedit{of} a policy is much cheaper than training. % , which \jhedit{takes} only a few episodes, 
\cwwinline{Thus, in this work,} \jhedit{we use} the real generalization performance \Tyedit{to calculate the best available performance \tyyedit{$\max_{x^\prime\in x_{1:k-1}} J(\pi_{x^\prime}, y_n)$}, \cwwinline{as opposed to \tyyedit{the estimated performance} in~\citep{MBTL}}. Thus, we get the acquisition function for task selection as follows:}
% , when calculating acquisition function for task selection, as follows:
\begin{equation}\label{eq:acquisition_revise}
\begin{aligned}
    a(x;x_{1:k-1})=&\mathbb{E}_{y_n \in \tyyedit{\mathcal{U}(Y)}}[[\mu_{k-1}(x)+\beta_k^{1/2}\sigma_{k-1}(x)\\&-\Delta \hat{J}(\pi_x,y_n) - \max_{x^\prime\in x_{1:k-1}} J(\pi_{x^\prime}, y_n)]_+],
\end{aligned}
\end{equation}

\textbf{Slope Learning.} The original GP-MBTL \jhedit{operates on a single $\theta$ value for generalization gap estimation ($\Delta J(\pi_x, y)=\theta |x-y|$, where $\theta\in\mathbb{R}^+$) under Assumption \ref{assump:distance}.} When extending the generalization gap model to multi-dimensional cases, we may consider different slopes for different \tyyedit{directions and different} dimension\jhedit{s}.
% \cw{Why call out different dimensions but not different directions?}
% \cw{Be consistent in naming throughout the paper. I see GP-MBTL and MBTL above.}
\Tyedit{Also, since the slopes are not necessarily positive as we discussed in Slope Criterion in Section \ref{sec:detection_approach}, we consider $\theta\in\mathbb{R}$.}
Recall that $\text{diff}_{\kappa, n}=[[x_\kappa-y_{\Tyedit{n}}]_-, [x_\kappa-y_{\Tyedit{n}}]_+]$. \Tyedit{Similar to Slope Criterion}, we propose to use a linear relationship $\hat{J}(\pi_{x_\kappa}, y_n)=\theta^\intercal \text{diff}_{\kappa, n}$, and learn the slope $\theta$ by linear regression:
\begin{equation}
\label{eq:linear_regression_2}
\begin{aligned}
    \theta =&\text{Linear Regression}(\{J(\pi_\kappa, y_n)\}_{n\in[N], \kappa\in[k]}, \\&\{\text{diff}_{\kappa, n}\}_{n\in[N], \kappa\in[k]}),
\end{aligned}
\end{equation}
where $\pi_\kappa=\pi_{x_{\kappa}}$.
In this way, GP-MBTL can automatically adapt to CMDP problems with varying slopes in the generalization gap of different dimensions (\jhedit{Algorithm~\ref{alg:GP-MBTL}}).
% \jhfn{we use different acquisition function from equation~\ref{eq:acquisition}.}

\begin{algorithm}[tb]
   \caption{GP-MBTL}
   \label{alg:GP-MBTL}
    \begin{algorithmic}
        \STATE {\bfseries Input:} Task set $Y=\{y_n\}_{n\in[N]}$.
        \STATE Discretize $X$ to $\overline{X}$
        \STATE Initialize $J,V=0,\forall x\in \overline{X}$
        \FOR{$k\in [K]$}
            \IF{\jhedit{$k=1$}}
            \STATE \jhedit{$x_1 \gets \mathrm{median}(\overline X)$}
            \ELSE
            \STATE $\mu\jhedit{_{k-1}}, \sigma\jhedit{_{k-1}}\jhedit{\gets}\mathcal{GP}(\mathbb{E}\jhedit{_{x\in x_{1:k-1}}}[J(\pi_x,x)], \jhedit{\kappa}(x,\tilde{x}))$
            \STATE Calculate $a(x;x_{1:k-1})$ with Eq. \ref{eq:acquisition_revise}
            \STATE $x_k=\arg\max_x a(x;x_{1:k-1})$
            \ENDIF
            \STATE Train on $x_k$, get $\pi_{x_k}$
            \STATE Transfer to target tasks, receive $\{J(\pi_{x_k}, y_n)\}_{n\in[N]}$
            \STATE Calculate $\theta$ with Eq. \ref{eq:linear_regression_2}
       \ENDFOR
    \end{algorithmic}
\end{algorithm}

\subsection{\cwwinline{GP library}}
\tyyedit{Another modification concerns the Gaussian-process implementation in GP-MBTL. While \citet{MBTL} relies on \texttt{GaussianProcessRegressor} from \textsf{scikit-learn}, which must be retrained from scratch at every iteration $k$ and thus becomes a computational bottleneck as $k$ grows, we instead adopt \texttt{gpytorch.models.ExactGP} \citep{gpytorch2018}. Its neural-network parameterization of the mean and covariance enables efficient online updates, eliminating the need for full retraining at each step.}
% \cwwinline{Another modification...}\cw{Include the change in GP library and show a figure illustrating the computational scaling.}
\section{\tyyedit{Comparison between M-MBTL and GP-MBTL}}\label{sec:comparison}
\tyyedit{
We compare M-MBTL and GP-MBTL through the generalization-performance decomposition introduced in Definition \ref{def:general_model}.
Under \textsc{Mountain} structure we set $f(x)=C_{1}$ and $h(x,y)=\text{dist}(x,y)$; consequently, M-MBTL needs only to estimate $g(y)$ via Eq. \eqref{eq:approx_J_target}.
GP-MBTL \cite{MBTL}, by contrast, assumes a linear generalization gap $\Delta J(\pi_{x},y)\;=\;J(\pi_{x},x)-J(\pi_{x},y)\;=\;g(x)-g(y)-h(x,y)$
which can be interpreted as a constant task difficulty $g(y)=C$ plus a linear interaction $h(x,y)$. GP-MBTL estimates the source-task performance using a Gaussian Process. With $g(y)$ constant, it is equivalent to estimating $f(x)$. However, when task difficulty is not constant, this approximation becomes inaccurate and GP-MBTL may underperform M-MBTL.
However, our revised GP-MBTL (\cwwinline{Appendix}~\ref{sec:GP-MBTL}) mitigates this issue by using actual generalization performance to compute the acquisition function. As a result, errors in estimating the generalization performance of trained policies do not directly affect task selection, reducing the impact of this problem. Consequently, GP-MBTL can still perform better than M-\jhedit{MBTL} in some cases, especially when \textsc{Mountain} structure is violated. Thus, we use GP-MBTL as a more general solution for M/GP-MBTL when \textsc{Mountain} is not detected.}

\section{Details in Structure Detection Model-Based Transfer Learning}\label{sec:detail_SD-MBTL}
Below, we present the details about key algorithms that form the backbone of our SD-MBTL framework.
Algorithm~\ref{alg:SD-MBTL} outlines the generic Structure Detection Model-Based Transfer Learning (SD-MBTL) framework. \Tyedit{The input of SD-MBTL includes a task set $Y$, a CMDP structures set $S$, a detection algorithm $\text{Detect}:\mathbb{R}^{k\times N}\to S$, and MBTL algorithms ($\text{Alg}_1,\text{Alg}_2,...,\text{Alg}_{|S|}$).} In each decision round, the algorithm first detects the underlying CMDP structure based on observed generalization performance and then uses a mapping function to select the corresponding MBTL algorithm. The chosen algorithm then selects the next training task, which is subsequently used to update the performance estimates.

\begin{algorithm}[htbp]
   \caption{SD-MBTL}
   \label{alg:SD-MBTL}
    \begin{algorithmic}
        \STATE {\bfseries Input:} Task set $Y=\{y_n\}_{n\in[N]}$, CMDP structures set $S$, Detection algorithm $\text{Detect}$, MBTL algorithms $\text{Alg}_1,\text{Alg}_2,..,\text{Alg}_A$.
        \FOR{$k\in[K]$}
            \STATE $\Tyedit{s_i}=\text{Detect}(\{J(\pi_\kappa, y_n)\}_{n\in[N], \kappa\in[k-1]})$
            % \jhfn{You may need mapping from $s_i$ to Alg$_i$?}
            % \STATE $a=\text{Map}(s)$
            \STATE Run \Tyedit{$\text{Alg}_i$ corresponding to the structure $s_i$}, get training task $x_k$
            \STATE Train on $x_k$, receive $\{J(\pi_k, y_n)\}_{n\in[N]}$
        \ENDFOR
    \end{algorithmic}
\end{algorithm}
\section{Oracle}\label{sec:oracle}
\subsection{Definitions}\label{sec:oracle_def}
% \cwwinline{We consider two Oracle definitions, which coincide with optimally solving the STS problem and an optimal myopic solution, respectively. Due to the small state and action space, the SSTS problem lends itself to be solved exactly using dynamic programming. An optimal myopic solution is one in which the policy gets to observe the current state but refuses to consider the future, corresponding to the best bandit-like policy.}
We define Myopic Oracle as an optimal myopic solution to solve the \jhedit{G}STS problem.

\begin{definition}[\cwwinline{Myopic Oracle}]
    \tyyedit{Myopic oracle greedily selects the training task that maximizes the expected generalization performance over target task $y$ and $W$ trial in each step $k$.}
    \begin{equation}
    \begin{aligned}
        x_{k} = \argmax_{x\in X \setminus \{x_{1:k-1}\}} \frac{1}{W} \sum_{w=1}^W \mathbb{E}_{y\in\mathcal{U}(Y)}[\max_{x^\prime\in x_{1:k-1}\cup\{x\}} J_{(w)}(\pi_{x^\prime},y)]
    \end{aligned}
    \end{equation}
\end{definition}
% \subsection{Theoretical analysis}
% \cw{Provide a Lemma on the optimality of the Oracle.}

% \cw{Provide a Lemma on the optimality of the Myopic Oracle.}

\subsection{Discussion}
For completeness, we provide a formal definition of the oracle introduced in \cite{MBTL}, termed the ``Sequential Oracle." Unlike the myopic oracle, the Sequential Oracle selects a different training task for each experimental trial, as though it knew the random seed in advance---an idealized capability that is unattainable in practice.
\begin{definition}[Sequential Oracle]
    \tyyedit{For each trial $w$, Sequential oracle greedily selects the training task $x_k^w$ that maximizes the expected generalization performance over target task $y$ in each step $k$.}
    \begin{equation}
        \tyyedit{x_{k}^{w} = \argmax_{x\in X \setminus \{x_{1:k-1}\}} \mathbb{E}_{y\in\mathcal{U}(Y)}[\max_{x^\prime\in x_{1:k-1}\cup\{x\}} J_{(w)}(\pi_{x^\prime},y)]}
    \end{equation}
\end{definition}

\section{Aggregated Metric}\label{sec:aggregated_metric}
\tedit{On the CMDP benchmarks, we design an aggregated performance metric, an adaptation of the widely used human-normalized score (HNS) in \cite{agent57} and \cite{Mnih2015HumanlevelCT}
. This helps us evaluate how baseline algorithms perform across multiple benchmarks---particularly suited for MBTL-based methods. For each benchmark, we calculate an algorithm's performance by subtracting the Random baseline and then dividing by the difference between the Myopic Oracle's performance and the Random baseline. This yields a normalized performance. Formally,
\begin{equation}
    \text{Normalized Performance}_{j} = \frac{\text{Performance}_{j} - \text{Random}_j}{\text{Myopic Oracle}_j - \text{Random}_j},
\end{equation}
where the subscription $j$ represents the performance at benchmark $j$. We then average an algorithm’s normalized performance across the four benchmarks to obtain its aggregated performance, as recommended by \cite{rliable}:
\begin{equation}
    \text{Aggregated Performance} = \sum_{j}\,\text{Normalized Performance}_{j}
\end{equation}
Under this metric, a value of 0 represents the performance of the Random baseline, while a value of 1 represents the performance of the Myopic Oracle.
This metric scales each MBTL-based algorithm’s score between 0 and 1, indicating how much it outperforms the Random baseline and how closely it approaches the Myopic Oracle.}

\section{Implementation Details}\label{sec:implementation_details}
\subsection{\tyyedit{Construction of transfer matrix.}}
\jhedit{For the ease of our analysis, we first construct the transfer matrix. To do so,} we \jhedit{first} train each task in the discrete task set independently. \tyyedit{This provides the policies trained on all source tasks, making it easier to compare the performance of different algorithms in source task selection.} 
Based on the \jhedit{average performance} after the transfer, we construct a transfer matrix. The transfer matrix is an \( N \times N \) matrix, where the element at the \( i \)-th row and \( j \)-th column represents the \jhedit{generalization performance} of policy trained from $i$-th task and applied to $j$-th task. 
\jhedit{This also allows us to try many different types of multi-policy training} algorithms \jhedit{that} can \cwwinline{layer on top of} any DRL algorithm \jhedit{with the low cost of training}. 
\tyyedit{The above experiments were repeated three times.} 

\subsection{Benchmarks}\label{sec:benchmark_details}
\paragraph{CartPole}
For the CartPole task, we employ the default environment \ttyedit{context variables} provided by the CARL library \citep{benjamins2023contextualizecasecontext}: a pole length of 0.5, a cart mass of 1.0, and a pole mass of 0.1. To generate three-dimensional context variations, each variable was varied over a range from 0.2 to 2 times its default value, resulting in \(N = 9 \times 10 \times 10\) unique contexts. Each independent training run is conducted for five million simulation timesteps.
\begin{figure}[ht]
    \centering
    \includegraphics[width=0.9\linewidth]{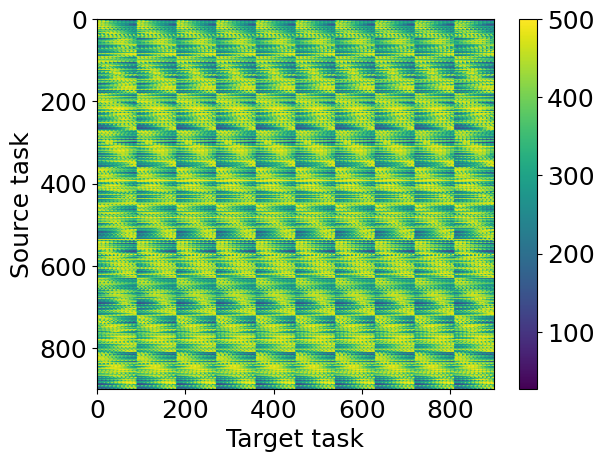}
    \caption{Transfer matrices of three context variables in the CartPole task (pole length, cart mass, pole mass). Bright\jhedit{er} colors indicate high\jhedit{er} \jhedit{generalization} performance.}
    \label{fig:m_cartpole} 
\end{figure}

\begin{figure*}[!t]%
    \centering
    \subfigure[CartPole: Pole length]{
        \label{fig:Len_pole}
        \includegraphics[width=0.33\textwidth]{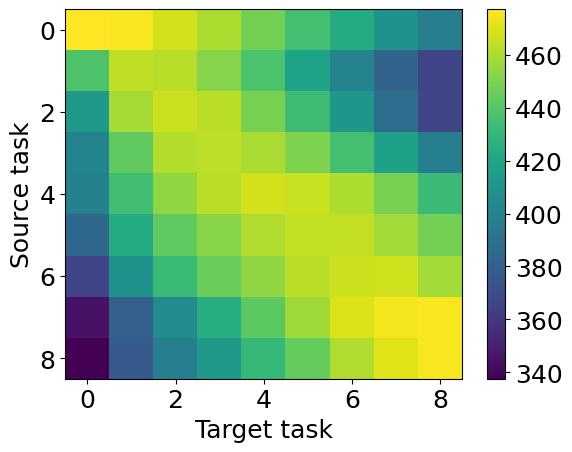}
        }%
    \subfigure[CartPole: Cart mass]{%
        \label{fig:Mass_cart}%
        \includegraphics[width=0.33\textwidth]{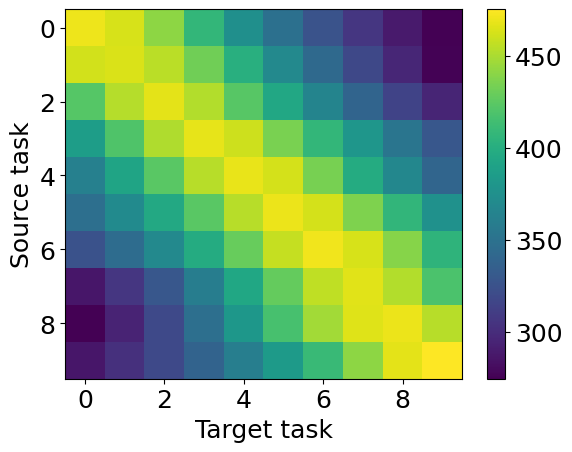}
        }%
    \subfigure[CartPole: Pole mass]{%
        \label{fig:Pole_mass}%
        \includegraphics[width=0.33\textwidth]{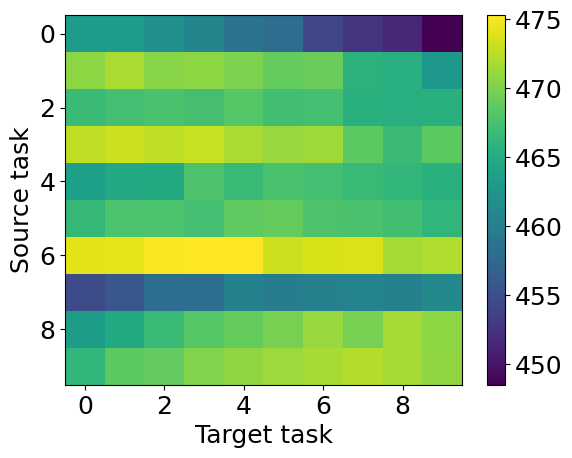}
        }%
    \caption{Average transfer matrices of three context variables in the CartPole task (pole length, cart mass, pole mass). The average transfer matrix is computed as the average of local transfer matrices, where each local transfer matrix varies only one variable while keeping the other two variables constant. Bright\jhedit{er} colors indicate high\jhedit{er} \jhedit{generalization} performance.}
    \label{fig:average_transfer_matrix_cartpole}
\end{figure*}
% \jhfn{Consider adding transfer matrices, overall, and three contexts each.}

\paragraph{BipedalWalker}
For the BipedalWalker CMDP, we similarly modify the environment using the CARL library. In this benchmark, we vary three context variables---friction, gravity, and scale---over a range from 0.2 to 1.6 times their default values, yielding \(N = 8 \times 8 \times 8\) distinct tasks. The default settings for BipedalWalker are a friction value of 2.5, a scale of 30, and a gravity of 10. As with CartPole, each independent training run is executed for five million time steps, and the best-performing model is selected for zero-shot transfer evaluation.

CARL benchmarks are distributed under the Apache License 2.0.
\begin{figure}[ht]
    \centering
    \includegraphics[width=0.9\linewidth]{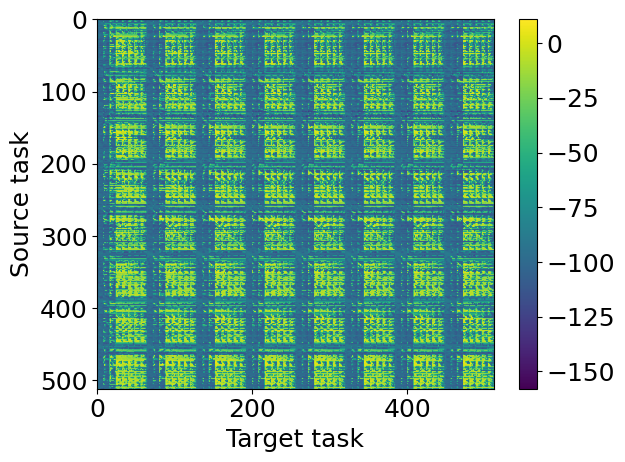}
    \caption{Transfer matrices of three context variables in the Bipedal Walker task (gravity, scale, friction). Bright\jhedit{er} colors indicate high\jhedit{er} \jhedit{generalization} performance.}
    \label{fig:m_walker} 
\end{figure}
% \jhfn{Consider adding transfer matrices, overall, and three contexts each.}
\begin{figure*}[!t]%
    \centering
    \subfigure[BipedalWalker: Gravity]{
        \label{fig:Gravity}
        \includegraphics[width=0.33\textwidth]{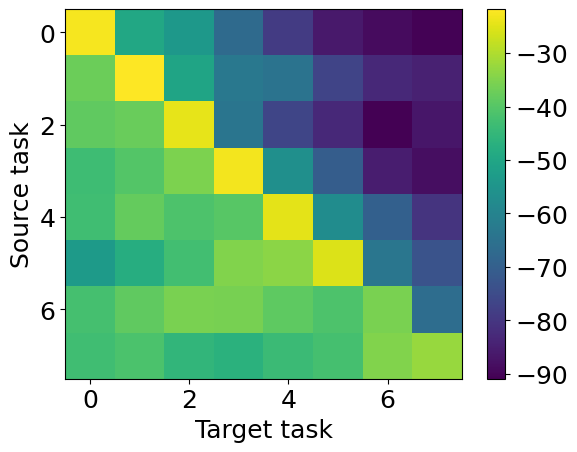}
        }%
    \subfigure[BipedalWalker: Scale]{%
        \label{fig:Scale}%
        \includegraphics[width=0.33\textwidth]{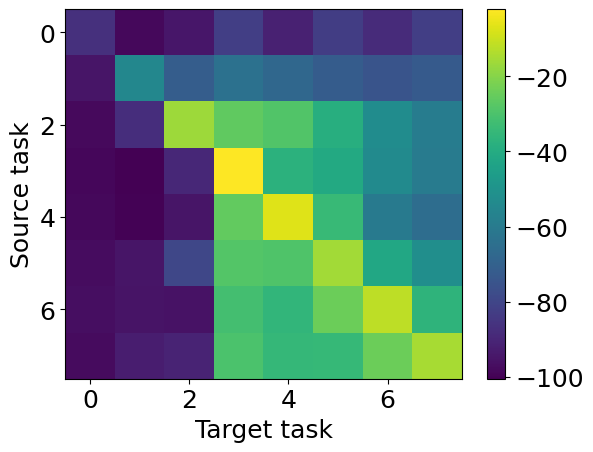}
        }%
    \subfigure[BipedalWalker: Friction]{%
        \label{fig:Friction}%
        \includegraphics[width=0.33\textwidth]{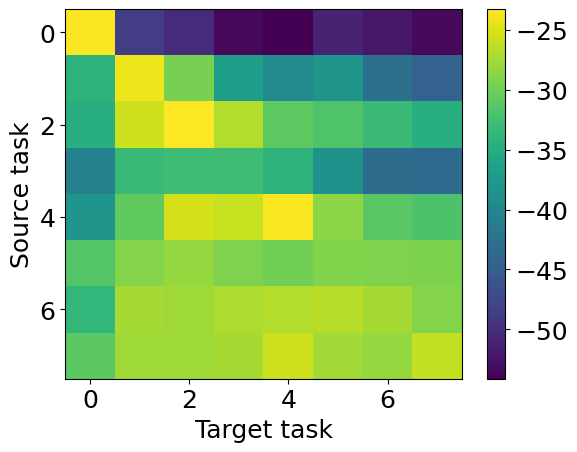}
        }%
    \caption{Average transfer matrices of three context variables in the Bipedal Walker task (gravity, scale, friction). The average transfer matrix is computed as the average of local transfer matrices, where each local transfer matrix varies only one variable while keeping the other two variables constant. Bright\jhedit{er} colors indicate high\jhedit{er} \jhedit{generalization} performance.}
    \label{fig:average_transfer_matrix_walker}
\end{figure*}
\paragraph{IntersectionZoo}
For the IntersectionZoo benchmark, we configure the simulation with a default inflow of 250 vehicles per hour, an autonomous vehicle penetration rate of 0.5, and a green phase duration of 20 seconds to replicate realistic urban traffic conditions. The CMDP context variables vary from 0.1 to 1.2 times their default values, resulting in a total of \(6^3\) distinct contexts. We employ version 1.16.0 of the microscopic traffic simulator, Simulation of Urban MObility (SUMO) \citep{SUMO2018} v.1.16.0.
For further details on the experimental setup and hyperparameter configurations, please refer to \cite{jayawardana2025intersectionzoo}.
The IntersectionZoo benchmark falls under the MIT License.

\begin{figure}[ht]
    \centering
    \includegraphics[width=0.9\linewidth]{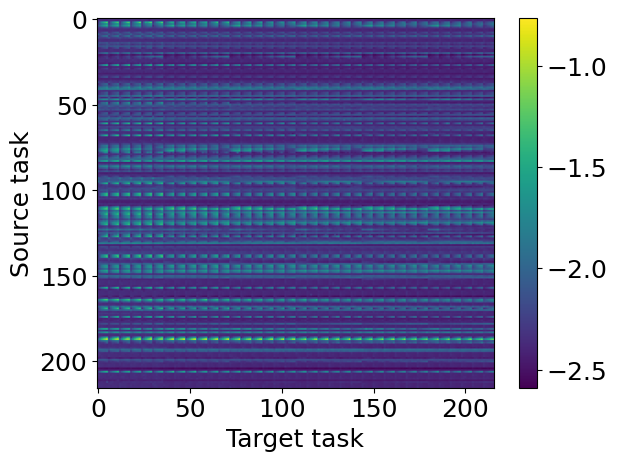}
    \caption{Transfer matrices of three context variables in the IntersectionZoo task (inflow rate, autonomous vehicle (AV) penetration rate, signal green-phase duration). Bright\jhedit{er} colors indicate high\jhedit{er} \jhedit{generalization} performance.}
    \label{fig:m_zoo} 
\end{figure}
% \jhfn{Consider adding transfer matrices, overall, and three contexts each.}
\begin{figure*}[!t]%
    \centering
    \subfigure[IntersectionZoo: Inflow rate]{
        \label{fig:zoo1}
        \includegraphics[width=0.33\textwidth]{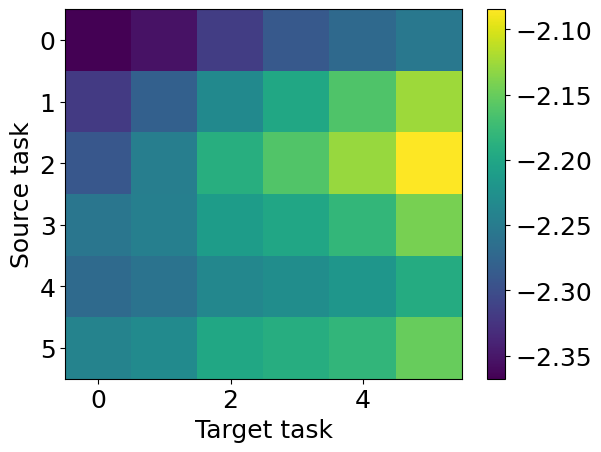}
        }%
    \subfigure[IntersectionZoo: AV penetration rate]{%
        \label{fig:zoo2}%
        \includegraphics[width=0.33\textwidth]{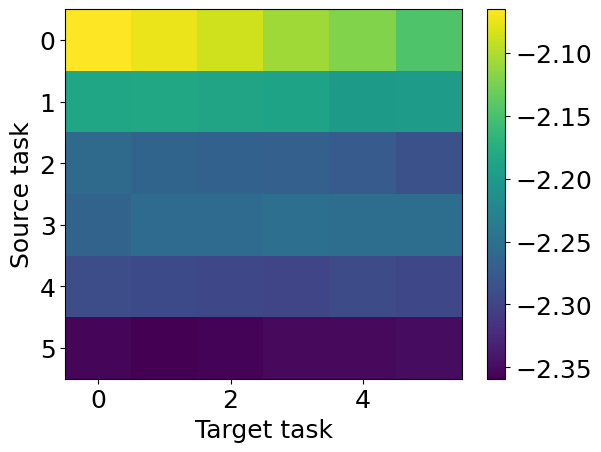}
        }%
    \subfigure[IntersectionZoo:  Green-phase duration]{%
        \label{fig:zoo3}%
        \includegraphics[width=0.33\textwidth]{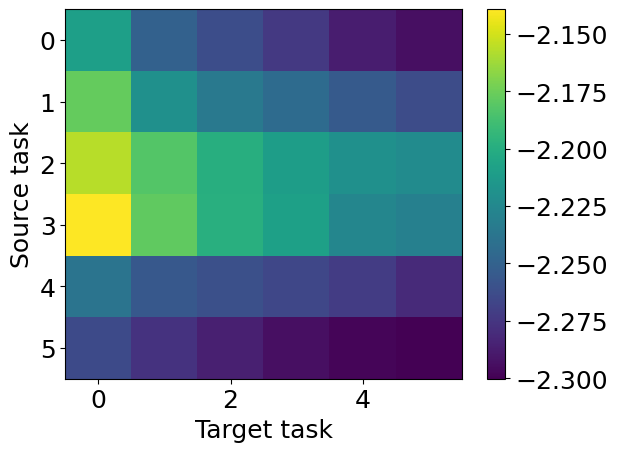}
        }%
    \caption{Average transfer matrices of three context variables in the IntersectionZoo task (inflow rate, autonomous vehicle (AV) penetration rate, signal green-phase duration). The average transfer matrix is computed as the average of local transfer matrices, where in each local transfer matrix, only one variable is changed while the other two variables are kept constant. Bright\jhedit{er} colors indicate high\jhedit{er} \jhedit{generalization} performance.}
    \label{fig:average_transfer_matrix_intersectionzoo}
\end{figure*}

\paragraph{CyclesGym}
For CyclesGym benchmark \citep{turchetta2022learning}, we start with default environmental context variables that simulate realistic agricultural conditions---specifically, a default precipitation of 10.0, a temperature of 20.0, and a sunlight (solar) intensity of 15.0. To generate a diverse set of contexts, we independently scale each of these context variables using scaling factors drawn from a predefined list (e.g., \jhedit{0.5, 1.0, 1.5, 2.0, 2.5, 3.0}), forming a three-dimensional grid of contexts (amounting to $6^3$ combinations). These context variables are set up to evaluate our method’s performance thoroughly across various simulated agricultural scenarios. Each trial is run three times (using seeds 0, 1, and 2) to ensure robustness under fixed weather and non-adaptive settings. 
CyclesGym benchmark falls under the BSD 3-Clause License.

\jhedit{Detailed context variables and their ranges are listed in Table~\ref{tab:env_context_params_detailed}.}

\begin{figure}[ht]
    \centering
    \includegraphics[width=0.9\linewidth]{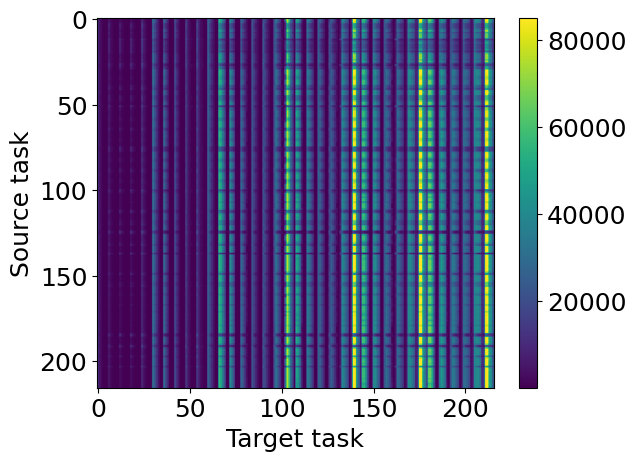}
    \caption{Transfer matrices of three context variables in the \jhedit{CyclesGym} (precipitation, temperature, sunlight). Bright\jhedit{er} colors indicate high\jhedit{er} \jhedit{generalization} performance.}
    \label{fig:m_crop} 
\end{figure}
% \jhfn{Consider adding transfer matrices, overall, and three contexts each.}
\begin{figure*}[!t]%
    \centering
    \subfigure[CyclesGym: Precipitation]{
        \label{fig:Crop1}
        \includegraphics[width=0.33\textwidth]{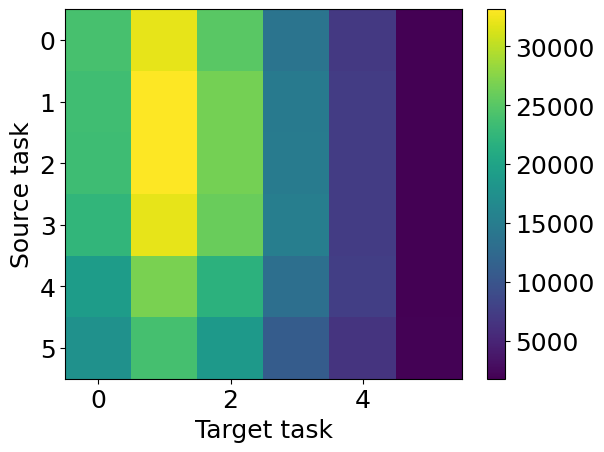}
        }%
    \subfigure[CyclesGym: Temperature]{%
        \label{fig:Crop2}%
        \includegraphics[width=0.33\textwidth]{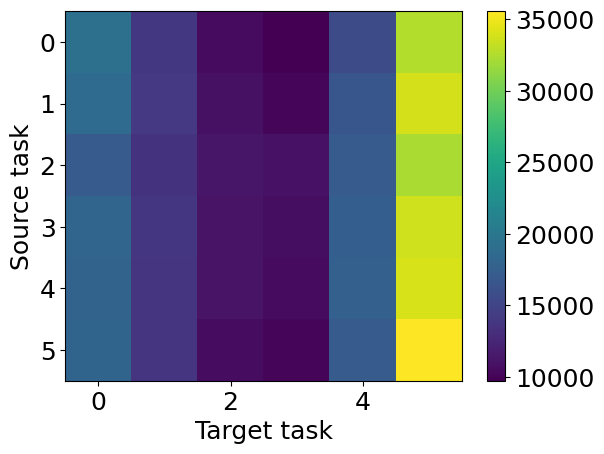}
        }%
    \subfigure[CyclesGym: Sunlight]{%
        \label{fig:Crop3}%
        \includegraphics[width=0.33\textwidth]{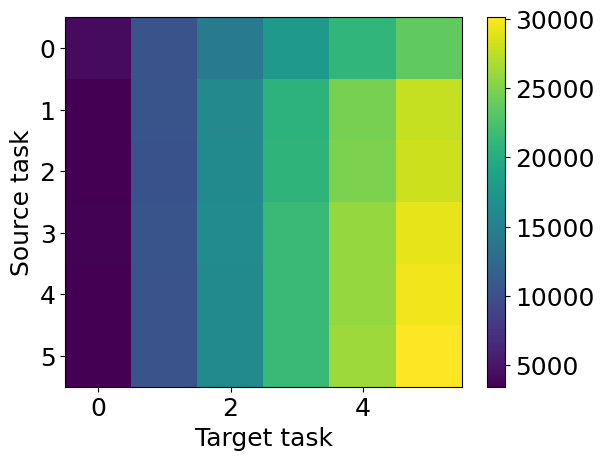}
        }%
    \caption{Average transfer matrices of three context variables in the \jhedit{CyclesGym} (precipitation, temperature, sunlight). The average transfer matrix is computed as the average of local transfer matrices, where each local transfer matrix varies only one variable while keeping the other two variables constant. Bright\jhedit{er} colors indicate high\jhedit{er} \jhedit{generalization} performance.}
    \label{fig:average_transfer_matrix_cyclesgym}
\end{figure*}

\begin{table*}[ht]
\centering
\caption{Environmental context variables for each benchmark.}
\label{tab:env_context_params_detailed}
\begin{tabular}{@{}c| c c c c@{}}
\toprule
\textbf{Benchmark}    & \textbf{Variable}        & \textbf{Default}            & \textbf{Variation Range}        & \textbf{\# Values} \\ 
\midrule
CartPole             & Pole length               & 0.5                         & 0.4 - 2.0 x default              & 9                  \\
CartPole             & Cart mass                 & 1.0                         & 0.2 - 2.0 x default              & 10                 \\
CartPole             & Pole mass                 & 0.1                         & 0.2 - 2.0 x default              & 10                 \\
\addlinespace
BipedalWalker        & Friction                  & 2.5                         & 0.2 - 1.6 x default              & 8                  \\
BipedalWalker        & Scale                      & 30                          & 0.2 - 1.6 x default              & 8                  \\
BipedalWalker        & Gravity                   & 10                          & 0.2 - 1.6 x default              & 8                  \\
\addlinespace
IntersectionZoo      & Inflow rate               & 250 veh/h                   & 0.1 - 1.2 x default              & 6                  \\
IntersectionZoo      & AV penetration rate       & 0.5                         & 0.1 - 1.2 x default              & 6                  \\
IntersectionZoo      & Green-phase duration      & 20 s                        & 0.1 - 1.2 x default              & 6                  \\
\addlinespace
CyclesGym            & Precipitation             & 10.0                        & \{\jhhedit{0.5, 1.0, 1.5, 2.0, 2.5, 3.0}\} x default & 6                  \\
CyclesGym            & Temperature               & 20.0                        & \{\jhhedit{0.75, 1.0, 1.25, 1.5, 1.75, 2.0}\} x default & 6                  \\
CyclesGym            & Sunlight        & 15.0                        & \{\jhhedit{0.5, 1.0, 1.5, 2.0, 2.5, 3.0}\} x default & 6                  \\
\bottomrule
\end{tabular}
\end{table*}

\subsection{\jhedit{Synthetic Data}}\label{sec:synthetic-data}
\begin{figure*}[!t]%
    \centering
    \subfigure[noise=0; $f(x)$ constant; $g(y)$ none; $h(x,y)$ L$_1$ norm]{
        \includegraphics[width=0.22\textwidth]{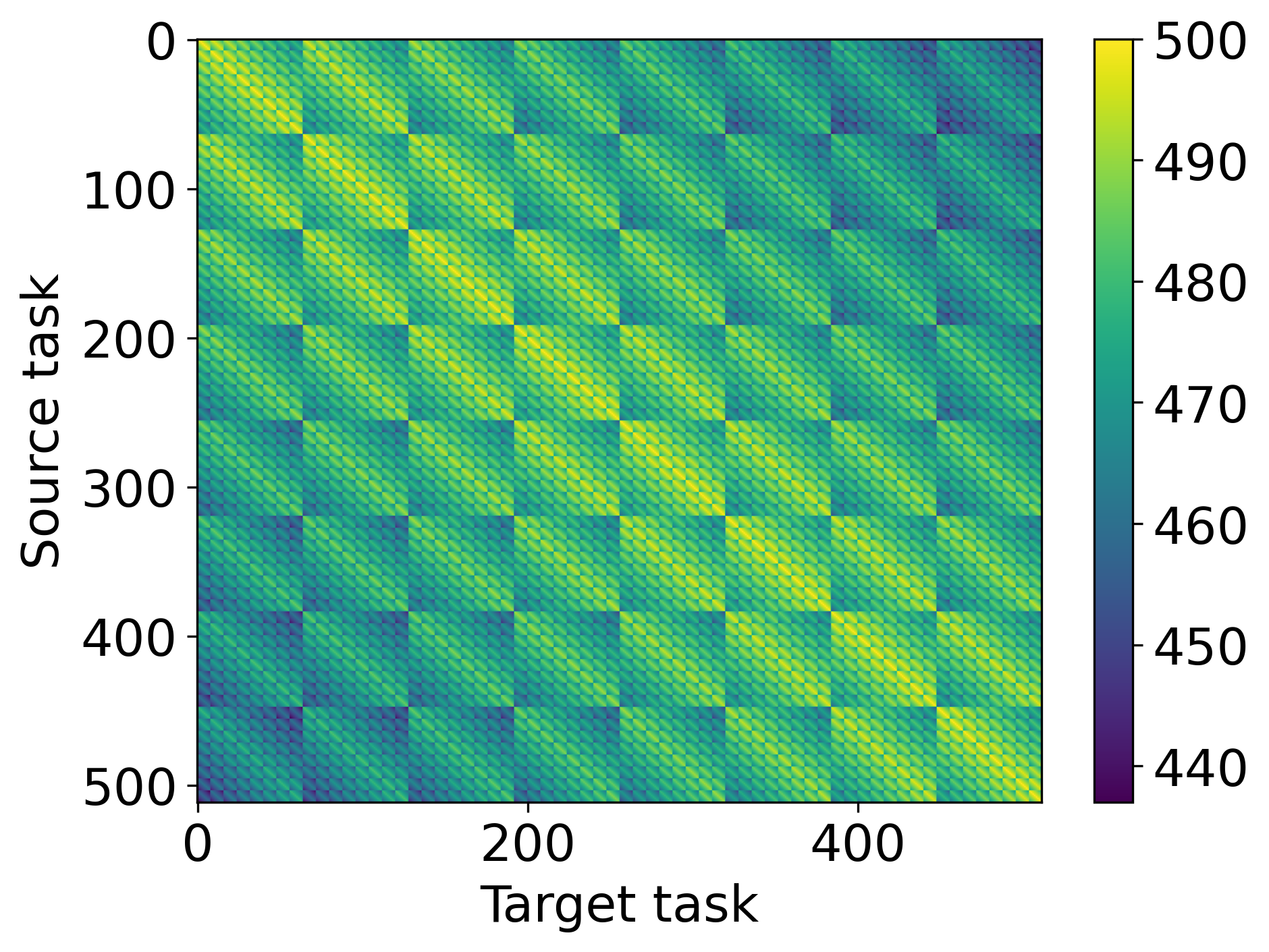}
        }%
    \hfill
    \subfigure[$\sigma=0$; $f(x)$ constant; $g(y)$ linear; $h(x,y)$ Non-distance]{
        \includegraphics[width=0.22\textwidth]{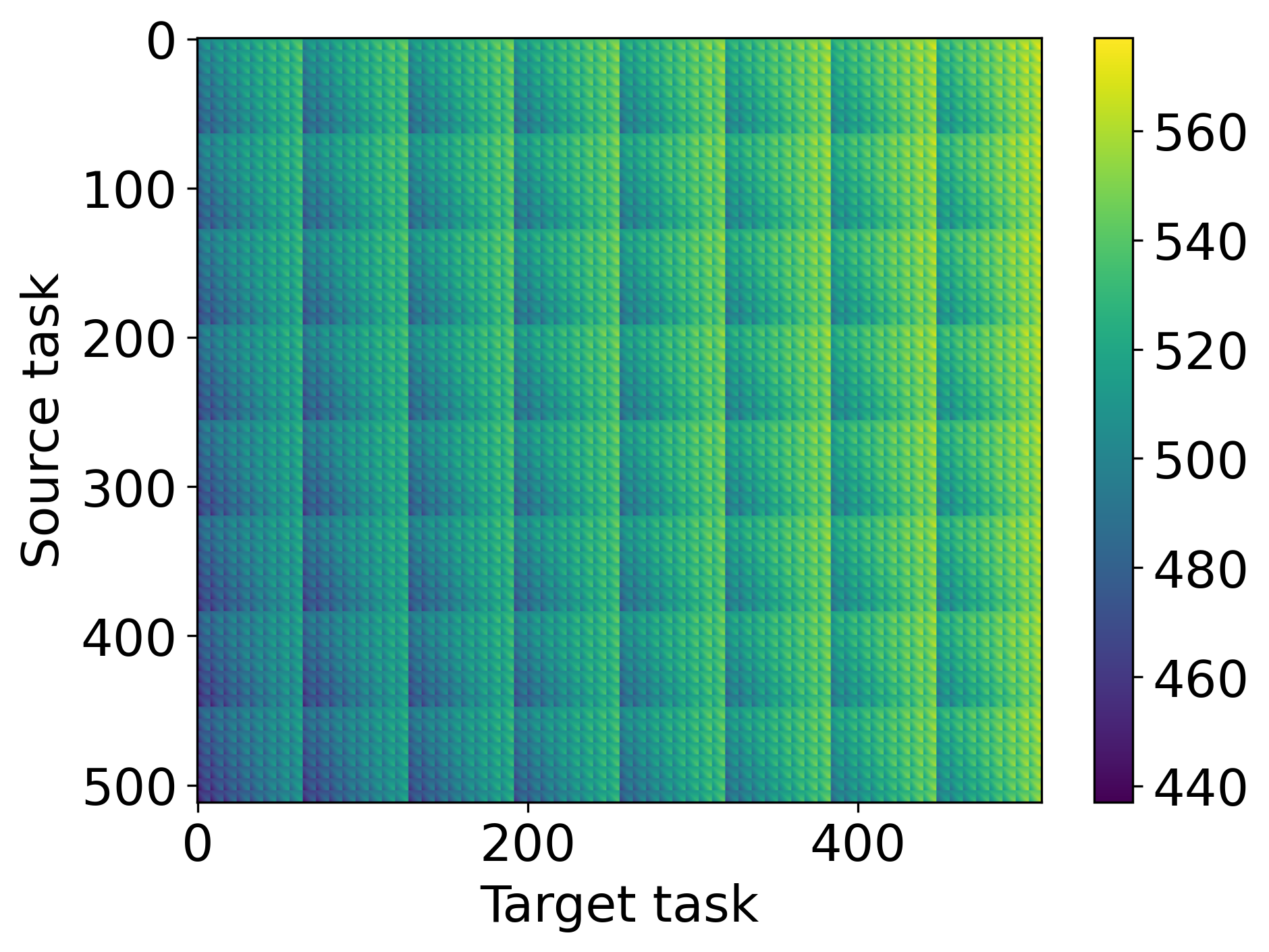}
        }%
    \hfill
    \subfigure[$\sigma=0$; $f(x)$ linear; $g(y)$ none; $h(x,y)$ L$_1$ norm]{
        \includegraphics[width=0.22\textwidth]{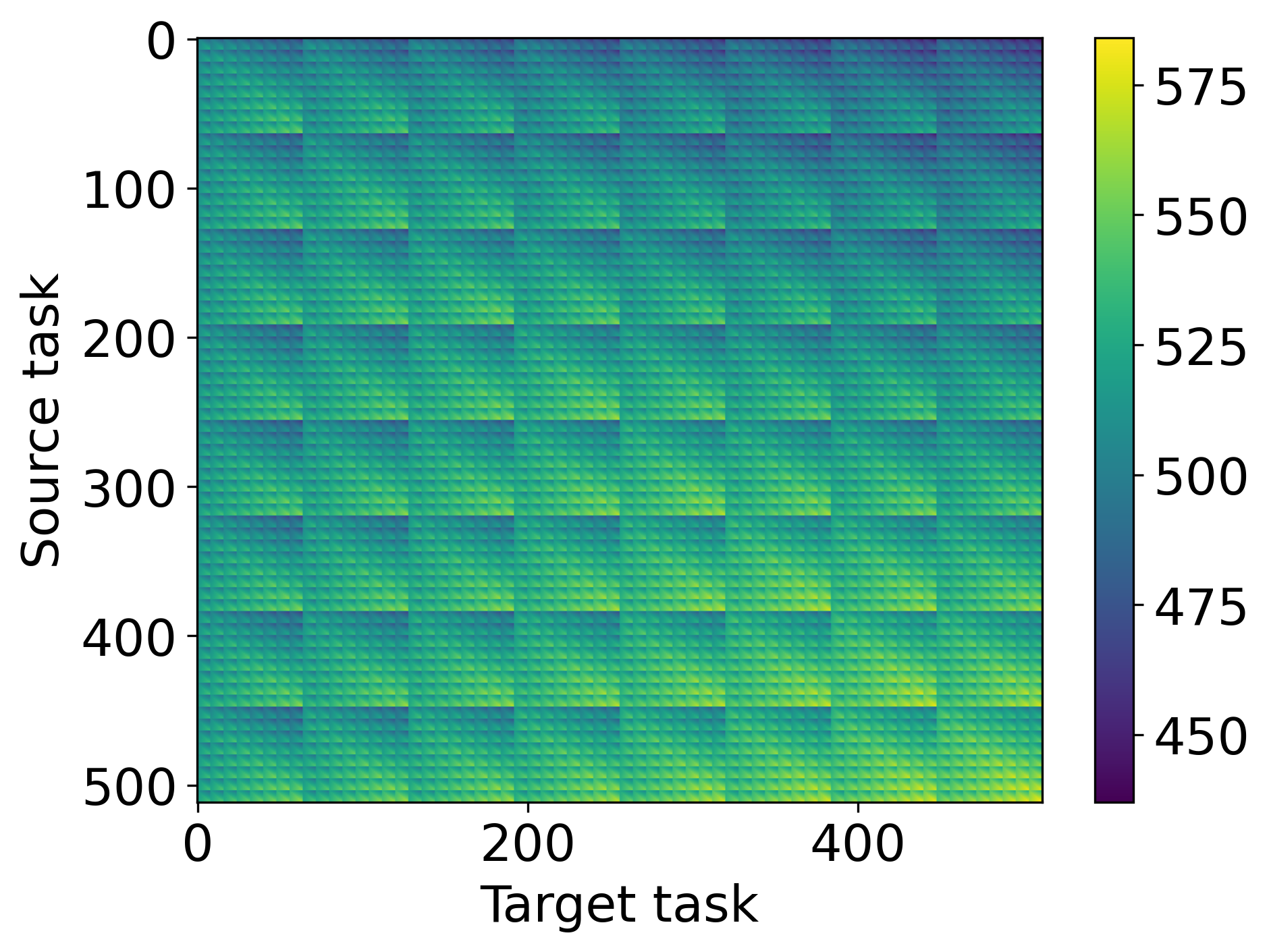}
        }%
    \hfill
    \subfigure[$\sigma=0$; $f(x)$ linear; $g(y)$ linear; $h(x,y)$ L$_1$ norm]{
        \includegraphics[width=0.22\textwidth]{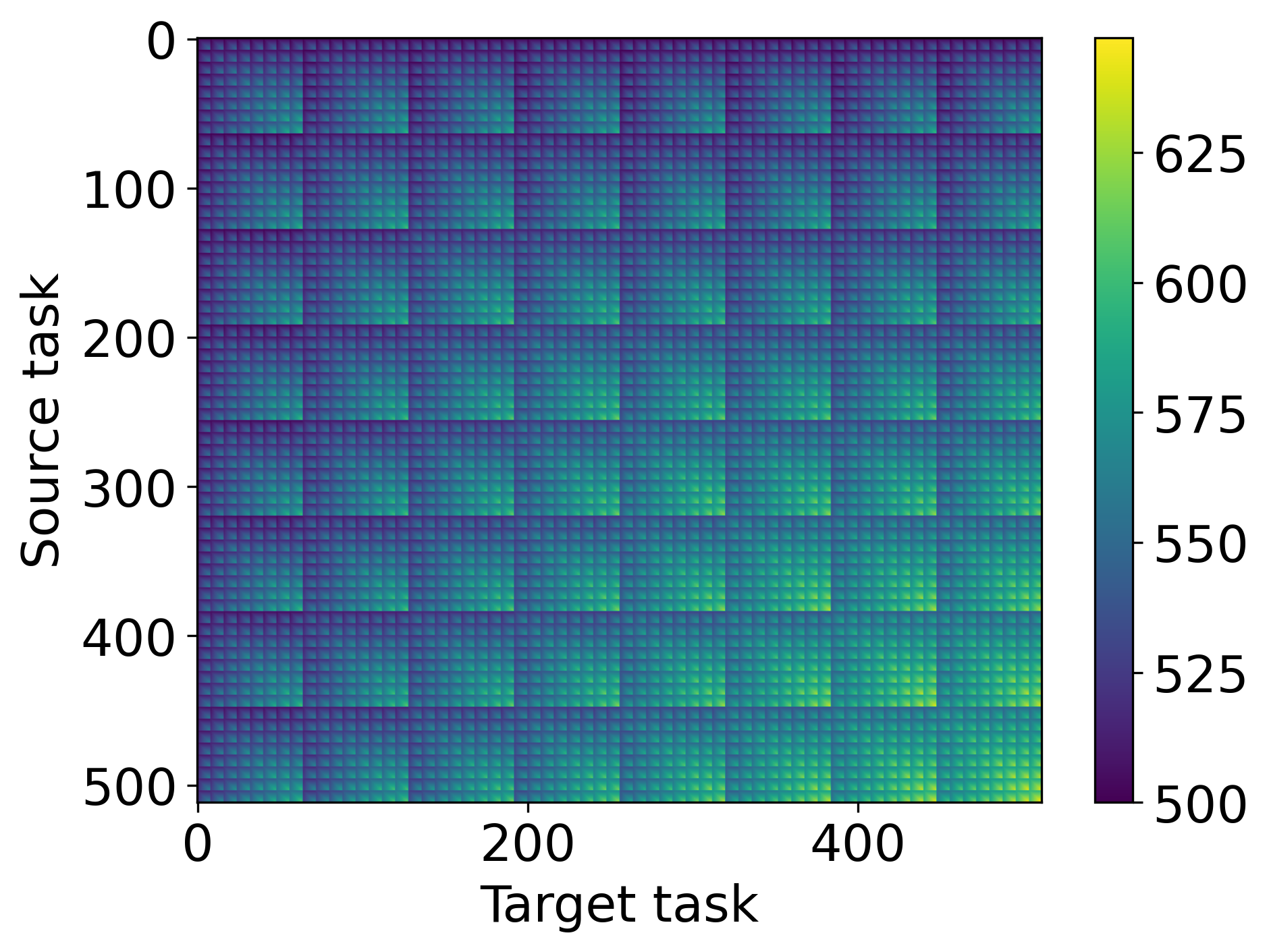}
        }%
    \\
    \subfigure[$\sigma=30$; $f(x)$ constant; $g(y)$ none; $h(x,y)$ L$_1$ norm]{
        \includegraphics[width=0.22\textwidth]{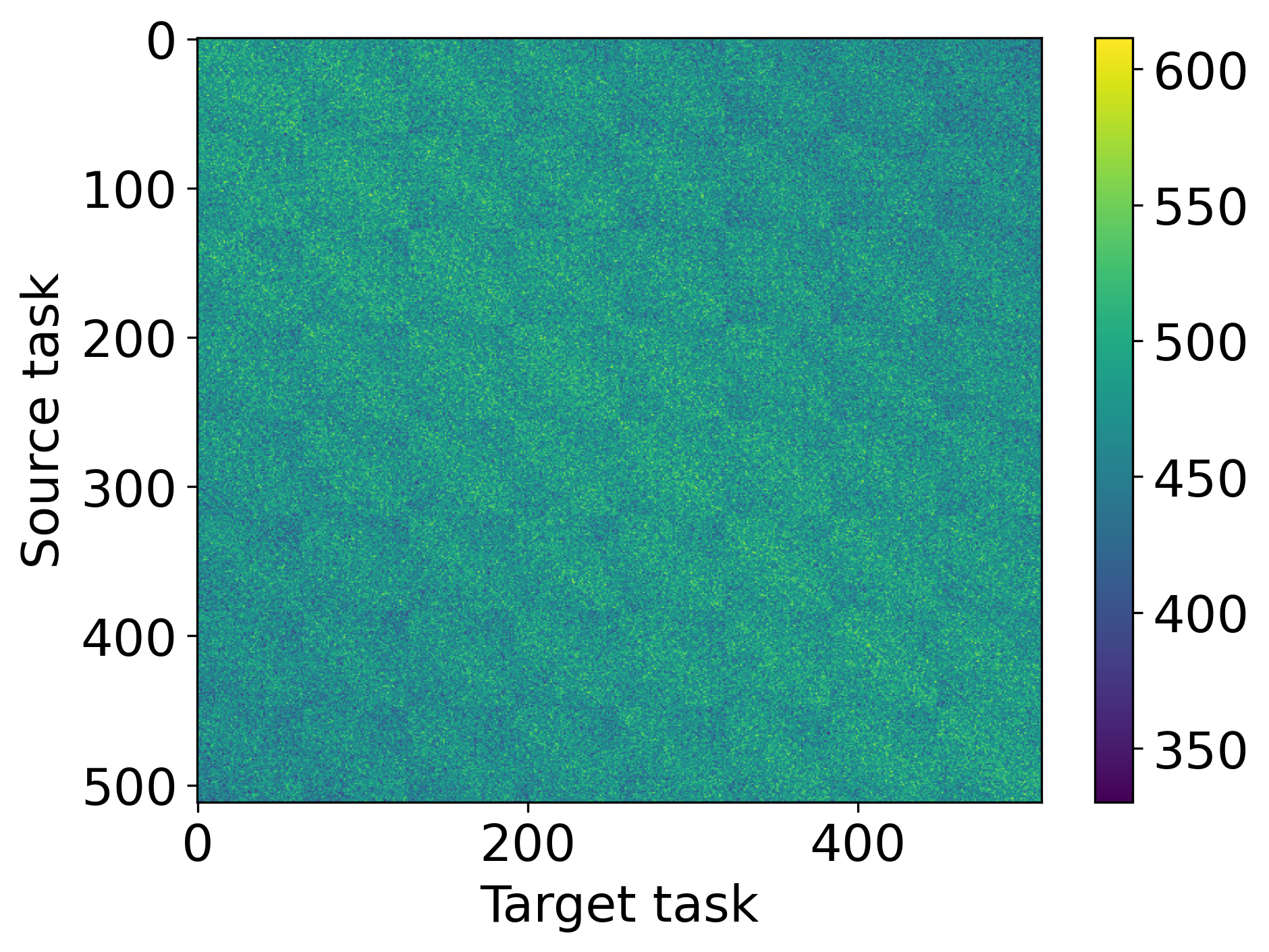}
        }%
    \hfill
    \subfigure[$\sigma=30$; $f(x)$ constant; $g(y)$ linear; $h(x,y)$ Non-distance]{
        \includegraphics[width=0.22\textwidth]{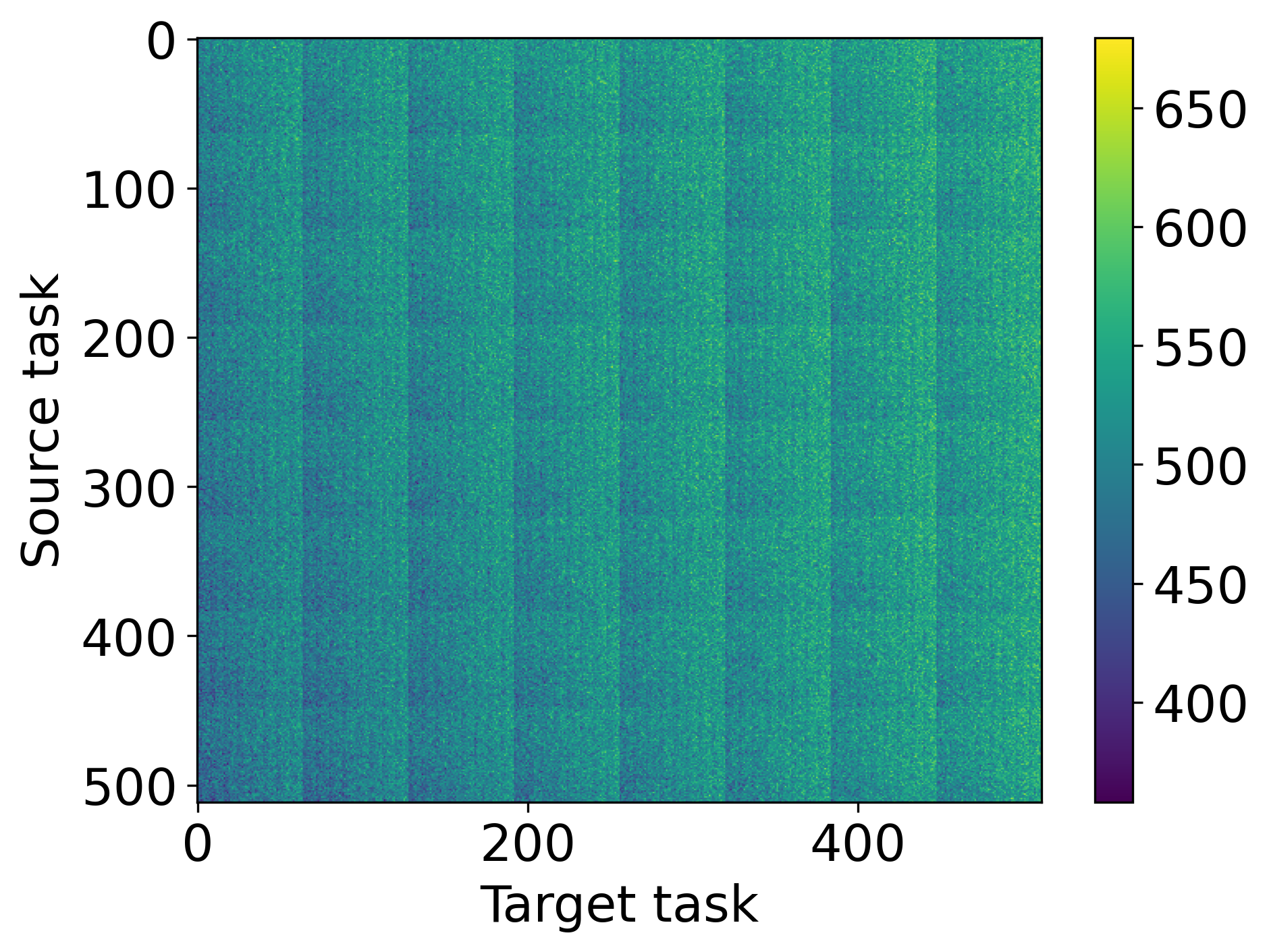}
        }%
    \hfill
    \subfigure[$\sigma=30$; $f(x)$ linear; $g(y)$ none; $h(x,y)$ L$_1$ norm]{
        \includegraphics[width=0.22\textwidth]{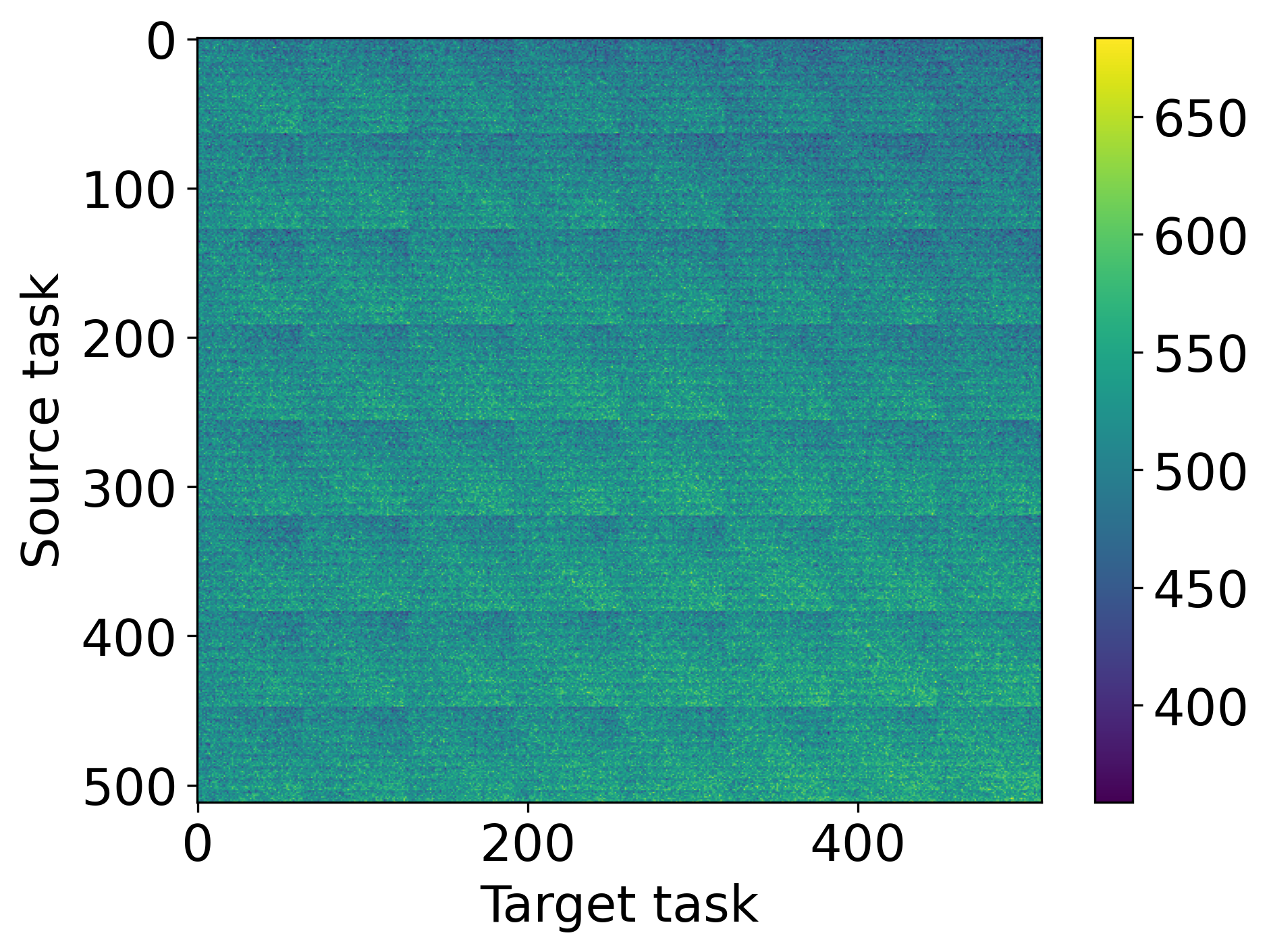}
        }%
    \hfill
    \subfigure[$\sigma=30$; $f(x)$ linear; $g(y)$ linear; $h(x,y)$ L$_1$ norm]{
        \includegraphics[width=0.22\textwidth]{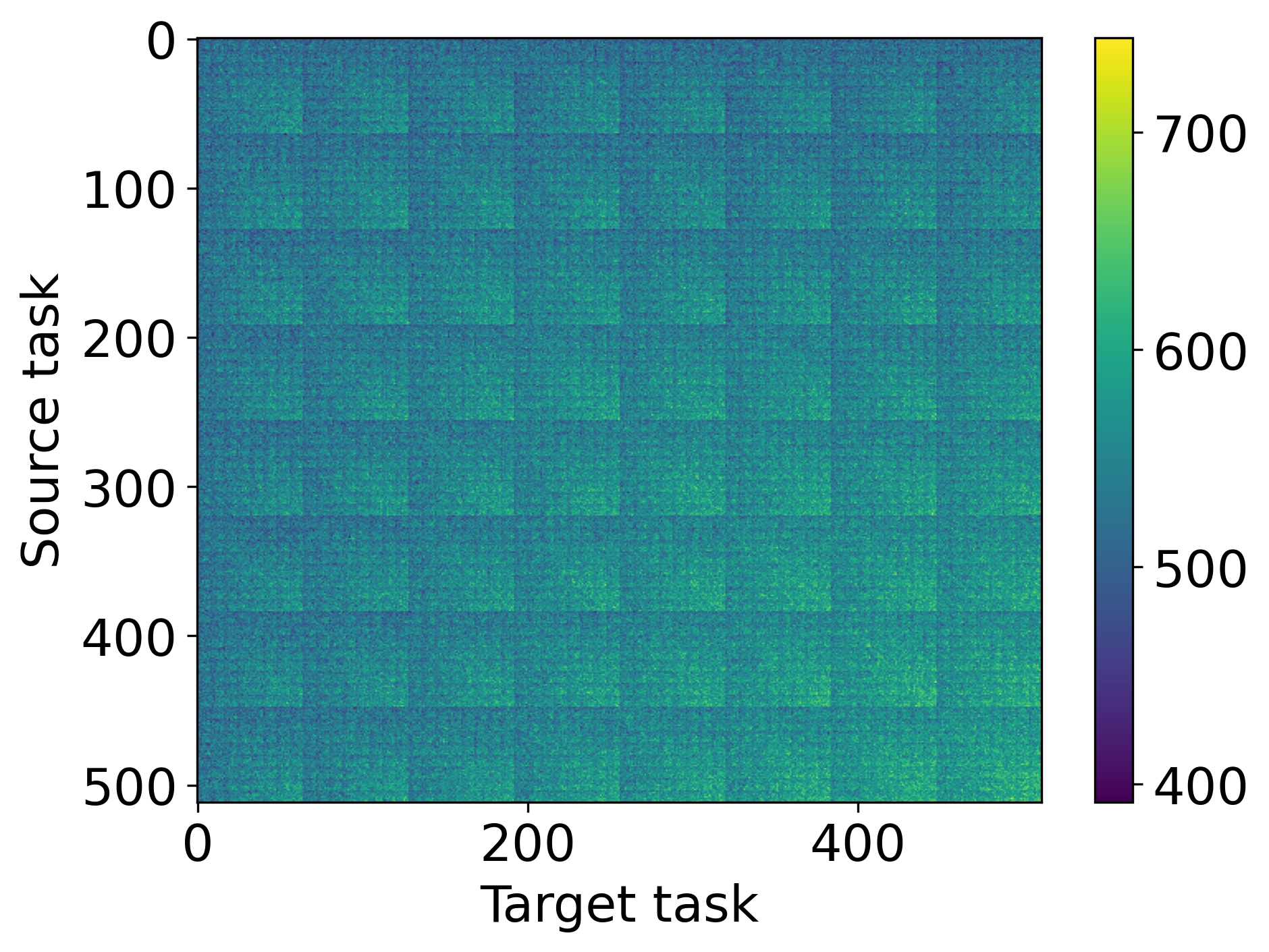}
        }%
    \caption{\jhedit{Transfer matrices of synthetic data with different configurations.
    Bright\jhedit{er} colors indicate high\jhedit{er} \jhedit{generalization} performance.}}
    \label{fig:transfer_matrix_synt}
\end{figure*}

\jhedit{\textbf{Transfer matrices.} To systematically evaluate transfer performance across controlled task variations, we generate a family of 512 "synthetic" tasks. Each task is defined by a 3-dimensional integer context vector \(x \in \{1,\dots,8\}^3\).
In Figure~\ref{fig:transfer_matrix_synt}, each subplot visualizes a 512x512 matrix whose \((i,j)\) entry is the average performance \(J(\pi_{x_i},x_j)\). Bright cells indicate high performance and dark cells indicate low performance. By comparing across panels, one can clearly see the effect of adding linear source (\(f(x)\)), target (\(g(y)\)) biases, and different interaction term $h(x,y)$ can produce structured transfer matrices, and how noise induces realistic transfer matrices.
} 

\subsection{Deep RL Training}\label{sec:DRL_training}
We employ the PPO implementation from the stable-baselines3 library \citep{stable-baselines3} for our deep reinforcement learning experiments. All training runs are executed on a distributed computing cluster featuring Intel(R) Xeon(R) CPU E5-2670 v2 @ 2.50GHz, with each independent training run utilizing four CPUs. \jhedit{Multi-task training is concurrently exposed to all target CMDPs, and the training is halted once the training convergence is achieved.}

\subsection{Hyperparameters}\label{sec:hyperparameters}
\paragraph{\tyyedit{Random Seeds}} \tyyedit{We test the algorithms with 100 bootstrapped transfer matrices. For the random selection algorithm, we take the average of 50 repeats for each transfer matrix, whereas other algorithms were repeated only once. For transfer matrix $w$ and repeat $j$, we use random seed $(w+1)\times(j+1)$.}
\paragraph{\jhedit{M-MBTL}}
M-MBTL uses the $L_1$ norm distance with slope \tyyedit{$0.01$} in each dimension. \tyyedit{We choose $L_1$ norm distance because we assume that the influence of different context dimensions is independent. Since we normalize the generalization performance to the range 0–1, choosing a fixed slope of 0.01 is a relatively moderate option. We assign the same slope to every dimension because we have no prior information about differences between dimensions; therefore, we assume that each dimension exerts an equal level of influence.} Since the number of tasks is relatively small, and the clustering process for each candidate takes a shorter time, we use the number of samples $M=N$.
% \jhfn{Add sequential clustering related hyperparameters, if any.}

\paragraph{GP-MBTL}
% We employ the \texttt{GaussianProcessRegressor} from "scikit-learn"\cw{Are we still using this?}, which implements Algorithm~2.1 from \cite{williams_gaussian_2006}. 
% Our kernel is formed as the product of a constant kernel and a radial basis function (RBF) kernel. We use the hyperparameters from \cite{MBTL}, a noise standard deviation of $\sigma = 0.001$ and random restarts of the hyperparameter optimizer of $15$.
% This set of hyperparameters is applied consistently throughout all experiments and benchmarks.
\tyyedit{GP-MBTL also uses $L_1$ norm distance with \jhedit{slope} \tyyedit{$0.01$} in each dimension. We used \texttt{gpytorch.kernels.RBFKernel} in \cite{gpytorch2018} with default hyperparameters as the kernel function for the Gaussian Process.}

\paragraph{RL Training}
Table~\ref{tab:benchmark_hyperparams} lists the RL‐training hyperparameters for CartPole, BipedalWalker, IntersectionZoo, and CyclesGym benchmarks. For CartPole and BipedalWalker, we employ PPO from \texttt{stable\_baseline3} \cite{stable-baselines3} with default hyperparameters, including a learning rate of $3\times 10^-4$, \texttt{n\_steps}$=2048$, batch size $64$, discount factor $0.99$, GAE parameter $0.95$, clipping parameter $0.2$, entropy coefficient $0$, and a value function loss coefficient $0.5$.

For IntersectionZoo, we configure PPO with a policy clip of $0.03$ and began with a KL-divergence penalty coefficient of $0.1$, aiming to maintain KL-divergence around $0.02$ throughout training. The value network employ a clipping threshold of $\pm$3 and carried a loss weight of $1$. To foster exploration, we add an entropy bonus weighted at $0.005$. Each epoch comprises $10$ gradient‐update steps, and the model is trained for $5,000$ epochs, running $10$ episodes per epoch. Episodes last up to $1,500$ time steps, and we sample mini-batches of $40$ transitions. Optimization is handled by Adam with a learning rate of $1e-4$, weight decay of 0.97, and $\beta$-parameters $(0.9, 0.999)$. Our policy and value networks each feature four hidden layers of $256$ units, use tanh activations, and are initialized orthogonally. The simulator warm up for 50 steps before each episode, proceed with a 0.5-second step size, and we discount future rewards by $\gamma=0.99$.

For the CyclesGym experiments, we also run PPO for a total of $20,000$ time steps, evaluating performance every $1,000$ steps. Each update collects $80$ environment steps per rollout, with mini-batches of $64$ samples and $10$ optimization epochs per update. For the others, we use the default parameters from \texttt{stable\_baseline3} \cite{stable-baselines3}.

\begin{table*}[htbp]
    \centering
    \caption{RL‐training hyperparameters for CartPole, BipedalWalker, IntersectionZoo, and CyclesGym benchmarks.}
    \label{tab:benchmark_hyperparams}
    \begin{tabular}{lcccc}
        \toprule
        \textbf{Hyperparameter} & \textbf{CartPole} & \textbf{BipedalWalker} &  \textbf{IntersectionZoo} &\textbf{CyclesGym} \\
        \midrule
        Training steps (independent) & $5\times 10^6$ & $5\times 10^6$ & $5\times 10^6$ & $2\times 10^3$ \\
        Training steps (multi-task)   & $5\times 10^8$ & $7\times 10^8$ & $5\times10^8$ & $4\times 10^4$ \\
        Test episodes (train)                & $10$ & $10$ & $10$ & $10$  \\
        Evaluation episodes (transfer)   & $100$ & $100$ & $100$ & $100$ \\
        \bottomrule
    \end{tabular}
\end{table*}
\section{Results in Detail}
\subsection{\Tyedit{Synthetic Data with Different Noise Terms}}\label{sec:diff_noise_term}
\Tyedit{We also test our approaches on synthetic data with different noise terms.
Table~\ref{tab:synthetic-performance-noise0} presents the performance comparison for \( \epsilon = 0 \), showing results similar to those in Table \ref{tab:synthetic-performance}.
Table~\ref{tab:synthetic-performance-noise30} presents the performance comparison for \( \epsilon \sim \mathcal{N}(0, 30^2) \). With significantly increased noise, Assumptions~\ref{assump:constant} and \ref{assump:distance} are violated to some extent in all cases, causing the CMDP structure to deviate from Mountain. Since GP-MBTL uses real generalization performance to compute the acquisition function, it demonstrates greater robustness to noise.  
GP-MBTL generally performs better when \( g(y) \) is constant, whereas M-MBTL is more effective when \( g(y) \) is linear. M/GP-MBTL consistently achieves the highest or near-highest performance, highlighting the effectiveness of structure detection in efficiently solving CMDPs with varying structural properties.}

\begin{table*}[!ht]
\begin{small}
\begin{center}
  \caption{Performance comparison on Synthetic Data ($\epsilon=0$) with K=50.}
  \label{tab:synthetic-performance-noise0}
  \begin{tabular}{lllcccc|c}
    \toprule
    \textbf{$f(x)$} & \textbf{$g(y)$} & \textbf{$h(x,y)$}  & \textbf{Random} & \textbf{GP-MBTL} & \textbf{M-MBTL (Ours)} & \textbf{M/GP-MBTL (Ours)} & \textbf{Myopic Oracle}\\
    \midrule
    Constant & Linear & Non-distance & {0.7028} $\pm$ 0.0010 & \textbf{0.7250} $\pm$ 0.0000 & {0.7055} $\pm$ 0.0000 & \textbf{0.7250} $\pm$ 0.0000 & {0.7250} $\pm$ 0.0000\\
    \rowcolor{mountainrow} Constant & Linear & $L_1$ norm & {0.7122} $\pm$ 0.0002 & {0.7110} $\pm$ 0.0000 & \textbf{0.7187} $\pm$ 0.0000 & \textbf{0.7187} $\pm$ 0.0000 & {0.7190} $\pm$ 0.0000\\
    Constant & None & Non-distance & {0.8688} $\pm$ 0.0019 & \textbf{0.9091} $\pm$ 0.0000 & {0.8736} $\pm$ 0.0000 & \textbf{0.9091} $\pm$ 0.0000 & {0.9091} $\pm$ 0.0000\\
    \rowcolor{mountainrow} Constant & None & $L_1$ norm & {0.9244} $\pm$ 0.0005 & {0.9364} $\pm$ 0.0000 & \textbf{0.9374} $\pm$ 0.0000 & \textbf{0.9374} $\pm$ 0.0000 & {0.9382} $\pm$ 0.0000\\
    Linear & Linear & Non-distance & {0.5544} $\pm$ 0.0022 & \textbf{0.5714} $\pm$ 0.0000 & {0.5552} $\pm$ 0.0000 & \textbf{0.5714} $\pm$ 0.0000 & {0.5714} $\pm$ 0.0000\\
    Linear & Linear & $L_1$ norm & {0.5517} $\pm$ 0.0027 & \textbf{0.5714} $\pm$ 0.0000 & {0.5535} $\pm$ 0.0000 & \textbf{0.5714} $\pm$ 0.0000 & {0.5714} $\pm$ 0.0000\\
    Linear & None & Non-distance & {0.6761} $\pm$ 0.0032 & \textbf{0.7000} $\pm$ 0.0000 & {0.6772} $\pm$ 0.0000 & \textbf{0.7000} $\pm$ 0.0000 & {0.7000} $\pm$ 0.0000\\
    Linear & None & $L_1$ norm & {0.7660} $\pm$ 0.0028 & \textbf{0.7857} $\pm$ 0.0000 & {0.7678} $\pm$ 0.0000 & \textbf{0.7857} $\pm$ 0.0000 & {0.7857} $\pm$ 0.0000\\
\midrule
\multicolumn{3}{c}{\textbf{Aggregated Performance}} & {-0.0000} $\pm$ 0.0353 & {0.8368} $\pm$ 0.0268 & {0.3010} $\pm$ 0.0264 & \textbf{0.9873} $\pm$ 0.0015 & {1.0000} $\pm$ 0.0000 \\
    \bottomrule
  \end{tabular}
  \end{center}
  \scriptsize{* \textit{Note}: Values are reported as the mean performance ± half the width of the 95\% confidence interval. Rows shaded in light gray indicate the synthetic settings that satisfy our proposed \textsc{Mountain} structure assumption. Bold values represent the highest value(s) within the statistically significant range for each task, excluding the oracle.}
  \end{small}
\end{table*}

\begin{table*}[!ht]
\begin{small}
\begin{center}
  \caption{Performance comparison on Synthetic Data ($\epsilon=\mathcal{N}(0, 30^2)$) with K=50.}
  \label{tab:synthetic-performance-noise30}
  \begin{tabular}{lllcccc|c}
    \toprule
    \textbf{$f(x)$} & \textbf{$g(y)$} & \textbf{$h(x,y)$}  & \textbf{Random} & \textbf{GP-MBTL} & \textbf{M-MBTL (Ours)} & \textbf{M/GP-MBTL (Ours)} & \textbf{Myopic Oracle}\\
    \midrule
    Constant & Linear & Non-distance & {0.7083} $\pm$ 0.0005 & \textbf{0.7274} $\pm$ 0.0008 & {0.7091} $\pm$ 0.0003 & {0.7202} $\pm$ 0.0012 & {0.7319} $\pm$ 0.0003\\
    \rowcolor{mountainrow} Constant & Linear & $L_1$ norm & \textbf{0.6879} $\pm$ 0.0003 & {0.6825} $\pm$ 0.0008 & \textbf{0.6879} $\pm$ 0.0003 & \textbf{0.6876} $\pm$ 0.0005 & {0.6952} $\pm$ 0.0003\\
    Constant & None & Non-distance & {0.7036} $\pm$ 0.0006 & \textbf{0.7274} $\pm$ 0.0006 & {0.7045} $\pm$ 0.0004 & {0.7215} $\pm$ 0.0012 & {0.7295} $\pm$ 0.0003\\
    \rowcolor{mountainrow} Constant & None & $L_1$ norm & {0.7125} $\pm$ 0.0004 & \textbf{0.7181} $\pm$ 0.0006 & {0.7125} $\pm$ 0.0004 & {0.7150} $\pm$ 0.0006 & {0.7205} $\pm$ 0.0004\\
    Linear & Linear & Non-distance & {0.6815} $\pm$ 0.0008 & {0.6790} $\pm$ 0.0044 & {0.6818} $\pm$ 0.0003 & \textbf{0.6912} $\pm$ 0.0024 & {0.7135} $\pm$ 0.0003\\
    Linear & Linear & $L_1$ norm & {0.6840} $\pm$ 0.0010 & {0.6810} $\pm$ 0.0062 & {0.6844} $\pm$ 0.0003 & \textbf{0.6953} $\pm$ 0.0037 & {0.7234} $\pm$ 0.0003\\
    Linear & None & Non-distance & \textbf{0.7240} $\pm$ 0.0008 & {0.7202} $\pm$ 0.0035 & {0.7244} $\pm$ 0.0003 & \textbf{0.7272} $\pm$ 0.0024 & {0.7586} $\pm$ 0.0003\\
    Linear & None & $L_1$ norm & {0.7246} $\pm$ 0.0010 & \textbf{0.7366} $\pm$ 0.0028 & {0.7250} $\pm$ 0.0003 & \textbf{0.7393} $\pm$ 0.0023 & {0.7670} $\pm$ 0.0003\\
\midrule
\multicolumn{3}{c}{\textbf{Aggregated Performance}} & {-0.0000} $\pm$ 0.0108 & {0.2127} $\pm$ 0.0514 & {0.0147} $\pm$ 0.0089 & \textbf{0.3099} $\pm$ 0.0281 & {1.0000} $\pm$ 0.0082 \\
    \bottomrule
  \end{tabular}
  \end{center}
  \scriptsize{* \textit{Note}: Values are reported as the mean performance ± half the width of the 95\% confidence interval. Rows shaded in light gray indicate the synthetic settings that satisfy our proposed \textsc{Mountain} structure assumption. Bold values represent the highest value(s) within the statistically significant range for each task, excluding the oracle.}
  \end{small}
\end{table*}

\begin{figure*}[ht]
  \centering
  \includegraphics[width=0.9\linewidth]{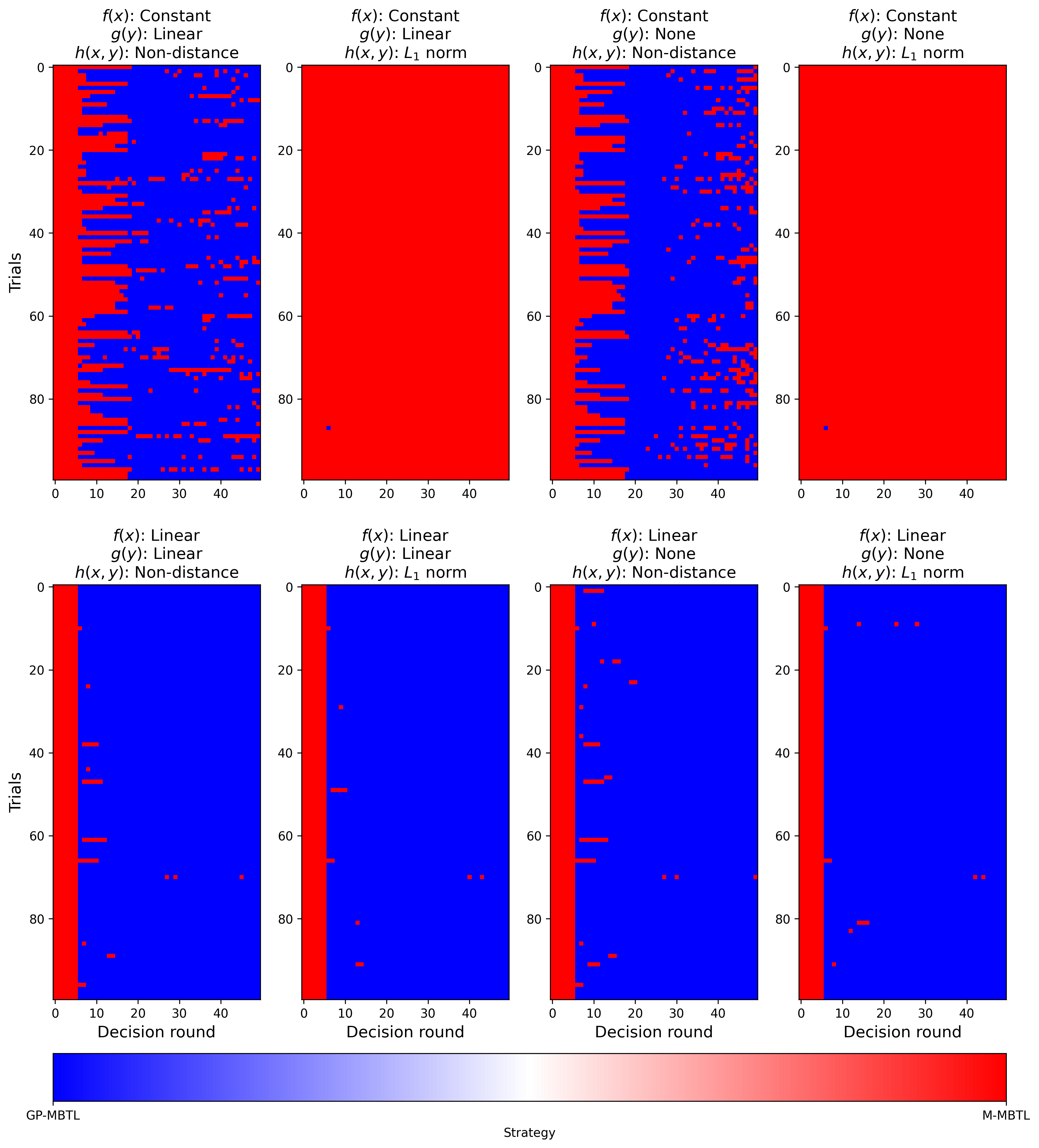}
  \caption{\jhedit{Per-round algorithm choice over 100 trials ($K=50$) for eight synthetic CMDP configurations with $\epsilon\sim\mathcal{N}(0,5^2)$. \textcolor{red}{Red} = M-MBTL; \textcolor{blue}{blue} = GP-MBTL. Only the data with \textsc{Mountain} structure (constant $f(x)$, $L_1$-norm $h(x,y)$) consistently trigger M-MBTL; all others default to GP-MBTL.}}
  \label{fig:algswitch_synt}
\end{figure*}

\begin{figure*}[!t]%
    \centering
    \subfigure[$f(x)$ constant; $g(y)$ none; $h(x,y)$ Non-distance]{%
        \label{fig:synt_g_noise5_x_weightNone_y_weightNone_dist_left11-3_dist_right333}%
        \includegraphics[width=0.4\textwidth]{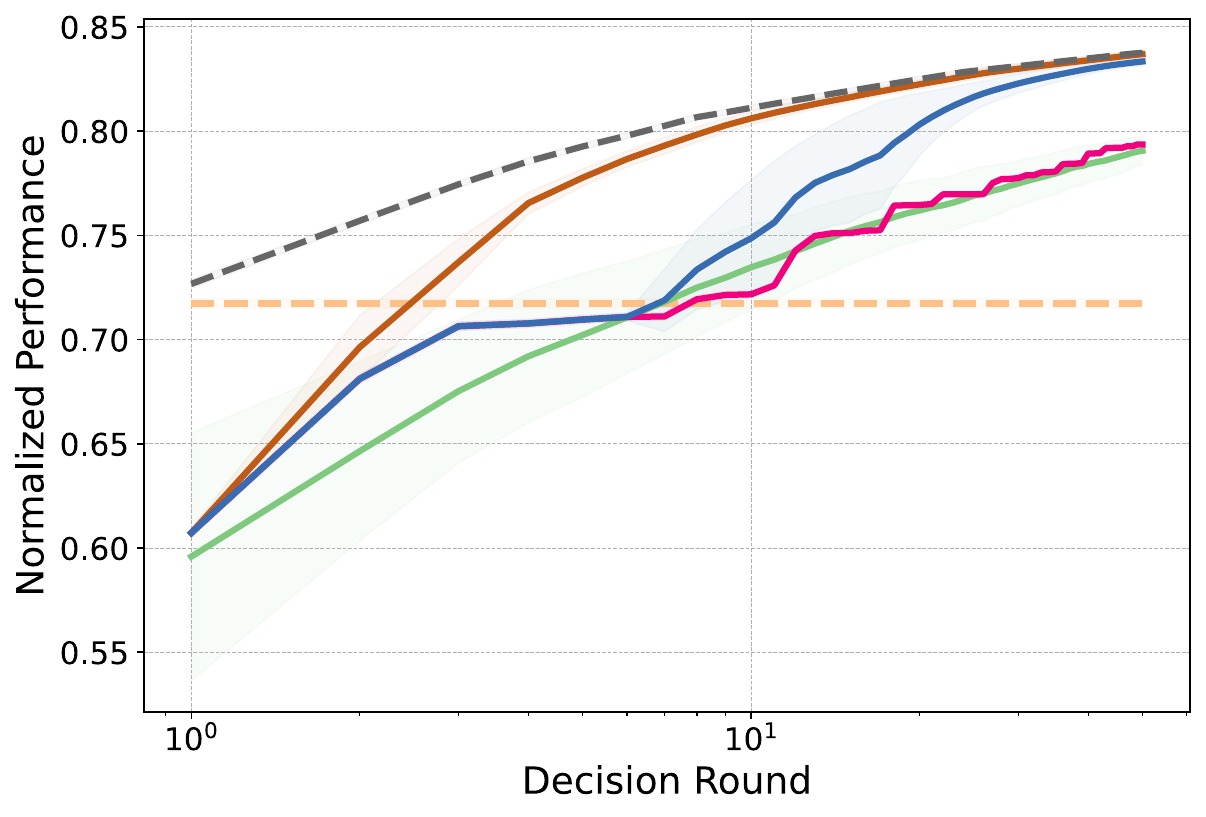}
        }%
    \subfigure[$f(x)$ constant; $g(y)$ none; $h(x,y)$ L$_1$ norm]{
        \label{fig:synt_g_noise5_x_weightNone_y_weightNone_dist_left-3-3-3_dist_right333}
        \includegraphics[width=0.4\textwidth]{images/result_synt_g_noise5_x_weightNone_y_weightNone_dist_left11-3_dist_right333_K50.pdf}
        }%
    \\
    \subfigure[$f(x)$ constant; $g(y)$ linear; $h(x,y)$ Non-distance]{%
        \label{fig:synt_g_noise5_x_weightNone_y_weightLinear_dist_left11-3_dist_right333}%
        \includegraphics[width=0.4\textwidth]{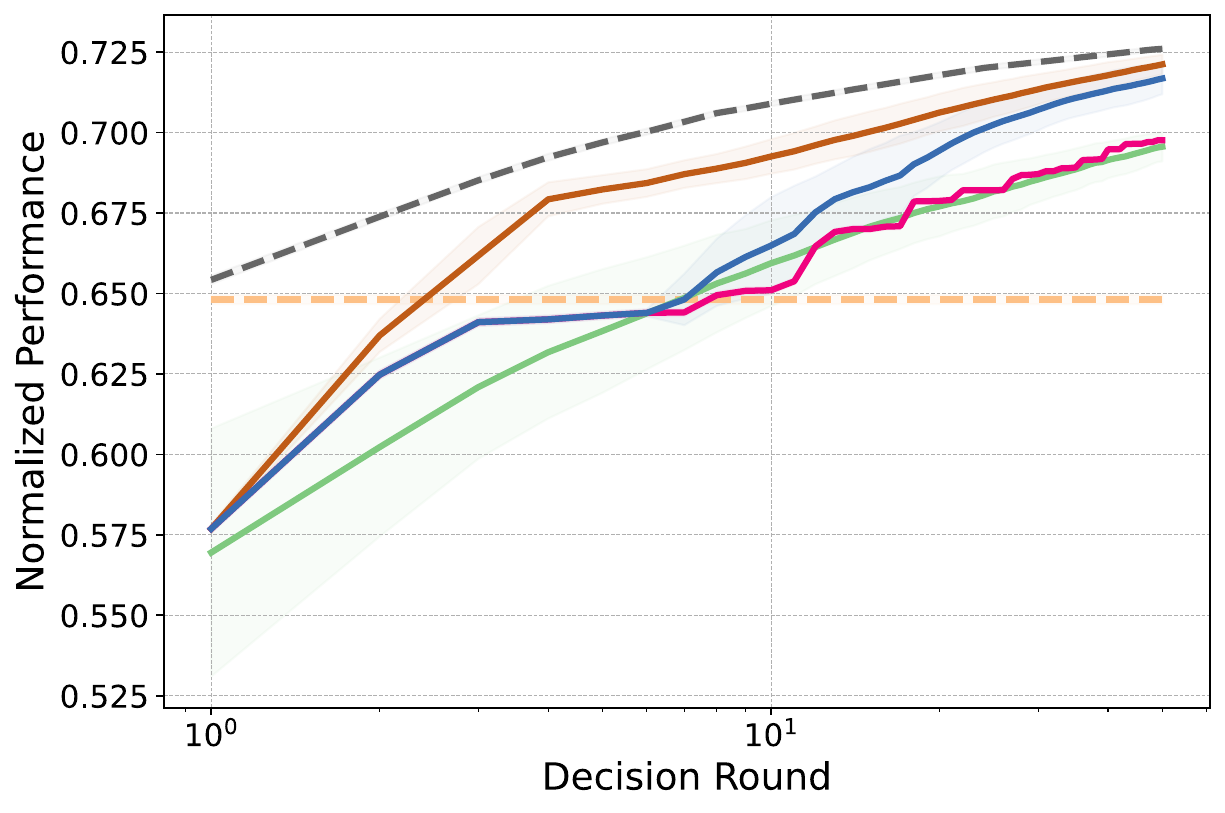}
        }%
    \subfigure[$f(x)$ constant; $g(y)$ linear; $h(x,y)$ L$_1$ norm]{
        \label{fig:synt_g_noise5_x_weightNone_y_weightLinear_dist_left-3-3-3_dist_right333}
        \includegraphics[width=0.4\textwidth]{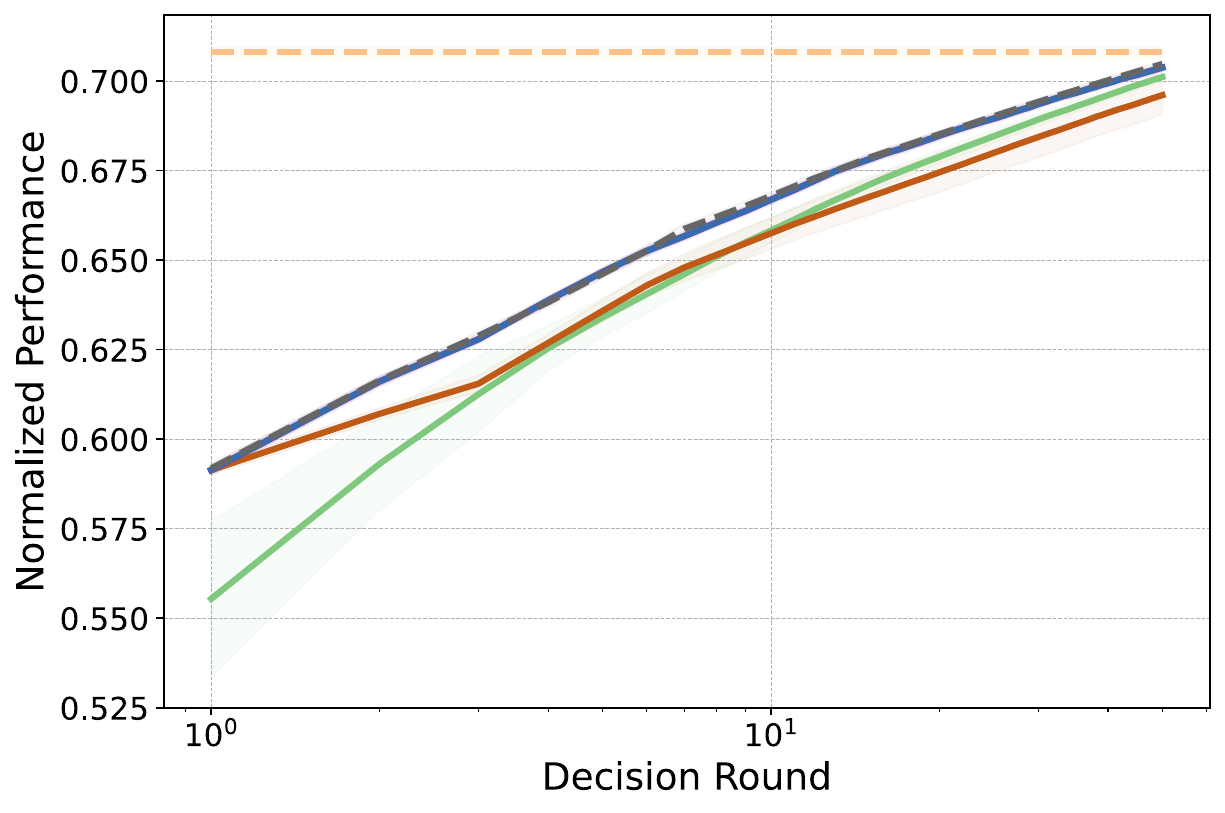}
        }%
    \\
    \subfigure[$f(x)$ linear; $g(y)$ none; $h(x,y)$ Non-distance]{%
        \label{fig:synt_g_noise5_x_weightlinear_y_weightNone_dist_left11-3_dist_right333}%
        \includegraphics[width=0.4\textwidth]{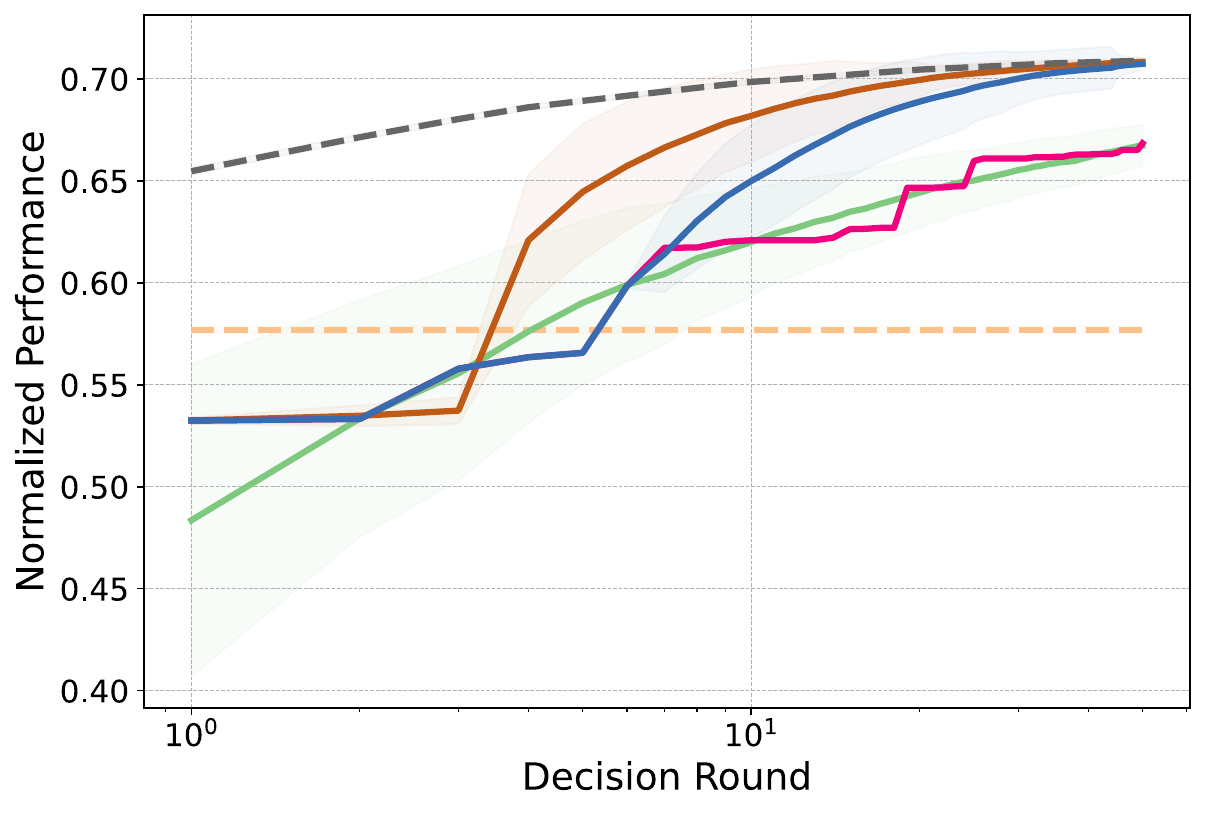}
        }%
    \subfigure[$f(x)$ linear; $g(y)$ none; $h(x,y)$ L$_1$ norm]{
        \label{fig:synt_g_noise5_x_weightlinear_y_weightNone_dist_left-3-3-3_dist_right333}
        \includegraphics[width=0.4\textwidth]{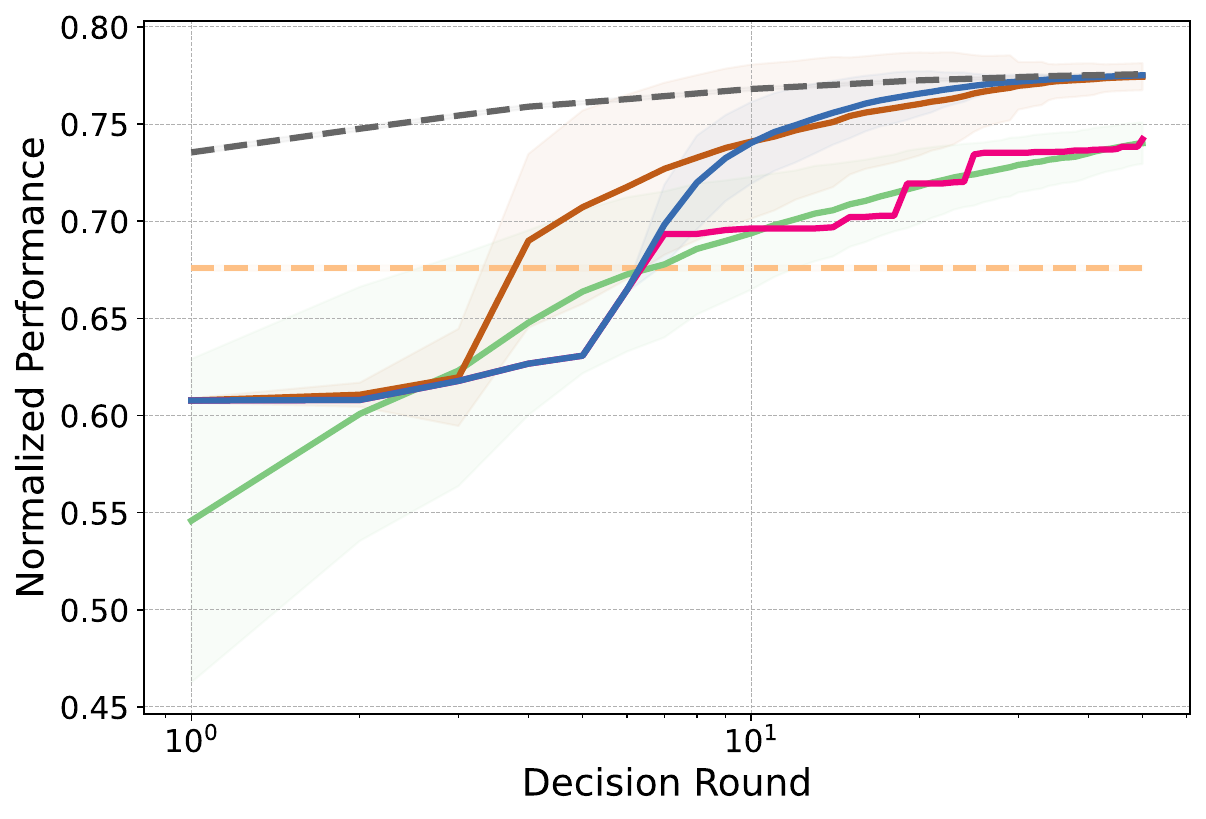}
        }%
    \\
    \subfigure[$f(x)$ linear; $g(y)$ linear; $h(x,y)$ Non-distance]{%
        \label{fig:synt_g_noise5_x_weightlinear_y_weightLinear_dist_left11-3_dist_right333}%
        \includegraphics[width=0.4\textwidth]{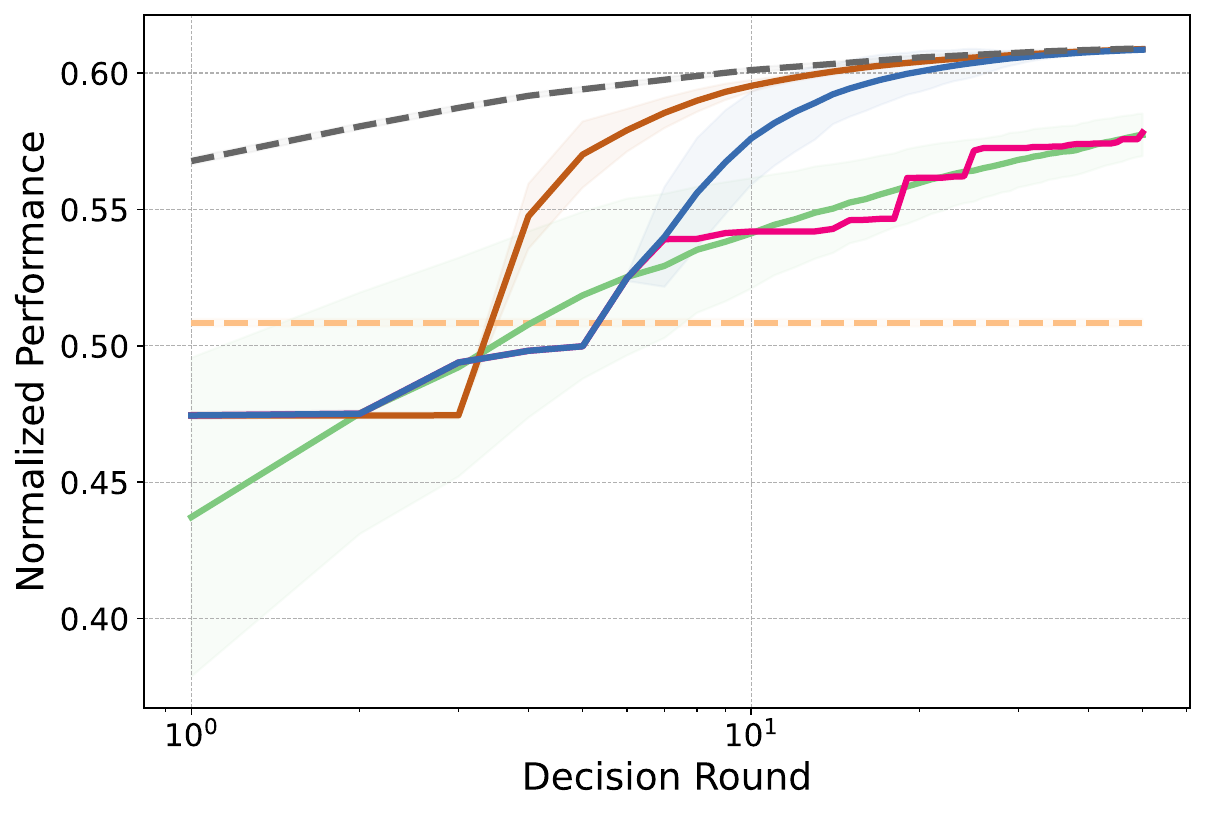}
        }%
    \subfigure[$f(x)$ linear; $g(y)$ linear; $h(x,y)$ L$_1$ norm]{
        \label{fig:synt_g_noise5_x_weightlinear_y_weightLinear_dist_left-3-3-3_dist_right333}
        \includegraphics[width=0.4\textwidth]{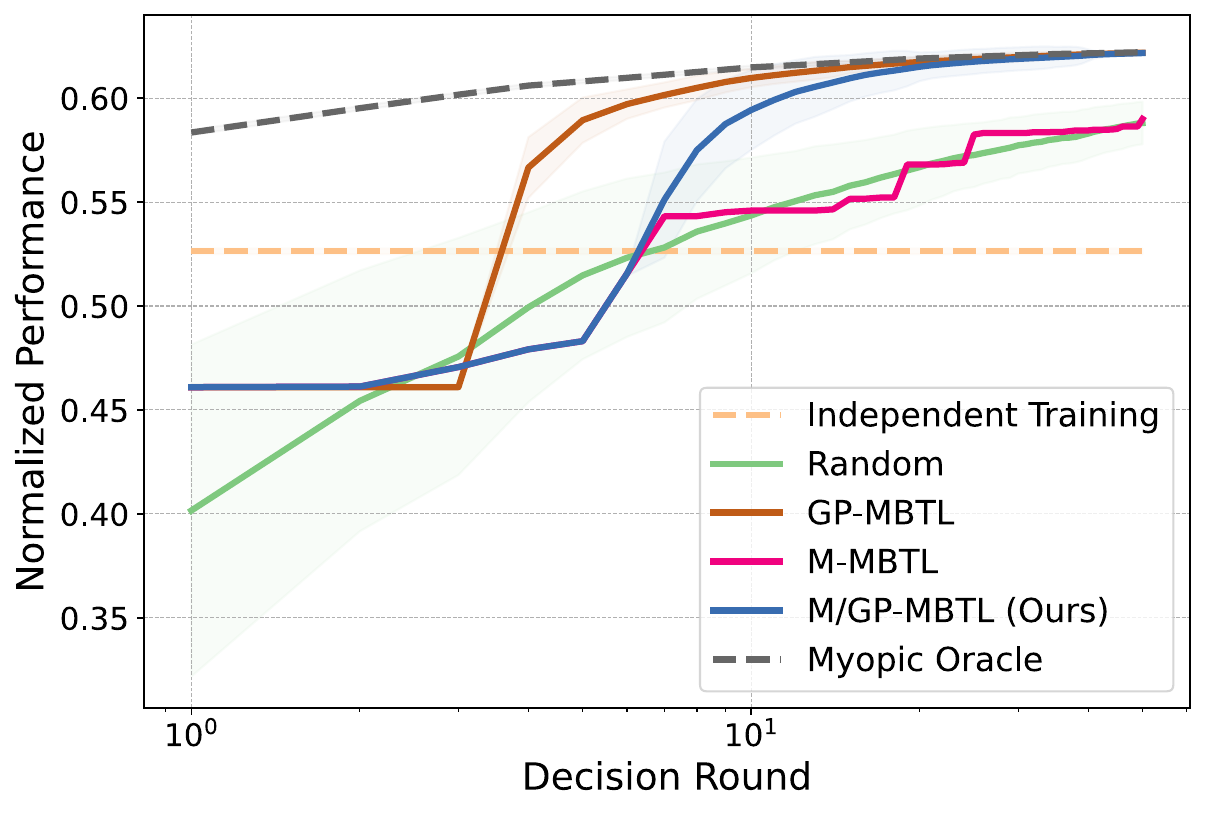}
        }%
    \\
    \caption{\jhedit{Normalized performance comparison} of \jhedit{M/GP}-MBTL \jhedit{with} other baselines (Independent Training, \jhedit{Stochastic} Oracle, Random, and MBTL baselines) \jhedit{on} the \jhedit{synthetic data ($\epsilon\sim\mathcal{N}(0, 5^2)$)) with $K=50$}.
    }
    \label{fig:result_synthetic}
\end{figure*}

\jhedit{Figure~\ref{fig:algswitch_synt} shows, for each of the eight synthetic CMDP configurations (varying $f(x)$, $g(y)$, and $h(x,y)$), which sub-algorithm was chosen over 100 trials and 50 rounds ($K=50$). Red entries indicate rounds where M-MBTL was deployed; blue where GP-MBTL was deployed. In the two \textsc{Mountain} settings (constant $f$, $L_1$-norm $h$, regardless of $g$), M-MBTL dominates almost every round. In all non-mountain settings (either non-distance $h$ or non-constant $f$), GP-MBTL is usually selected. This confirms that our online detector correctly identifies the underlying structure and switches to the appropriate task-selection strategy in each case.}
\jhedit{Figure~\ref{fig:result_synthetic} also shows the progression of the normalized performance of our proposed M/GP-MBTL algorithm compared with various baselines in terms of decision rounds. M/GP-MBTL achieves the best---or nearly the best---performance, successfully detecting whether a CMDP satisfies \textsc{Mountain} structure and choosing the appropriate algorithm accordingly.}

\begin{figure*}
    \centering
\includegraphics[width=0.6\textwidth]{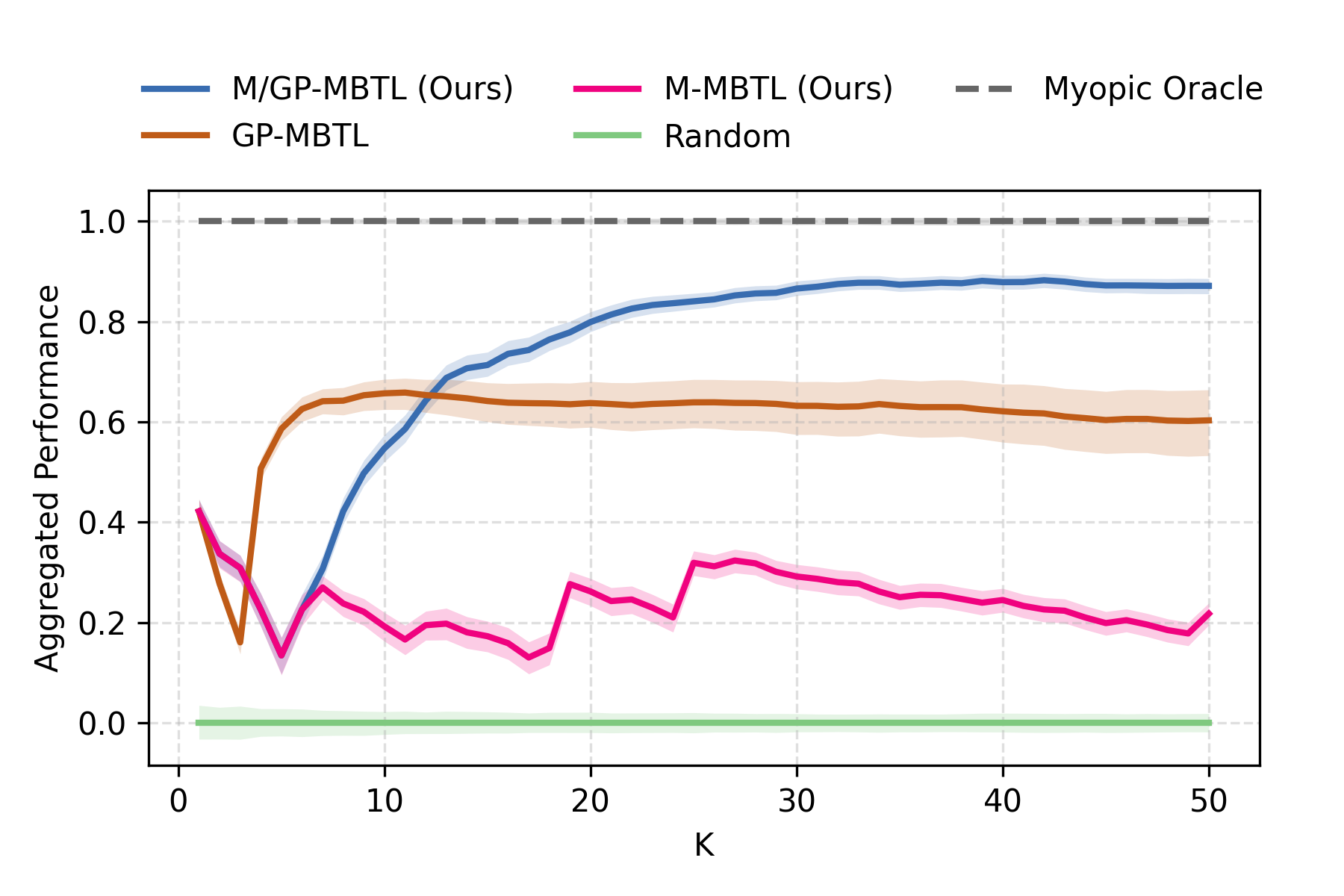}
\caption{The plot of the aggregated metric versus (K) on the 3D synthetic dataset.}
    \label{fig:any_K}
\end{figure*}
Figure~\ref{fig:any_K} shows how the aggregated metric varies with K on the 3D synthetic dataset. As K increases, the performance of M/GP-MBTL improves rapidly, surpassing GP-MBTL at around (K = 12), and then continues to maintain the best performance thereafter.

\subsection{\Tyedit{CMDP Benchmarks}}\label{sec:IQM}

\begin{figure*}[!t]
    \setcounter{subfigure}{0}
    \centering
    \subfigure[CartPole]{
        \label{fig:epsilon_cartpole}
        \includegraphics[width=0.45\textwidth]{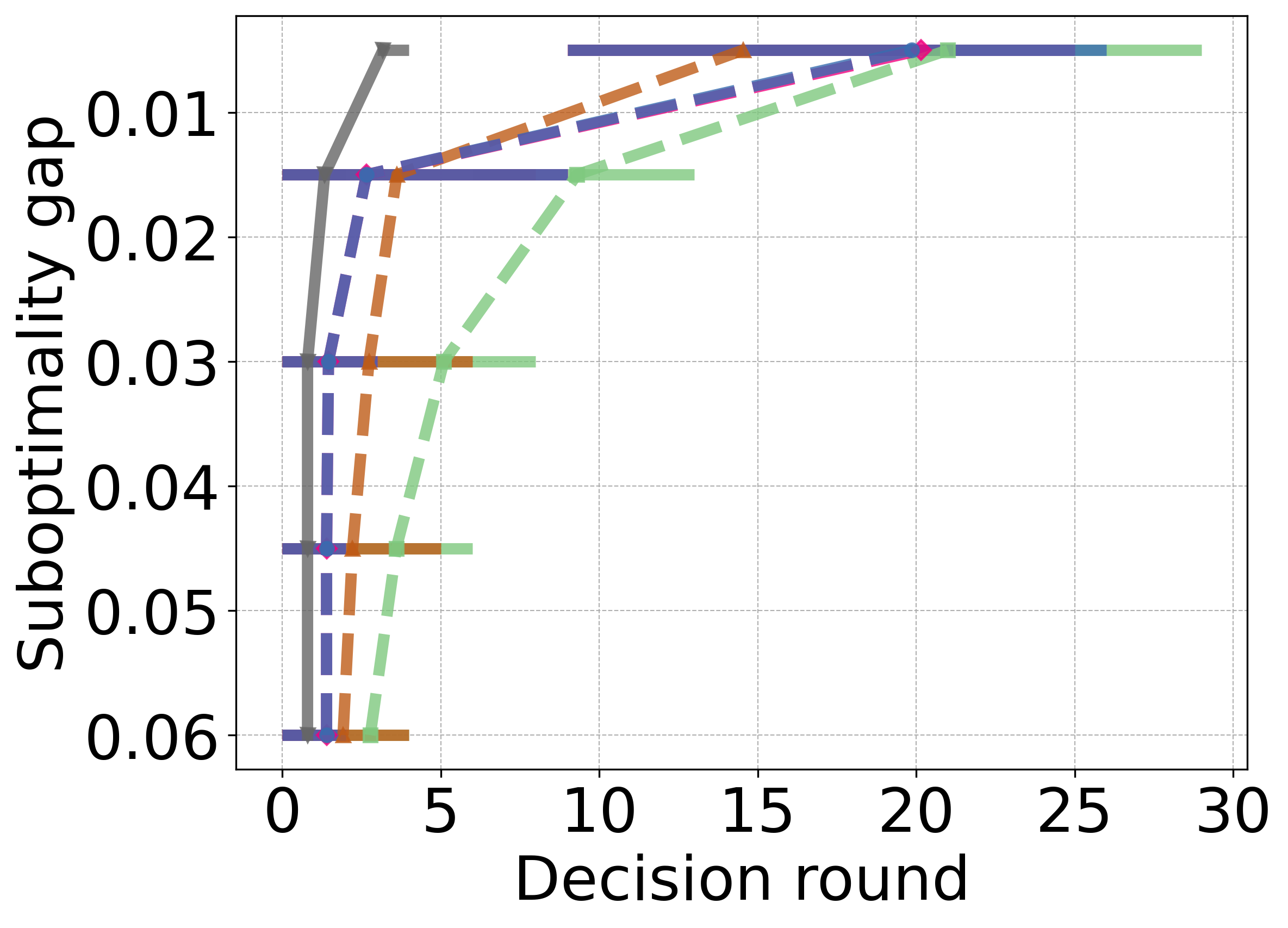}
        }
    \subfigure[BipedalWalker]{%
        \label{fig:epsilon_walker}%
        \includegraphics[width=0.45\textwidth]{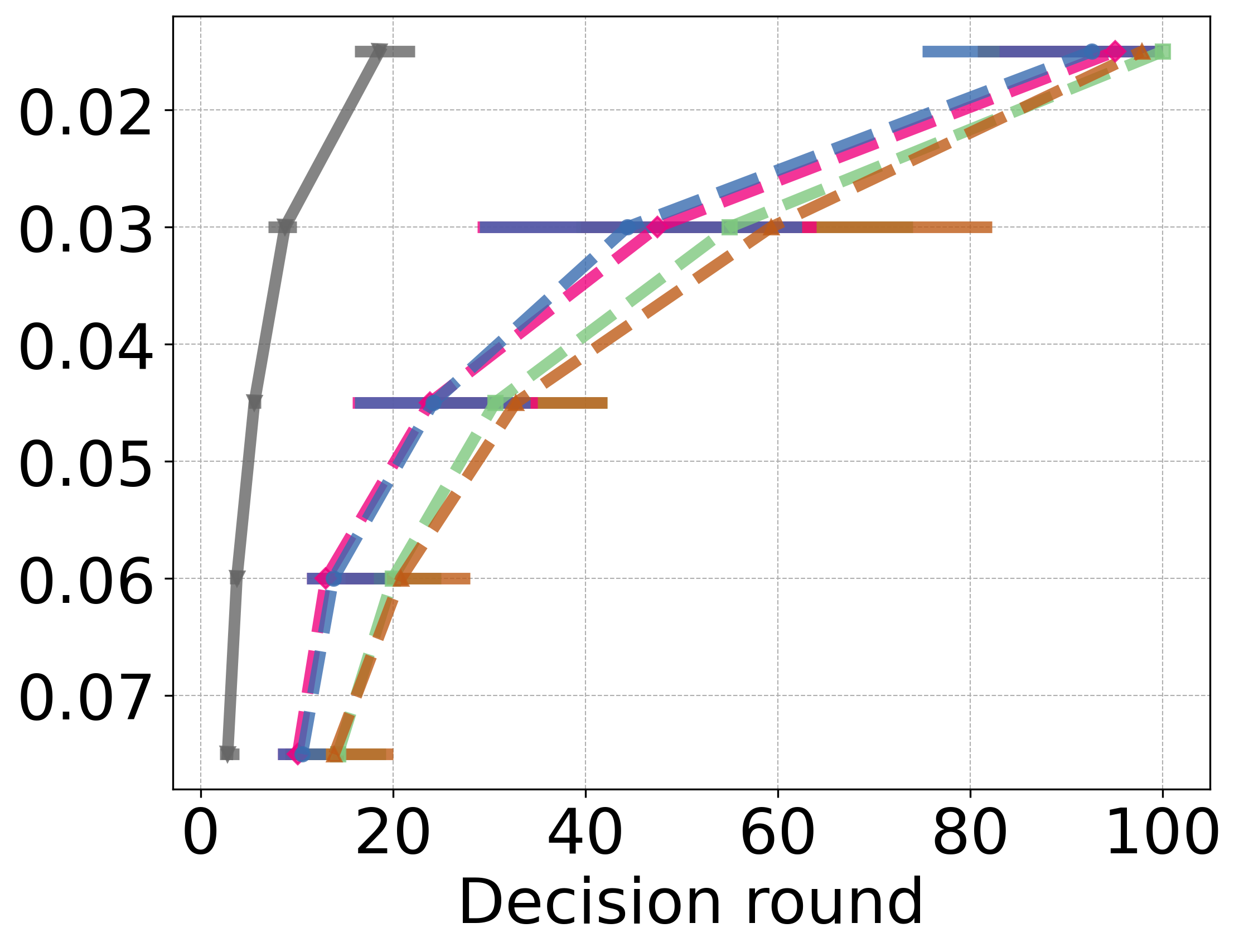}
        }%    
        \\
    \subfigure[IntersectionZoo]{%
        \label{fig:epsilon_intersectionzoo}%
        \includegraphics[width=0.45\textwidth]{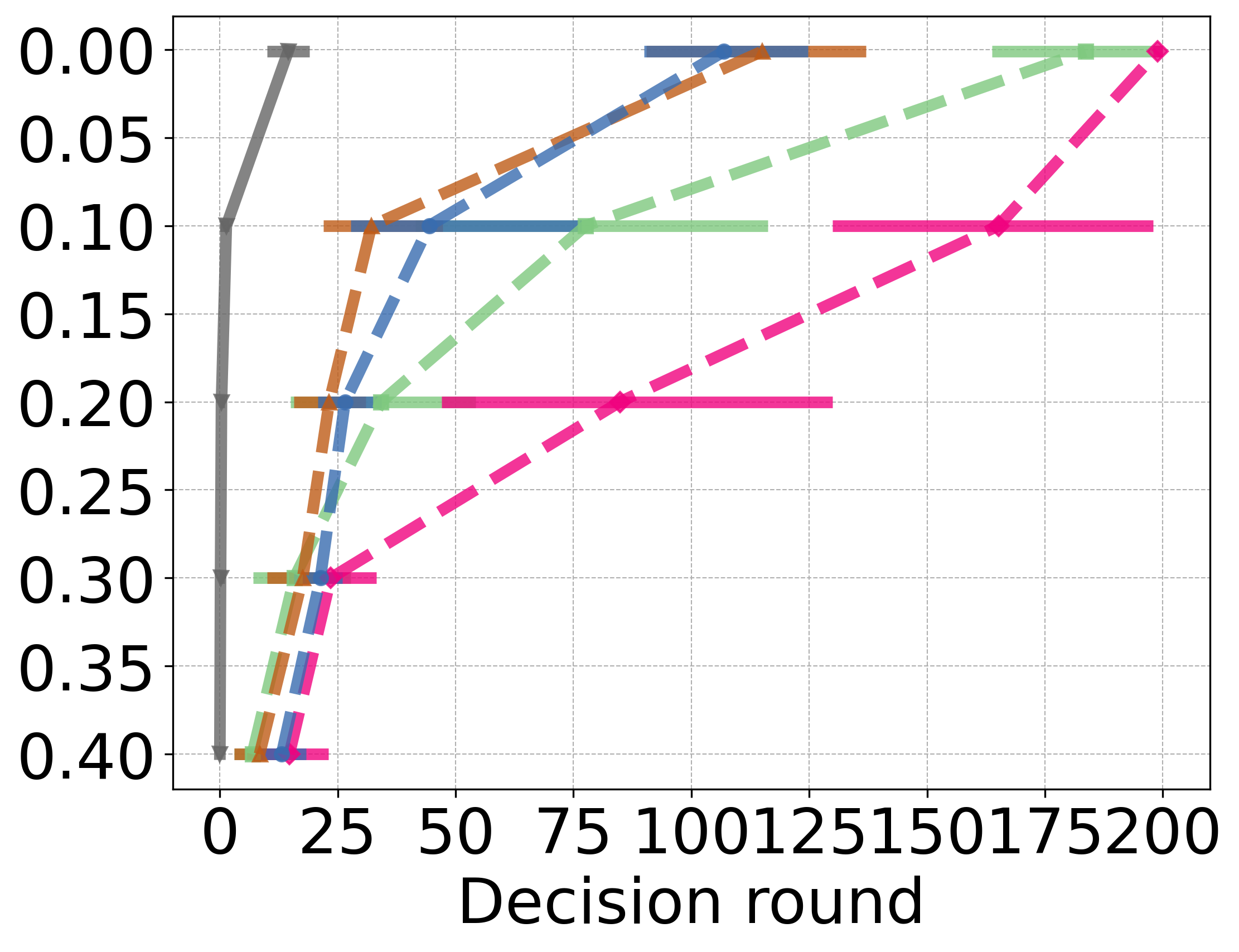}
        }%
    \subfigure[\tyyedit{\jhedit{CyclesGym}}]{%
        \label{fig:epsilon_cyclesgym}%
        \includegraphics[width=0.45\textwidth]{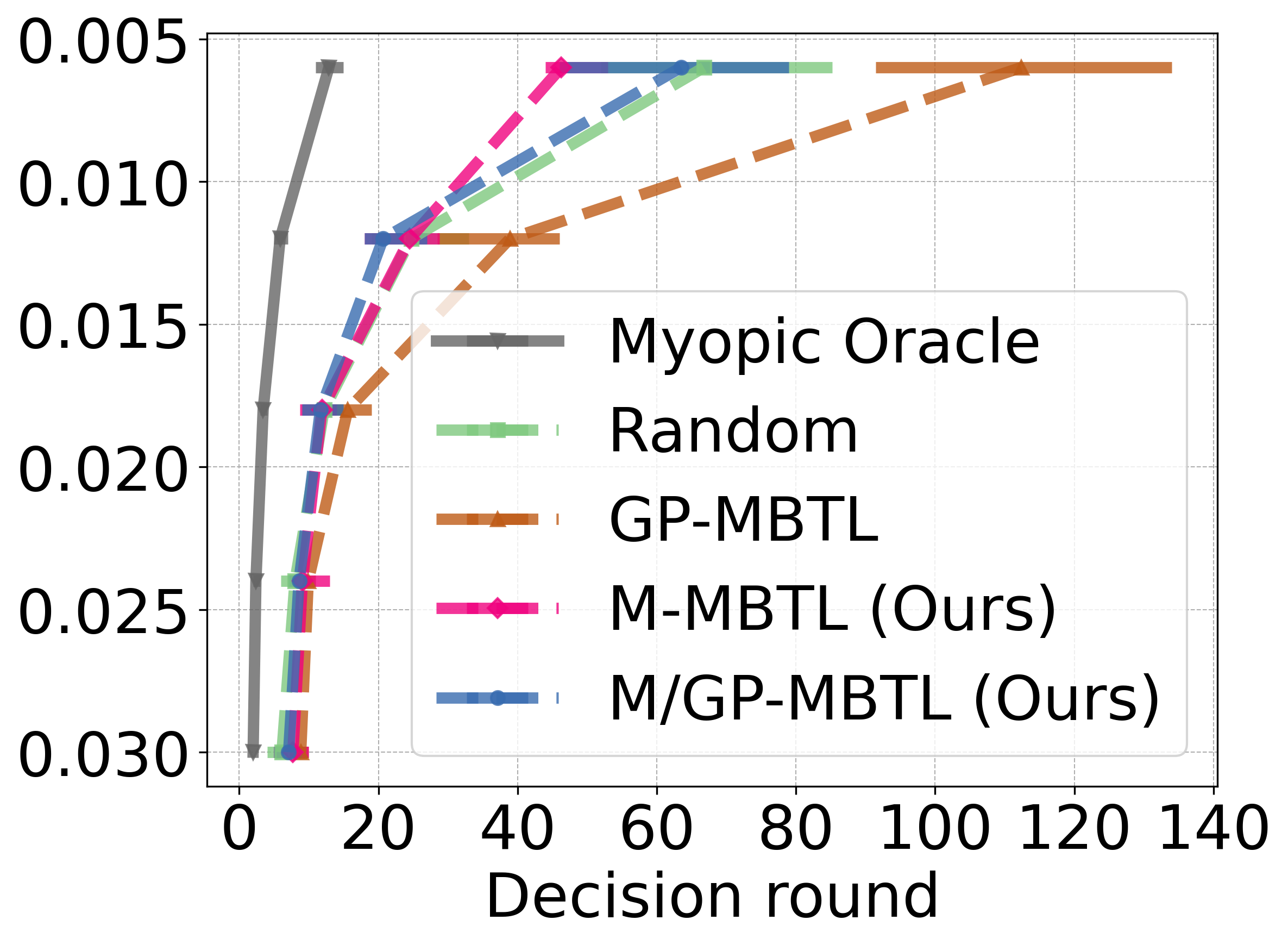}
        }%
    \caption{\jhedit{Number of source-trained policies required to achieve $\epsilon$-suboptimal generalization performance in CMDP benchmarks. Lower values indicate faster convergence to near-optimal performance.}}
    \label{fig:epsilon_all}
\end{figure*}

\tedit{Figure~\ref{fig:epsilon_all} shows the number of source-trained policies required to achieve $\epsilon$-suboptimal generalization performance in CMDP benchmarks. Figure~\ref{fig:epsilon_cartpole} shows that M-MBTL outperforms GP-MBTL in the early stage but is eventually caught up by GP-MBTL. The reason is that, early on, the Gaussian Process in GP-MBTL is trained on only a small amount of data, while M-MBTL makes more reliable training-task selections by leveraging the clustering loss. As the number of decision rounds grows, the prediction of the Gaussian Process becomes progressively more accurate.}

\begin{figure*}[!t]%
    \centering
    \subfigure[CartPole]{
        \label{fig:result_cartpole_k100}
        \includegraphics[width=0.4\textwidth]{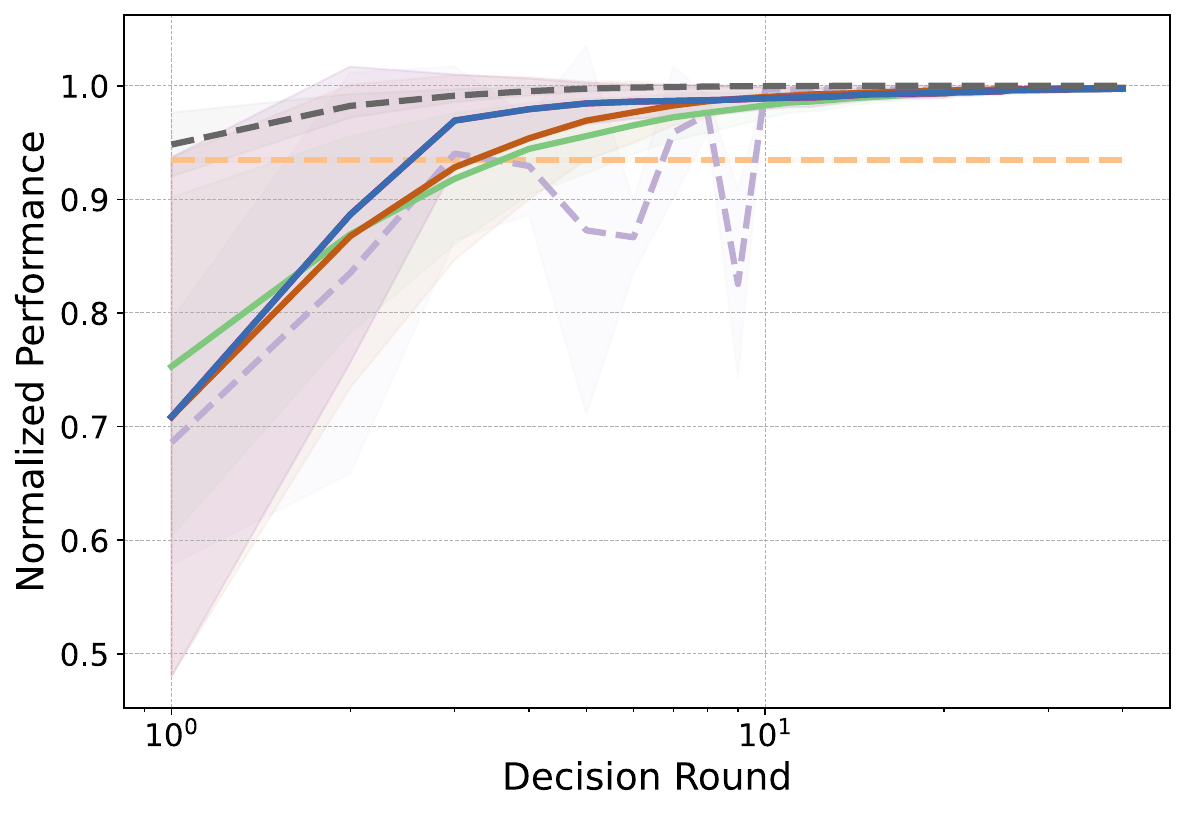}
        }%
    \subfigure[BipedalWalker]{%
        \label{fig:result_walker_k256}%
        \includegraphics[width=0.4\textwidth]{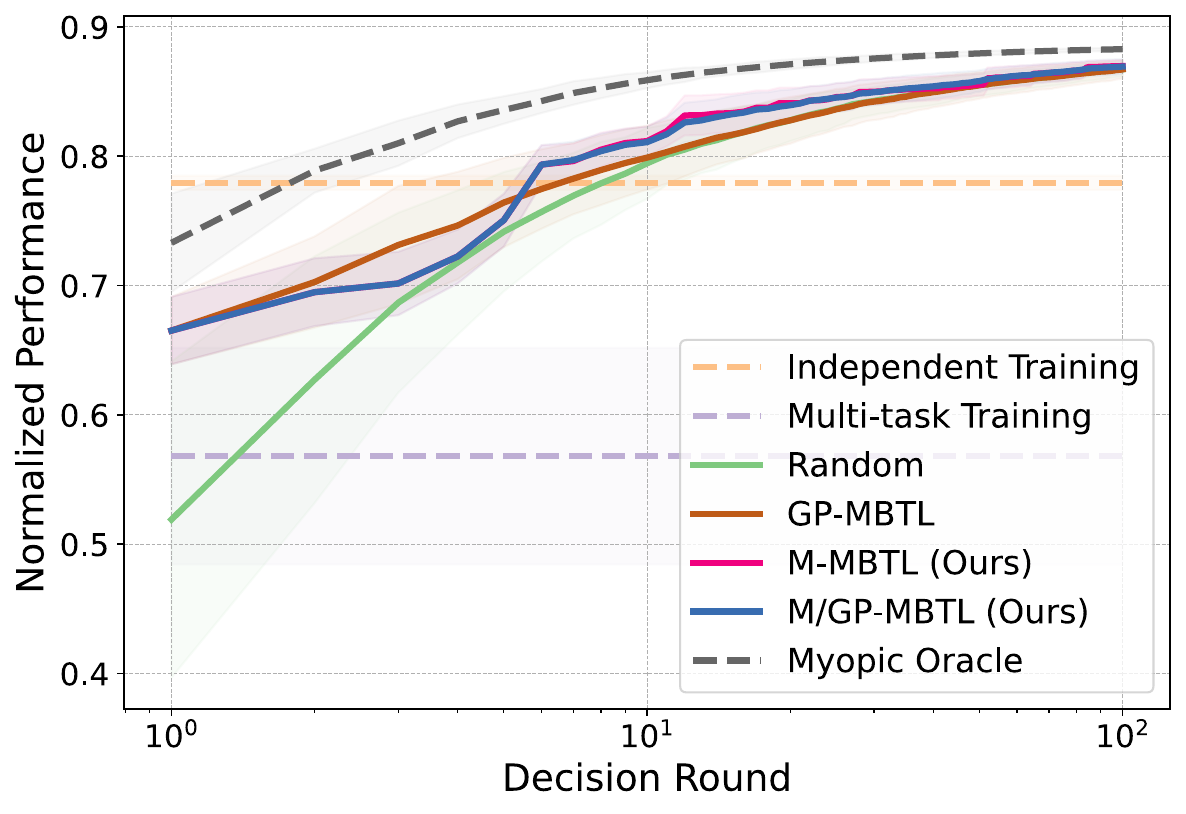}
        }%
    \\
    \subfigure[IntersectionZoo]{
        \label{fig:result_intersectionzoo_k216}
        \includegraphics[width=0.4\textwidth]{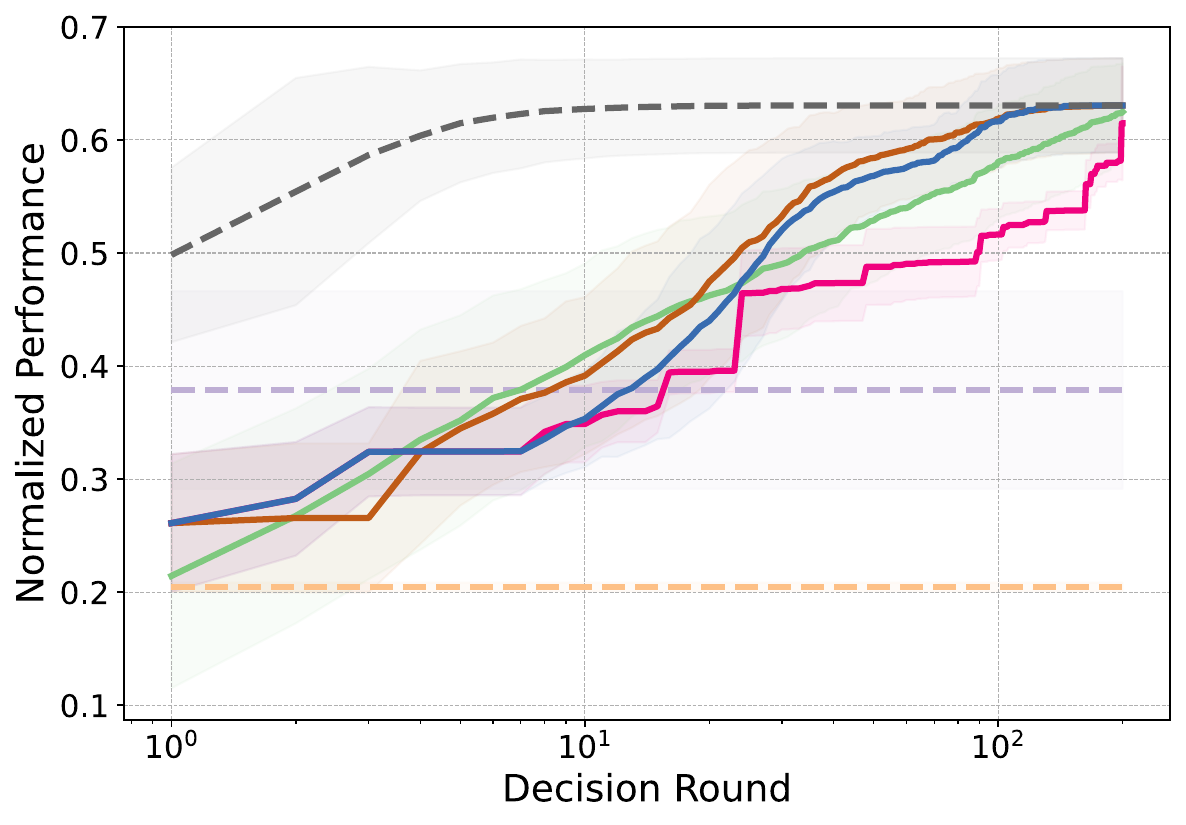}
        }%
    \subfigure[CyclesGym]{%
        \label{fig:result_crop_k216}%
        \includegraphics[width=0.4\textwidth]{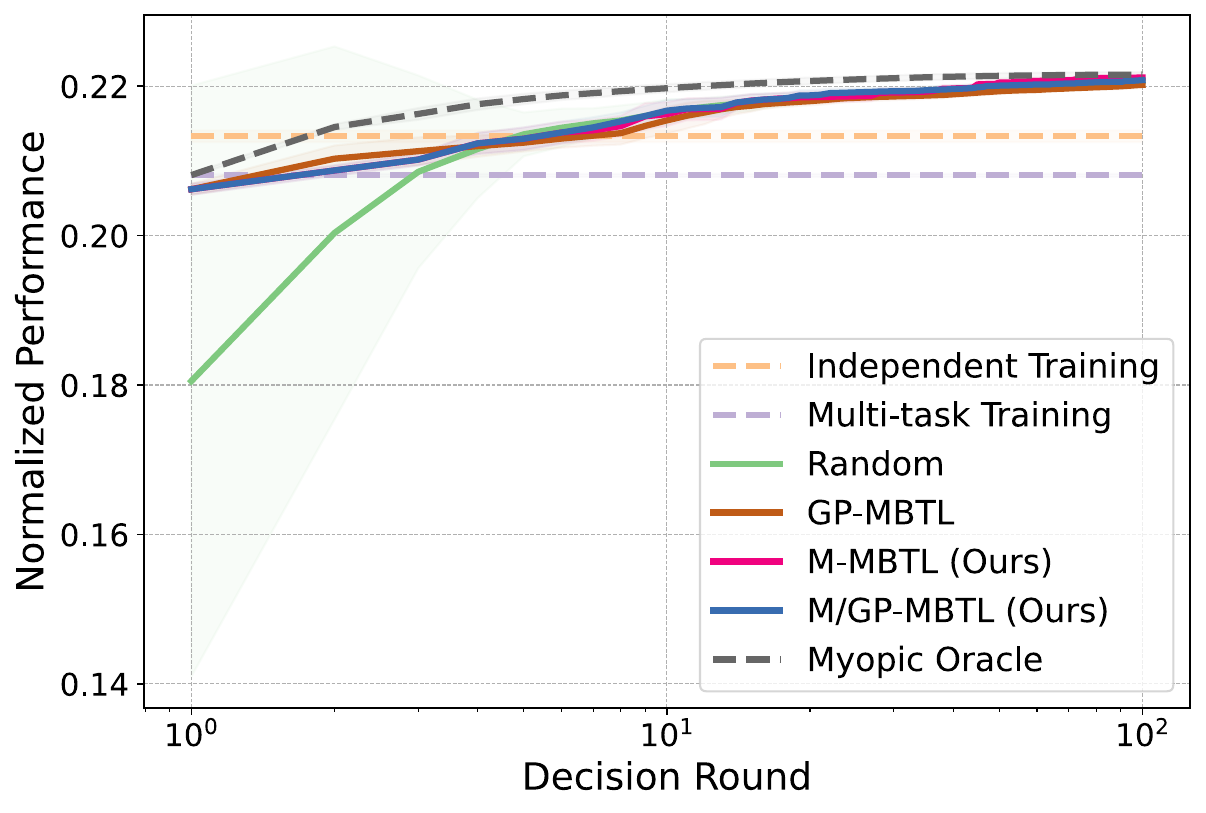}
        }%
    \caption{\jhedit{Normalized performance comparison} of \jhedit{M/GP}-MBTL \jhedit{with} other baselines (Independent Training, \jhedit{Multi-task Training, Stochastic} Oracle, Random, and MBTL baselines) \jhedit{on} the \jhedit{CartPole, BipedalWalker, IntersectionZoo, and CyclesGym benchmarks}. 
    % The x-axis represents the number of decision rounds, while the y-axis represents the normalized average performance.
    }
    \label{fig:result_all}
\end{figure*}

In Figure~\ref{fig:result_all}, we present the normalized performance of our proposed M/GP-MBTL algorithm compared with various baselines (including Independent Training, Multi-task Training, Stochastic Oracle, Random, and GP-MBTL) across four CMDP benchmarks in terms of decision rounds. \tyyedit{These results demonstrate that our M/GP-MBTL approach performs almost as well as the best of M-MBTL and GP-MBTL, and approaches the myopic oracle quickly across all benchmarks.}

\begin{figure*}[!t]%
    \centering
    \subfigure[CartPole ($K=12$)]{
        \label{fig:iqm_cartpole}
        \includegraphics[width=0.8\textwidth]{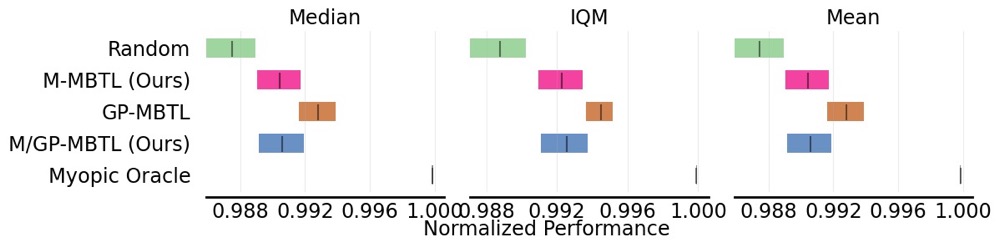}
        }%

    \subfigure[BipedalWalker ($K=12$)]{%
        \label{fig:iqm_walker}%
        \includegraphics[width=0.8\textwidth]{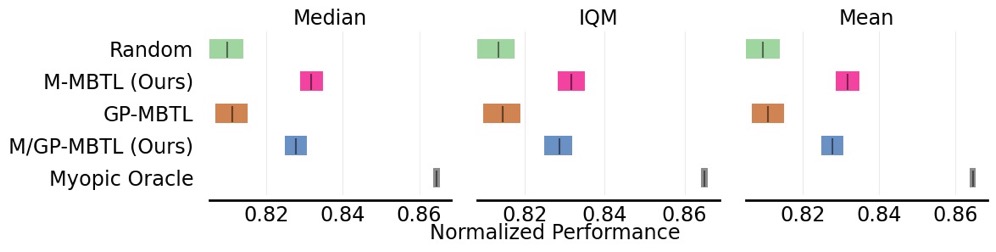}
        }%

    \subfigure[IntersectionZoo ($K=50$)]{%
        \label{fig:iqm_intersectionzoo}%
        \includegraphics[width=0.8\textwidth]{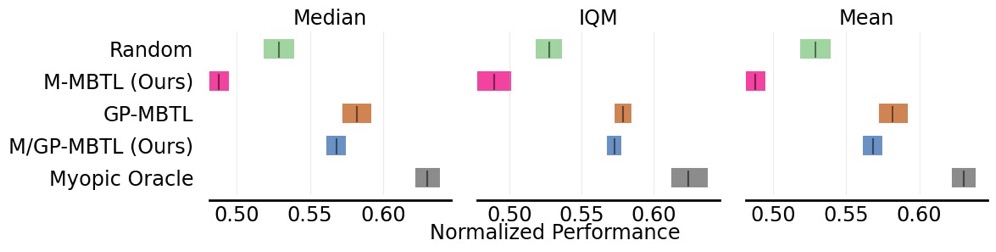}
        }%

    \subfigure[CyclesGym ($K=50$)]{%
        \label{fig:iqm_cyclesgym}%
        \includegraphics[width=0.8\textwidth]{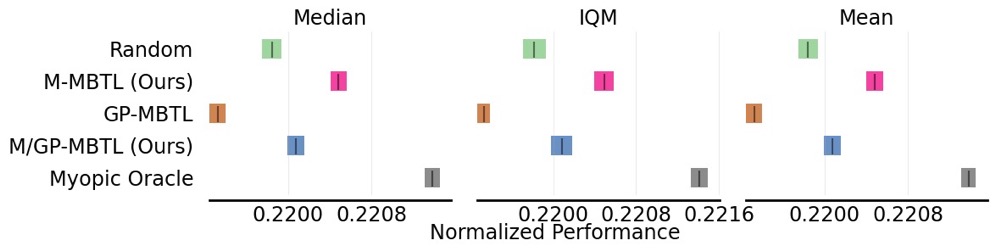}
        }%
    \caption{Performance comparison of M/GP-MBTL with baselines in terms of median, IQM, and Mean. \ttyedit{Colored bars indicate 95\% bootstrap confidence intervals for each method.}}
    \label{fig:iqm_all}
\end{figure*}

Figure~\ref{fig:iqm_all} also provides a detailed performance analysis using three different statistical measures: median, Interquartile Mean (IQM), and mean performance. IQM, which is robust to outliers as validated by \cite{agarwal2021deep}, serves as a reliable metric to compare the consistency of performance across different methods. These figures illustrate that M/GP-MBTL not only delivers high overall performance but also exhibits lower variability across experiment trials and benchmarks.

\Tyedit{Table~\ref{tab:performance_AUC} presents a comparison of area under the curve (AUC) performance. AUC measures the average performance in $K$ decision rounds, and highlights how quickly the performance of each algorithm improves. M/GP-MBTL consistently achieves the highest or near-highest AUC performance across all benchmarks.} \ttyedit{It achieves better performance than multi-task training in the CartPole benchmark, since multi-task training performs worse when the number of training steps is small.} However, the AUC performance shows a larger standard deviation because the performance of each algorithm fluctuates more when $k$ is small, especially for the random algorithm. Consequently, the performance of several algorithms may all fall within the statistically significant range.

\begin{table*}[!ht]
    \begin{small}
\begin{center}
  \caption{Performance comparison of different methods on CMDP benchmarks}
  \label{tab:performance_AUC}
  \begin{tabular}{ccccccc|c}
    \toprule
    \textbf{Benchmark (CMDP)} & \textbf{Independent} & \textbf{Multi-task} & \textbf{Random} & \textbf{GP-MBTL} & \textbf{M-MBTL (Ours)} & \textbf{M/GP-MBTL (Ours)} & \textbf{Myopic Oracle} \\
    \midrule
\begin{tabular}[c]{@{}c@{}}\textbf{CartPole}\\($K=12$)\end{tabular} & \begin{tabular}[c]{@{}c@{}}0.9346\\$\pm$ 0.0003\end{tabular} & \begin{tabular}[c]{@{}c@{}}\textbf{0.9246}\\$\pm$ 0.0393\end{tabular} & \begin{tabular}[c]{@{}c@{}}0.9502\\$\pm$ 0.0046\end{tabular} & \begin{tabular}[c]{@{}c@{}}\textbf{0.9542}\\$\pm$ 0.0069\end{tabular} & \begin{tabular}[c]{@{}c@{}}\textbf{0.9612}\\$\pm$ 0.0056\end{tabular} & \begin{tabular}[c]{@{}c@{}}\textbf{0.9612}\\$\pm$ 0.0056\end{tabular} & \begin{tabular}[c]{@{}c@{}}0.9939\\$\pm$ 0.0005\end{tabular} \\
\begin{tabular}[c]{@{}c@{}}\textbf{BipedalWalker}\\($K=12$)\end{tabular} & \begin{tabular}[c]{@{}c@{}}\textbf{0.7794}\\$\pm$ 0.0011\end{tabular} & \begin{tabular}[c]{@{}c@{}}0.5680\\$\pm$ 0.0919\end{tabular} & \begin{tabular}[c]{@{}c@{}}0.7482\\$\pm$ 0.0061\end{tabular} & \begin{tabular}[c]{@{}c@{}}0.7735\\$\pm$ 0.0043\end{tabular} & \begin{tabular}[c]{@{}c@{}}\textbf{0.7800}\\$\pm$ 0.0020\end{tabular} & \begin{tabular}[c]{@{}c@{}}\textbf{0.7788}\\$\pm$ 0.0019\end{tabular} & \begin{tabular}[c]{@{}c@{}}0.8383\\$\pm$ 0.0015\end{tabular} \\
\begin{tabular}[c]{@{}c@{}}\textbf{IntersectionZoo}\\($K=50$)\end{tabular} & \begin{tabular}[c]{@{}c@{}}0.2045\\$\pm$ 0.0008\end{tabular} & \begin{tabular}[c]{@{}c@{}}\textbf{0.3788}\\$\pm$ 0.1059\end{tabular} & \begin{tabular}[c]{@{}c@{}}0.4580\\$\pm$ 0.0107\end{tabular} & \begin{tabular}[c]{@{}c@{}}\textbf{0.4824}\\$\pm$ 0.0093\end{tabular} & \begin{tabular}[c]{@{}c@{}}0.4183\\$\pm$ 0.0040\end{tabular} & \begin{tabular}[c]{@{}c@{}}0.4626\\$\pm$ 0.0066\end{tabular} & \begin{tabular}[c]{@{}c@{}}0.6238\\$\pm$ 0.0087\end{tabular} \\
\begin{tabular}[c]{@{}c@{}}\textbf{CyclesGym}\\($K=50$)\end{tabular} & \begin{tabular}[c]{@{}c@{}}0.2133\\$\pm$ 0.0002\end{tabular} & \begin{tabular}[c]{@{}c@{}}0.2081\\$\pm$ 0.0002\end{tabular} & \begin{tabular}[c]{@{}c@{}}0.2119\\$\pm$ 0.0009\end{tabular} & \begin{tabular}[c]{@{}c@{}}0.2138\\$\pm$ 0.0002\end{tabular} & \begin{tabular}[c]{@{}c@{}}\textbf{0.2141}\\$\pm$ 0.0002\end{tabular} & \begin{tabular}[c]{@{}c@{}}\textbf{0.2143}\\$\pm$ 0.0002\end{tabular} & \begin{tabular}[c]{@{}c@{}}0.2181\\$\pm$ 0.0001\end{tabular} \\
\midrule
\begin{tabular}[c]{@{}c@{}}\textbf{Aggregated}\\\textbf{Performance}\end{tabular} & \begin{tabular}[c]{@{}c@{}}-0.5824\\$\pm$ 0.0796\end{tabular} & \begin{tabular}[c]{@{}c@{}}-1.3987\\$\pm$ 0.6393\end{tabular} & \begin{tabular}[c]{@{}c@{}}0.0000\\$\pm$ 0.0432\end{tabular} & \begin{tabular}[c]{@{}c@{}}\textbf{0.1494}\\$\pm$ 0.0471\end{tabular} & \begin{tabular}[c]{@{}c@{}}\textbf{0.1472}\\$\pm$ 0.0412\end{tabular} & \begin{tabular}[c]{@{}c@{}}\textbf{0.2198}\\$\pm$ 0.0364\end{tabular} & \begin{tabular}[c]{@{}c@{}}1.0000\\$\pm$ 0.0155\end{tabular} \\
\bottomrule
    \end{tabular}
\end{center}
    \scriptsize{* \textit{Note}: Values are reported as the mean performance ± half the width of the 95\% confidence interval. Bold values represent the highest value(s) within the statistically significant range for each task, excluding the oracle.\\}
\end{small}
\end{table*}

\begin{figure*}[ht]
  \centering
  \includegraphics[width=0.9\linewidth]{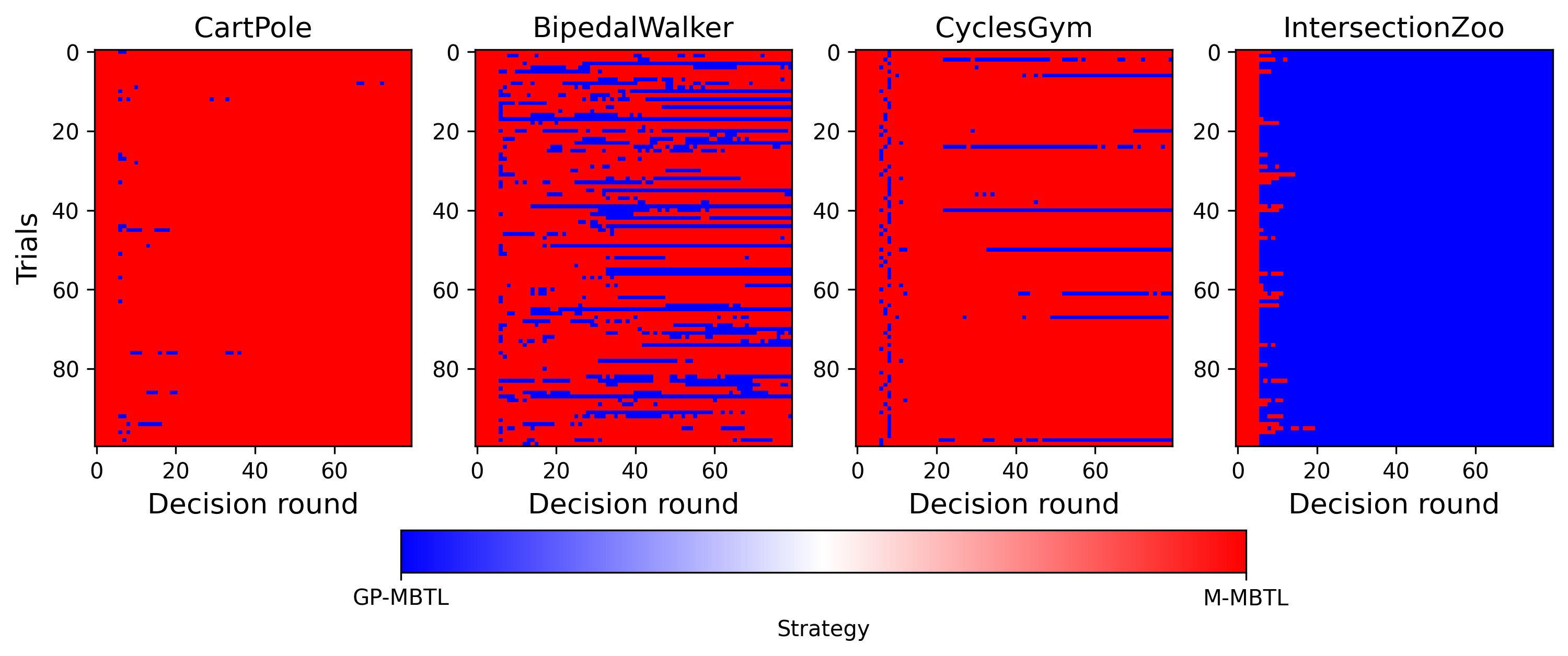}
  \caption{\jhedit{Per‐round algorithm selection over 100 trials ($K=80$) across four benchmarks. Red pixels indicate rounds where M‐MBTL was chosen; blue where GP‐MBTL was chosen. CartPole and \jhedit{CyclesGym} remain almost entirely in M‐MBTL (mountain structure), IntersectionZoo almost entirely in GP‐MBTL (non‐mountain), and BipedalWalker exhibits a dynamic mix.}}
  \label{fig:algswitch_benchmarks}
\end{figure*}

\jhedit{In Figure \ref{fig:algswitch_benchmarks}, we visualize, for each of 100 independent trials and over 80 decision rounds ($K=80$), which sub‐algorithm our detector chose at each step: red entries mark rounds where M‐MBTL was used, and blue entries mark GP‐MBTL. As expected, in CartPole and \jhedit{CyclesGym} CMDP, which both exhibit strong \textsc{Mountain} structure, M‐MBTL dominates almost all rounds; in IntersectionZoo, which violates \textsc{Mountain}, GP‐MBTL is selected nearly throughout; and in BipedalWalker, where the structure only partially holds, the detector switches dynamically, reflecting a mixture of both strategies. This confirms that our online structure detector is correctly steering the algorithm towards the most appropriate source‐task selection method in each setting.}

\begin{figure*}
\centering
\subfigure[Cartpole \& BipedalWalker]{%
        \label{fig:any_K_cartpole_walker}
        \includegraphics[width=0.4\textwidth]{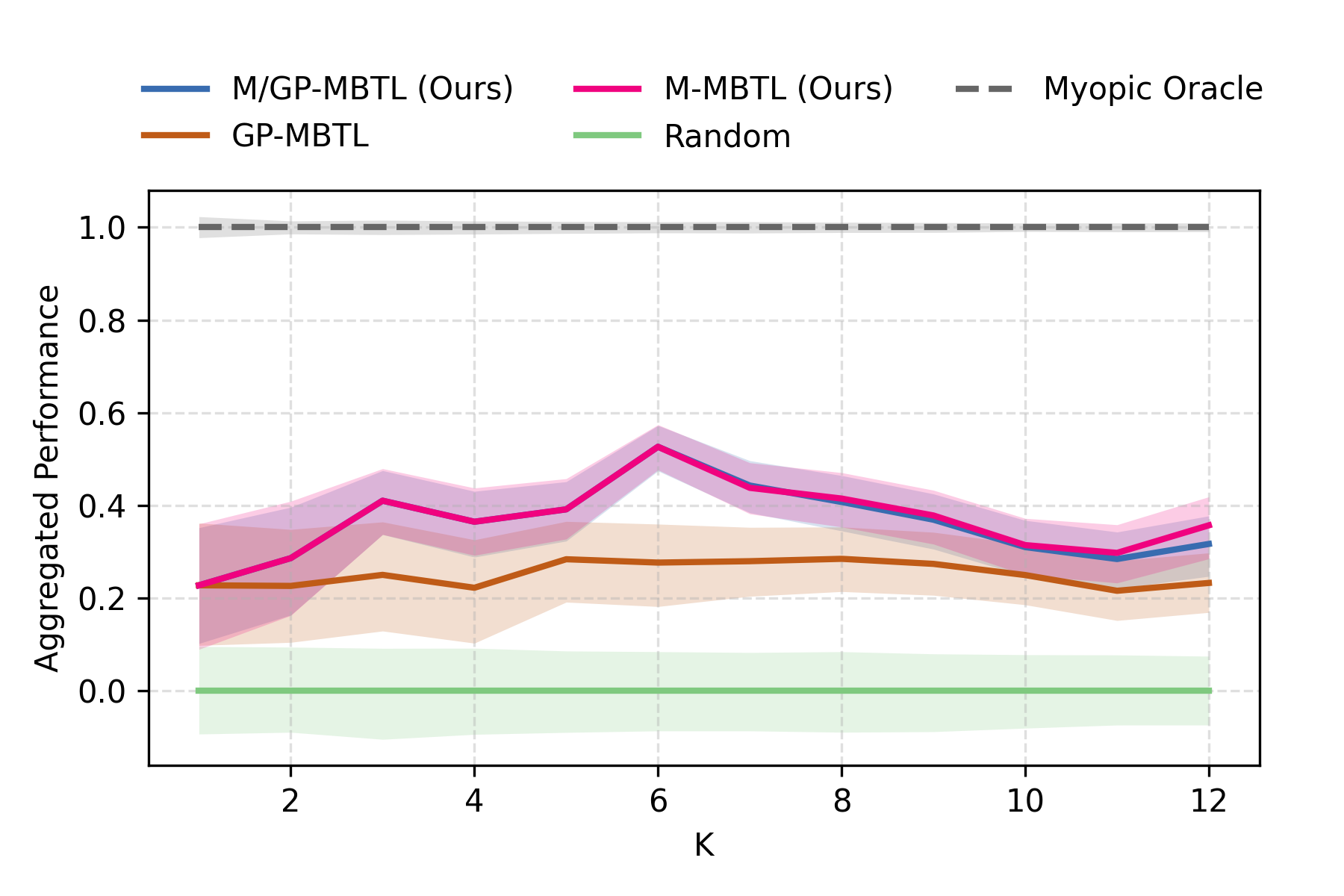}
        }%
\subfigure[IntersectionZoo \& CyclesGym]{%
        \label{fig:any_K_Zoo_Crop}
        \includegraphics[width=0.4\textwidth]{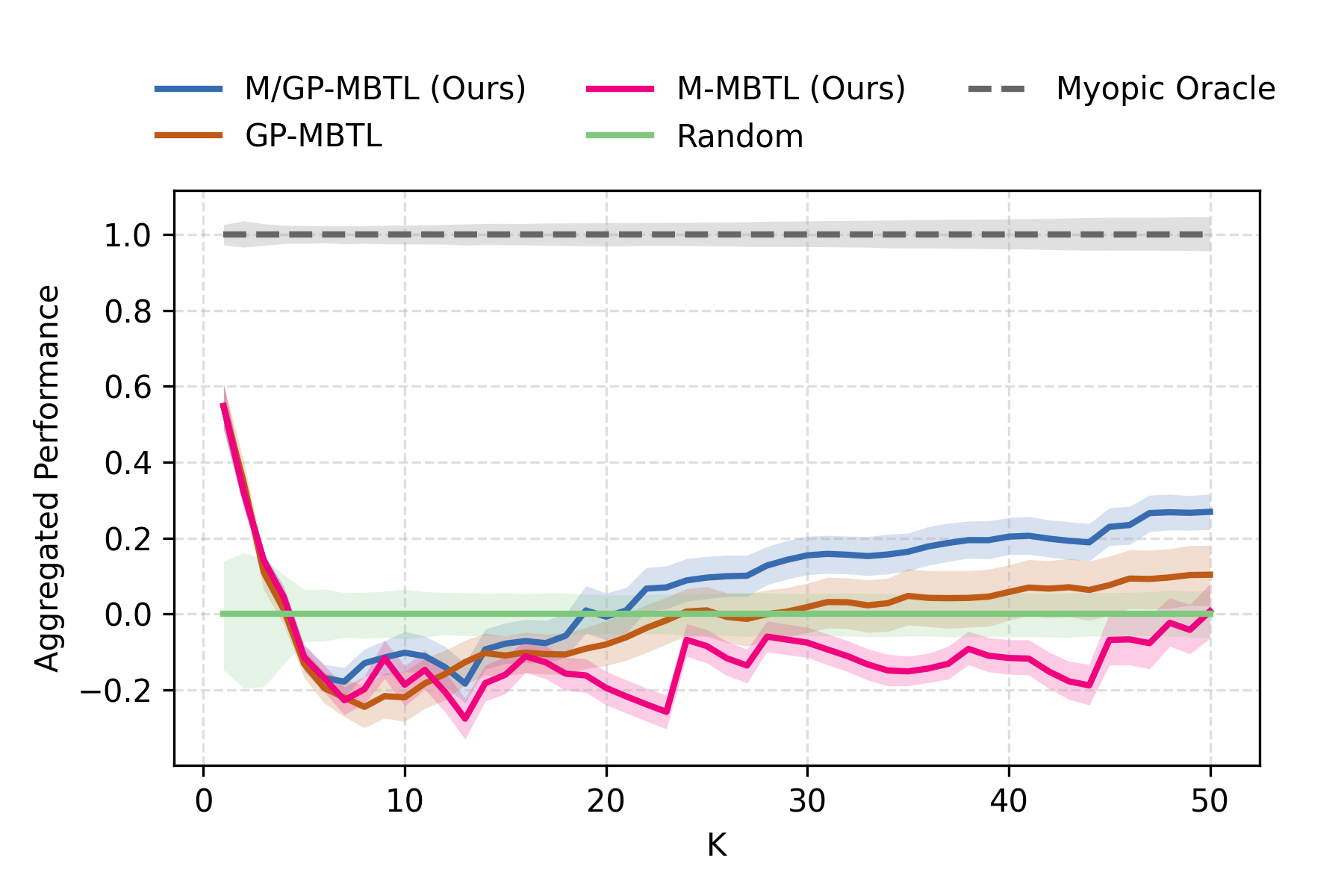}
        }%
        \caption{The plot of the aggregated metric versus K on the benchmark dataset.}
        \label{fig:any_K_benchmark}
    
\end{figure*}
Figure~\ref{fig:any_K_benchmark} shows how the aggregated metric varies with K on benchmark dataset. 
Different benchmarks require different values of K for the algorithms to converge to stable performance. Classic control tasks such as Cartpole and Walker converge with relatively small K, whereas IntersectionZoo and CyclesGym require much larger K. Therefore, we group the four benchmarks into two sets and plot the aggregated metric curves separately. Figure~\ref{fig:any_K_cartpole_walker} shows the aggregated metrics for Cartpole and Walker, where M/GP-MBTL relies mainly on M-MBTL, resulting in similar performance between the two. Figure~\ref{fig:any_K_Zoo_Crop} shows the aggregated metrics for IntersectionZoo and CyclesGym, where the performance of M/GP-MBTL rises rapidly after around K = 7 and surpasses the other baselines.

\subsection{Abalation Study}\label{sec:ablation}
\begin{figure*}[ht]
    \centering
    \includegraphics[width=0.99\linewidth]{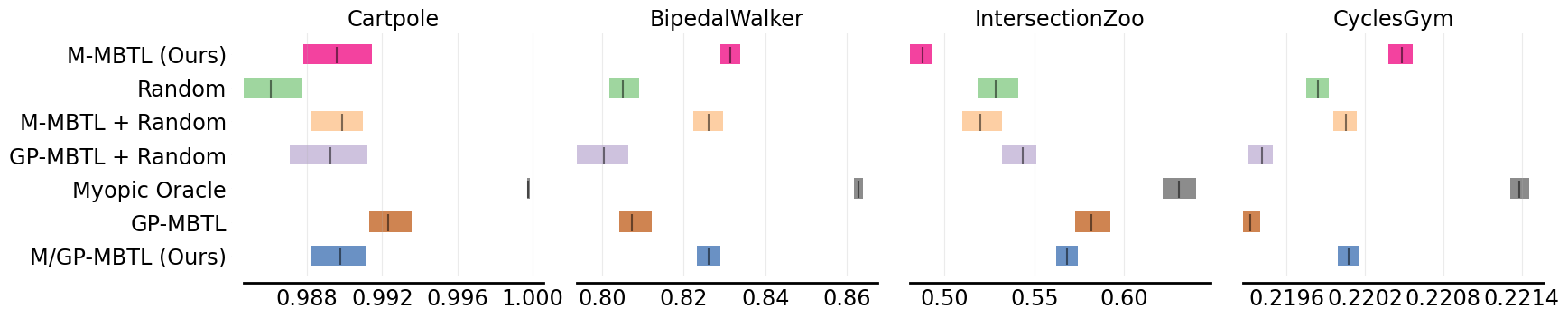}
    \caption{\tyyedit{Ablation study comparing M/GP-MBTL with four variants---stand-alone M-MBTL, GP-MBTL, M-MBTL+Random, and GP-MBTL+Random.} \ttyedit{Colored bars indicate 95\% bootstrap confidence intervals for each method.}}
    \label{fig:ablation_study} 
\end{figure*}

% \begin{wrapfigure}[8]{r}{.8\textwidth}
%     \centering
%     \includegraphics[width=0.99\linewidth]{images/ablation.png}
%     \caption{\tyyedit{Ablation study comparing M/GP-MBTL with four variants---stand-alone M-MBTL, GP-MBTL, M-MBTL+Random, and GP-MBTL+Random.}}
%     \label{fig:ablation_study} 
% \end{wrapfigure}
% \tyyedit{\textbf{Ablation Study.}}
To verify the effectiveness of individual algorithms and the online structure detection module in M/GP-MBTL, we compare M/GP-MBTL against standalone M-MBTL, standalone GP-MBTL, and two hybrid baselines that each combine an algorithm with a random policy.
M-MBTL+Random uses M-MBTL only when the Mountain structure is detected, and GP-MBTL+Random applies GP-MBTL only when its linear-gap assumption holds. In all other cases, they are reverted to random selection.

Figure~\ref{fig:ablation_study} shows the ablation study results.
In CartPole, BipedalWalker, and \jhedit{CyclesGym}, M-MBTL+Random performs nearly as well as M/GP-MBTL because those tasks largely satisfy Mountain structure. Across BipedalWalker, \jhedit{CyclesGym}, and IntersectionZoo, \jhedit{M/GP-MBTL} consistently outperforms GP-MBTL+Random, and it also exceeds M-MBTL+Random on IntersectionZoo. These results demonstrate that combining M-MBTL and GP-MBTL enables the framework to leverage the strengths of both methods, and that M/GP-MBTL selects the appropriate algorithm under various CMDP structures.

\subsection{Performance Distribution: Real vs. Bootstrapped Data}\label{sec:bootstrap_distribution}
Due to the substantial computational resources required to independently train each CMDP, we were limited to only three trials per task. Even these three trials took several months to complete. To address this limitation---and leveraging the fact that each context-MDP training is independent---we generated 97 bootstrapped datasets by combining the available trials in different ways.
For the construction of the transfer matrix, each row (representing a source task) was randomly selected from one of the three available training runs. Consequently, the rows of the matrix may originate from different combinations of these trials.

\begin{figure*}
    \centering
    \includegraphics[width=\textwidth]{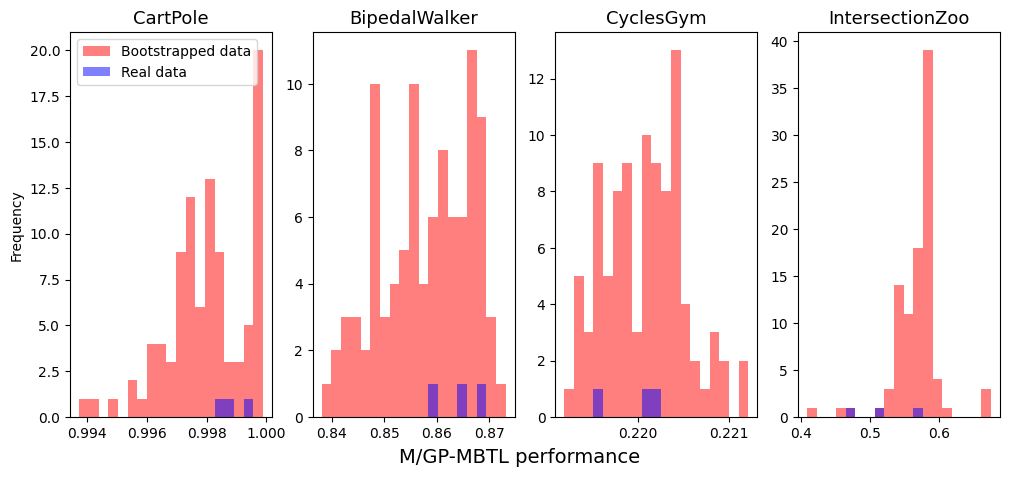}
    \caption{Histogram of M/GP-MBTL performance for real data and bootstrapped data}
    \label{fig:hist-bootstrap}
\end{figure*}

Figure~\ref{fig:hist-bootstrap} presents histograms that compare M/GP-MBTL performance. This comparison is drawn between performance achieved using the original, limited real DRL data and performance derived from the 97 bootstrapped datasets, across four distinct CMDP benchmarks: CartPole, BipedalWalker, CyclesGym, and IntersectionZoo. The results depicted in the figure demonstrate a consistency between the performance distributions obtained from the bootstrapped data and the discrete performance points observed from the real trials. The bootstrapping process generates a more continuous spectrum of performance values. Crucially, the denser regions within these bootstrapped distributions generally align well with the specific performance levels indicated by the sparse real data.

\subsection{Higher Dimensional Experiments}\label{sec:higher_dim}
Tables \ref{tab:synthetic-performance-5D-noise5} and \ref{tab:synthetic-performance-7D-noise5} report the performance comparison between M/GP-MBTL and other baselines under 5-dimensional and 7-dimensional context settings.

In the 5-dimensional experiments, we use a linear function $f(x)=[2,2,2,2,2]\cdot x$ and $g(y)=[2,2,2,2,2]\cdot y$ for non-constant case, and $f(x)=0$, $g(y)=0$ for constant case. When distance metric holds, $h(x,y)=-[3,3,3,3,3]\cdot |x-y|$. We test two non-distance cases. In case 1, $h(x,y)=-([3,3,3,3,3]\cdot [x-y]_+ +[1,1,1,-3,-3]\cdot [x-y]_-)$. In case 2, $h(x,y)=-([3,3,3,3,3]\cdot [x-y]_+ +[1,1,1,1,-3]\cdot [x-y]_-)$. 

Since testing in high-dimensional settings typically requires substantial computation time, in the 7-dimensional experiments we only tested the four cases regarding whether $g(y)$ is constant and whether $h(x,y)$ satisfies the properties of a distance metric.
We use a linear function $g(y)=[2,2,2,2,2,2,2]\cdot y$ for non-constant case, and $g(y)=0$ for constant case. When distance metric holds, $h(x,y)=-[3,3,3,3,3,3,3]\cdot |x-y|$. Otherwise, $h(x,y)=-([3,3,3,3,3,3,3]\cdot [x-y]_+ +[1,1,1,1,1,-3,-3]\cdot [x-y]_-)$.

In both the 5-dimensional and 7-dimensional experiments, M/GP-MBTL demonstrated the best overall performance with relatively small confidence intervals, indicating greater robustness. In contrast, GP-MBTL showed much larger confidence intervals, reflecting its instability across different CMDP structures.

\begin{table*}[!ht]
\begin{small}
\begin{center}
  \caption{Performance comparison on 5D Synthetic Data ($\epsilon=\mathcal{N}(0, 5^2)$) with K=50.}
  \label{tab:synthetic-performance-5D-noise5}
  \begin{tabular}{lllcccc|c}
    \toprule
    \textbf{$f(x)$} & \textbf{$g(y)$} & \textbf{$h(x,y)$}  & \textbf{Random} & \textbf{GP-MBTL} & \textbf{M-MBTL} & \textbf{M/GP-MBTL (Ours)} & \textbf{Myopic Oracle}\\
    \midrule
    Linear & Linear & $L_1$ norm & {0.5996} $\pm$ 0.0039 & \textbf{0.6229} $\pm$ 0.0003 & {0.5993} $\pm$ 0.0010 & {0.6122} $\pm$ 0.0091 & {0.6265} $\pm$ 0.0002\\
    Linear & Linear & Non-distance 1 & \textbf{0.6142} $\pm$ 0.0012 & {0.6120} $\pm$ 0.0022 & \textbf{0.6156} $\pm$ 0.0002 & \textbf{0.6160} $\pm$ 0.0006 & {0.6236} $\pm$ 0.0004\\
    Linear & Linear & Non-distance 2 & {0.6128} $\pm$ 0.0018 & {0.6140} $\pm$ 0.0020 & {0.6142} $\pm$ 0.0005 & \textbf{0.6172} $\pm$ 0.0008 & {0.6262} $\pm$ 0.0005\\
    Linear & Constant & $L_1$ norm & {0.7348} $\pm$ 0.0040 & \textbf{0.7440} $\pm$ 0.0071 & {0.7344} $\pm$ 0.0011 & \textbf{0.7461} $\pm$ 0.0061 & {0.7626} $\pm$ 0.0002\\
    Linear & Constant & Non-distance 1 & \textbf{0.6891} $\pm$ 0.0016 & {0.6810} $\pm$ 0.0060 & \textbf{0.6910} $\pm$ 0.0003 & \textbf{0.6910} $\pm$ 0.0010 & {0.7016} $\pm$ 0.0005\\
    Linear & Constant & Non-distance 2 & \textbf{0.7007} $\pm$ 0.0023 & {0.6930} $\pm$ 0.0076 & \textbf{0.7024} $\pm$ 0.0006 & \textbf{0.7012} $\pm$ 0.0031 & {0.7175} $\pm$ 0.0006\\
    \rowcolor{mountainrow} Constant & Linear & $L_1$ norm & {0.7049} $\pm$ 0.0006 & {0.6887} $\pm$ 0.0089 & \textbf{0.7090} $\pm$ 0.0004 & \textbf{0.7089} $\pm$ 0.0003 & {0.7117} $\pm$ 0.0004\\
    Constant & Linear & Non-distance 1 & {0.6982} $\pm$ 0.0068 & \textbf{0.7463} $\pm$ 0.0006 & {0.6917} $\pm$ 0.0019 & {0.7421} $\pm$ 0.0014 & {0.7482} $\pm$ 0.0002\\
    Constant & Linear & Non-distance 2 & {0.7000} $\pm$ 0.0027 & \textbf{0.7457} $\pm$ 0.0021 & {0.6987} $\pm$ 0.0015 & \textbf{0.7441} $\pm$ 0.0017 & {0.7503} $\pm$ 0.0001\\
    \rowcolor{mountainrow} Constant & Constant & $L_1$ norm & {0.7374} $\pm$ 0.0007 & \textbf{0.7411} $\pm$ 0.0010 & \textbf{0.7423} $\pm$ 0.0005 & \textbf{0.7422} $\pm$ 0.0004 & {0.7457} $\pm$ 0.0005\\
    Constant & Constant & Non-distance 1 & {0.7482} $\pm$ 0.0089 & \textbf{0.8134} $\pm$ 0.0012 & {0.7396} $\pm$ 0.0026 & \textbf{0.8129} $\pm$ 0.0002 & {0.8150} $\pm$ 0.0003\\
    Constant & Constant & Non-distance 2 & {0.7548} $\pm$ 0.0035 & \textbf{0.8189} $\pm$ 0.0002 & {0.7531} $\pm$ 0.0020 & {0.8167} $\pm$ 0.0008 & {0.8202} $\pm$ 0.0001\\
\midrule
\multicolumn{3}{c}{\textbf{Aggregated Performance}} & {0.0000} $\pm$ 0.0356 & {0.1528} $\pm$ 0.2728 & {0.1124} $\pm$ 0.0602 & \textbf{0.5327} $\pm$ 0.0906 & {1.0000} $\pm$ 0.0090 \\
    \bottomrule
  \end{tabular}
  \end{center}
  \scriptsize{* \textit{Note}: Values are reported as the mean performance ± half the width of the 95\% confidence interval. Rows shaded in light gray indicate the synthetic settings that satisfy our proposed \textsc{Mountain} structure assumption. Bold values represent the highest value(s) within the statistically significant range for each task, excluding the oracle.}
  \end{small}
\end{table*}

\begin{table*}[!ht]
\begin{small}
\begin{center}
  \caption{Performance comparison on 7D Synthetic Data ($\epsilon=\mathcal{N}(0, 5^2)$) with K=50.}
  \label{tab:synthetic-performance-7D-noise5}
  \begin{tabular}{lllcccc|c}
    \toprule
    \textbf{$f(x)$} & \textbf{$g(y)$} & \textbf{$h(x,y)$}  & \textbf{Random} & \textbf{GP-MBTL} & \textbf{M-MBTL} & \textbf{M/GP-MBTL (Ours)} & \textbf{Myopic Oracle}\\
    \midrule
    \rowcolor{mountainrow} Constant & Constant & $L_1$ norm & {0.7136} $\pm$ 0.0002 & {0.7171} $\pm$ 0.0013 & \textbf{0.7199} $\pm$ 0.0002 & \textbf{0.7191} $\pm$ 0.0013 & {0.7239} $\pm$ 0.0001\\
    Constant & Constant & Non-distance & {0.7395} $\pm$ 0.0001 & \textbf{0.8151} $\pm$ 0.0009 & {0.7308} $\pm$ 0.0002 & \textbf{0.8139} $\pm$ 0.0003 & {0.8172} $\pm$ 0.0001\\
    \rowcolor{mountainrow} Constant & Linear & $L_1$ norm & {0.6845} $\pm$ 0.0001 & {0.6786} $\pm$ 0.0025 & \textbf{0.6895} $\pm$ 0.0002 & \textbf{0.6895} $\pm$ 0.0002 & {0.6928} $\pm$ 0.0001\\
    Constant & Linear & Non-distance & {0.6948} $\pm$ 0.0001 & \textbf{0.7466} $\pm$ 0.0051 & {0.6881} $\pm$ 0.0001 & \textbf{0.7435} $\pm$ 0.0046 & {0.7548} $\pm$ 0.0001\\
\midrule
\multicolumn{3}{c}{\textbf{Aggregated Performance}} & {-0.0000} $\pm$ 0.0051 & {0.3625} $\pm$ 0.3053 & {0.2473} $\pm$ 0.1465 & \textbf{0.7270} $\pm$ 0.0811 & {1.0000} $\pm$ 0.0032 \\
    \bottomrule
  \end{tabular}
  \end{center}
  \scriptsize{* \textit{Note}: Values are reported as the mean performance ± half the width of the 95\% confidence interval. Rows shaded in light gray indicate the synthetic settings that satisfy our proposed \textsc{Mountain} structure assumption. Bold values represent the highest value(s) within the statistically significant range for each task, excluding the oracle.}
  \end{small}
\end{table*}

\clearpage
\makeatletter
\@ifundefined{isChecklistMainFile}{
  % We are compiling a standalone document
  \newif\ifreproStandalone
  \reproStandalonetrue
}{
  % We are being \input into the main paper
  \newif\ifreproStandalone
  \reproStandalonefalse
}
\makeatother

\ifreproStandalone
\documentclass[letterpaper]{article}
\usepackage[submission]{aaai2026}
\setlength{\pdfpagewidth}{8.5in}
\setlength{\pdfpageheight}{11in}
\usepackage{times}
\usepackage{helvet}
\usepackage{courier}
\usepackage{xcolor}
\frenchspacing

\begin{document}
\fi
\setlength{\leftmargini}{20pt}
\makeatletter\def\@listi{\leftmargin\leftmargini \topsep .5em \parsep .5em \itemsep .5em}
\def\@listii{\leftmargin\leftmarginii \labelwidth\leftmarginii \advance\labelwidth-\labelsep \topsep .4em \parsep .4em \itemsep .4em}
\def\@listiii{\leftmargin\leftmarginiii \labelwidth\leftmarginiii \advance\labelwidth-\labelsep \topsep .4em \parsep .4em \itemsep .4em}\makeatother

\setcounter{secnumdepth}{0}
\renewcommand\thesubsection{\arabic{subsection}}
\renewcommand\labelenumi{\thesubsection.\arabic{enumi}}

\newcounter{checksubsection}
\newcounter{checkitem}[checksubsection]

\newcommand{\checksubsection}[1]{%
  \refstepcounter{checksubsection}%
  \paragraph{\arabic{checksubsection}. #1}%
  \setcounter{checkitem}{0}%
}

\newcommand{\checkitem}{%
  \refstepcounter{checkitem}%
  \item[\arabic{checksubsection}.\arabic{checkitem}.]%
}
\newcommand{\question}[2]{\normalcolor\checkitem #1 #2 \color{blue}}
\newcommand{\ifyespoints}[1]{\makebox[0pt][l]{\hspace{-15pt}\normalcolor #1}}

\section*{Reproducibility Checklist}

\vspace{1em}
\hrule
\vspace{1em}

\textbf{Instructions for Authors:}

This document outlines key aspects for assessing reproducibility. Please provide your input by editing this \texttt{.tex} file directly.

For each question (that applies), replace the ``Type your response here'' text with your answer.

\vspace{1em}
\noindent
\textbf{Example:} If a question appears as
\begin{center}
\noindent
\begin{minipage}{.9\linewidth}
\ttfamily\raggedright
\string\question \{Proofs of all novel claims are included\} \{(yes/partial/no)\} \\
Type your response here
\end{minipage}
\end{center}
you would change it to:
\begin{center}
\noindent
\begin{minipage}{.9\linewidth}
\ttfamily\raggedright
\string\question \{Proofs of all novel claims are included\} \{(yes/partial/no)\} \\
yes
\end{minipage}
\end{center}
Please make sure to:
\begin{itemize}\setlength{\itemsep}{.1em}
\item Replace ONLY the ``Type your response here'' text and nothing else.
\item Use one of the options listed for that question (e.g., \textbf{yes}, \textbf{no}, \textbf{partial}, or \textbf{NA}).
\item \textbf{Not} modify any other part of the \texttt{\string\question} command or any other lines in this document.\\
\end{itemize}

You can \texttt{\string\input} this .tex file right before \texttt{\string\end\{document\}} of your main file or compile it as a stand-alone document. Check the instructions on your conference's website to see if you will be asked to provide this checklist with your paper or separately.

\vspace{1em}
\hrule
\vspace{1em}

% The questions start here

\checksubsection{General Paper Structure}
\begin{itemize}

\question{Includes a conceptual outline and/or pseudocode description of AI methods introduced}{(yes/partial/no/NA)}
yes

\question{Clearly delineates statements that are opinions, hypothesis, and speculation from objective facts and results}{(yes/no)}
yes

\question{Provides well-marked pedagogical references for less-familiar readers to gain background necessary to replicate the paper}{(yes/no)}
yes

\end{itemize}
\checksubsection{Theoretical Contributions}
\begin{itemize}

\question{Does this paper make theoretical contributions?}{(yes/no)}
yes

	\ifyespoints{\vspace{1.2em}If yes, please address the following points:}
        \begin{itemize}
	
	\question{All assumptions and restrictions are stated clearly and formally}{(yes/partial/no)}
	yes

	\question{All novel claims are stated formally (e.g., in theorem statements)}{(yes/partial/no)}
	yes

	\question{Proofs of all novel claims are included}{(yes/partial/no)}
	yes

	\question{Proof sketches or intuitions are given for complex and/or novel results}{(yes/partial/no)}
	yes

	\question{Appropriate citations to theoretical tools used are given}{(yes/partial/no)}
	yes

	\question{All theoretical claims are demonstrated empirically to hold}{(yes/partial/no/NA)}
	yes

	\question{All experimental code used to eliminate or disprove claims is included}{(yes/no/NA)}
	yes
	
	\end{itemize}
\end{itemize}

\checksubsection{Dataset Usage}
\begin{itemize}

\question{Does this paper rely on one or more datasets?}{(yes/no)}
no

\ifyespoints{If yes, please address the following points:}
\begin{itemize}

	\question{A motivation is given for why the experiments are conducted on the selected datasets}{(yes/partial/no/NA)}
	NA

	\question{All novel datasets introduced in this paper are included in a data appendix}{(yes/partial/no/NA)}
	NA

	\question{All novel datasets introduced in this paper will be made publicly available upon publication of the paper with a license that allows free usage for research purposes}{(yes/partial/no/NA)}
	NA

	\question{All datasets drawn from the existing literature (potentially including authors' own previously published work) are accompanied by appropriate citations}{(yes/no/NA)}
	NA

	\question{All datasets drawn from the existing literature (potentially including authors' own previously published work) are publicly available}{(yes/partial/no/NA)}
	NA

	\question{All datasets that are not publicly available are described in detail, with explanation why publicly available alternatives are not scientifically satisficing}{(yes/partial/no/NA)}
	NA

\end{itemize}
\end{itemize}

\checksubsection{Computational Experiments}
\begin{itemize}

\question{Does this paper include computational experiments?}{(yes/no)}
yes

\ifyespoints{If yes, please address the following points:}
\begin{itemize}

	\question{This paper states the number and range of values tried per (hyper-) parameter during development of the paper, along with the criterion used for selecting the final parameter setting}{(yes/partial/no/NA)}
	yes

	\question{Any code required for pre-processing data is included in the appendix}{(yes/partial/no)}
	yes

	\question{All source code required for conducting and analyzing the experiments is included in a code appendix}{(yes/partial/no)}
	yes

	\question{All source code required for conducting and analyzing the experiments will be made publicly available upon publication of the paper with a license that allows free usage for research purposes}{(yes/partial/no)}
	yes
        
	\question{All source code implementing new methods have comments detailing the implementation, with references to the paper where each step comes from}{(yes/partial/no)}
	yes

	\question{If an algorithm depends on randomness, then the method used for setting seeds is described in a way sufficient to allow replication of results}{(yes/partial/no/NA)}
	yes

	\question{This paper specifies the computing infrastructure used for running experiments (hardware and software), including GPU/CPU models; amount of memory; operating system; names and versions of relevant software libraries and frameworks}{(yes/partial/no)}
	yes

	\question{This paper formally describes evaluation metrics used and explains the motivation for choosing these metrics}{(yes/partial/no)}
	yes

	\question{This paper states the number of algorithm runs used to compute each reported result}{(yes/no)}
	yes

	\question{Analysis of experiments goes beyond single-dimensional summaries of performance (e.g., average; median) to include measures of variation, confidence, or other distributional information}{(yes/no)}
	yes

	\question{The significance of any improvement or decrease in performance is judged using appropriate statistical tests (e.g., Wilcoxon signed-rank)}{(yes/partial/no)}
	yes

	\question{This paper lists all final (hyper-)parameters used for each model/algorithm in the paper’s experiments}{(yes/partial/no/NA)}
	yes

\end{itemize}
\end{itemize}
\ifreproStandalone
\end{document}
\fi

\section{\jhedit{Potential impacts}}\label{sec:impacts}
\jhedit{Our approach could substantially lower the computational resources required to address complex, real-world CMDP problems. Although we have not identified any immediate negative societal consequences, we will continue to investigate the broader effects of deploying the methods introduced in this paper.}
%%%%%%%%%%%%%%%%%%%%%%%%%%%%%%%%%%%%%%%%%%%%%%%%%%%%%%%%%%%%
\end{document}